%% file: main.tex
\documentclass[10pt,twocolumn,letterpaper]{article}

\usepackage{wacv}
\usepackage{times}
\usepackage{epsfig}
\usepackage{graphicx}
\usepackage{amsmath}
\usepackage{amssymb}

\usepackage{subfigure}
\usepackage{caption}
\usepackage{url}
\usepackage{multirow}
\usepackage{verbatim} 

\usepackage[pagebackref=true,breaklinks=true,letterpaper=true,colorlinks,bookmarks=false]{hyperref}

\wacvfinalcopy 

\def\wacvPaperID{535}

\ifwacvfinal\fi
\title{Self-Contained Stylization via Steganography\\for Reverse and Serial Style Transfer}

\author{Hung-Yu Chen$^{\ast\dagger}\qquad$ I-Sheng Fang$^{\ast\dagger}\qquad$ Wei-Chen Chiu\\
National Chiao Tung University, Taiwan\\
{\tt\small chen3381@purdue.edu $\quad$ nf0126@gmail.com $\quad$ walon@cs.nctu.edu.tw}
}

\ifwacvfinal\fi

\newcommand\blfootnote[1]{%
  \begingroup
  \renewcommand\thefootnote{}\footnote{#1}%
  \addtocounter{footnote}{-1}%
  \endgroup
}

\begin{document}

\twocolumn[{
    \maketitle
    \includegraphics[width=0.975\textwidth]{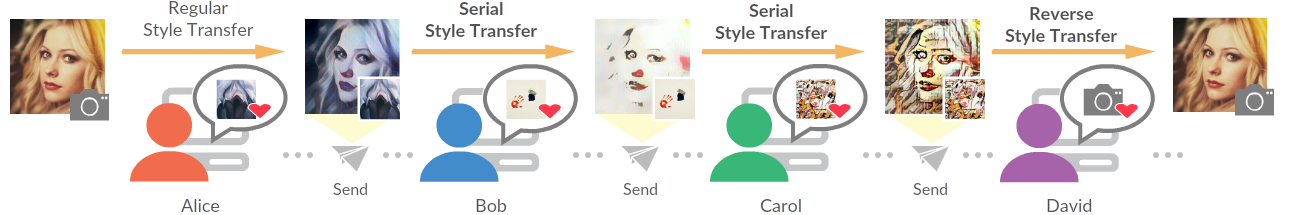}
    \captionsetup{hypcap=false} 
    \captionof{figure}{As style transfer is widely used in social networks such as Facebook or Instagram nowadays, two practical applications related to style transfer are studied in this paper and can be illustrated with a scenario here. \textbf{(1) Serial style transfer}: when Bob receives a stylized image from Alice, he can easily modify its style into any other arbitrary one and further share the output to others (e.g. Carol). \textbf{(2) Reverse style transfer}: any user (e.g. David) who receives a stylized image can easily reverse it back to its original photo, i.e. analogous to de-style, without requiring any additional information. Our proposed self-contained stylization approach is capable of tackling
    these two applications hence could be quite useful for the social networks, since the users of a social network tend to share photos and stylized images on the same platform.
    }
    \label{fig:teaser}
    \vspace{1em}
}]

 \captionsetup{hypcap=true} 

\begin{abstract}
Style transfer has been widely applied to give real-world images a new artistic look. However, given a stylized image, the attempts to use typical style transfer methods for de-stylization or transferring it again into another style usually lead to artifacts or undesired results. We realize that these issues are originated from the content inconsistency between the original image and its stylized output. Therefore, in this paper we advance to keep the content information of the input image during the process of style transfer by the power of steganography, with two approaches proposed: a two-stage model and an end-to-end model. We conduct extensive experiments to successfully verify the capacity of our models, in which both of them are able to not only generate stylized images of quality comparable with the ones produced by typical style transfer methods, but also effectively eliminate the artifacts introduced in reconstructing original input from a stylized image as well as performing multiple times of style transfer in series.
\end{abstract}
\blfootnote{$^\dagger$Hung-Yu Chen and I-Sheng Fang are now with Purdue University and National Cheng Chi University respectively.}
\blfootnote{$^\ast$Both authors contribute equally.}

\input{intro.tex}

\input{related.tex}

\input{method.tex}

\input{experiment.tex}

\input{conclusion.tex}

{\small
\bibliographystyle{ieee}
\bibliography{egbib}
}

\clearpage
\input{appendix.tex}

\end{document}

%% file: intro.tex
\section{Introduction}\label{sec:intro}
A style transfer approach typically aims to modify an input photo such that its content can be preserved but the associated style is revised as the one of a reference image. In comparison to the classical methods which generally rely on matching color statistics between the reference image and modified output~\cite{ImageAnalogies, shih2013data}, the recent development of deep learning has brought a great leap by being able to capture high-level representation for the content and style of images, thus producing more photorealistic stylization.
In particular, after the advent of first deep-learning style transfer work~\cite{Gatys2016ImageStyleTransferUsingCNN}, many research efforts~\cite{Gatys2015TextureSynthesisUsingCNN,Gatys2016ImageStyleTransferUsingCNN, Johnson2016Perceptual, Ulyanov2016TextureNetworks} have gone with the trend to propose faster, more visually appealing, and more universal algorithms for the task of style transfer.

Without loss of generality, as these approaches basically perform transformation on the content feature of input photo according to the style feature from reference image, the appearance of photo is usually altered to have various colors or textures, which inevitably causes changes to the fine-grained details in content information. Consequently, the stylized output no longer has the same content feature as its original photo, leading to some issues for two novel applications that we proposed in this paper: \textit{serial} and \textit{reverse} style transfer. The former attempts to transfer an image, which is already stylized, into another arbitrary style; while the latter aims to remove the stylization effect of a stylized image and turn back to its original photo, as the example scenario illustrated in the Figure~\ref{fig:teaser}. Particularly, what we expect to obtain for the serial style transfer is that, even after applying multiple times of different stylizations, the final output should be similar to the one which is produced by directly transferring the original photo into the latest style (i.e. not influenced by the previous stylization). 
We believe that having both serial and reverse style transfer can open the door to exciting new ways for users to interact with style transfer algorithms, not only allowing the freedom to perform numerous stylizations on a photo with having its content well preserved, but also providing the access to the original input by recovering from a stylized image.

\begin{figure}[t]
    \centering
	\includegraphics[width=\columnwidth]{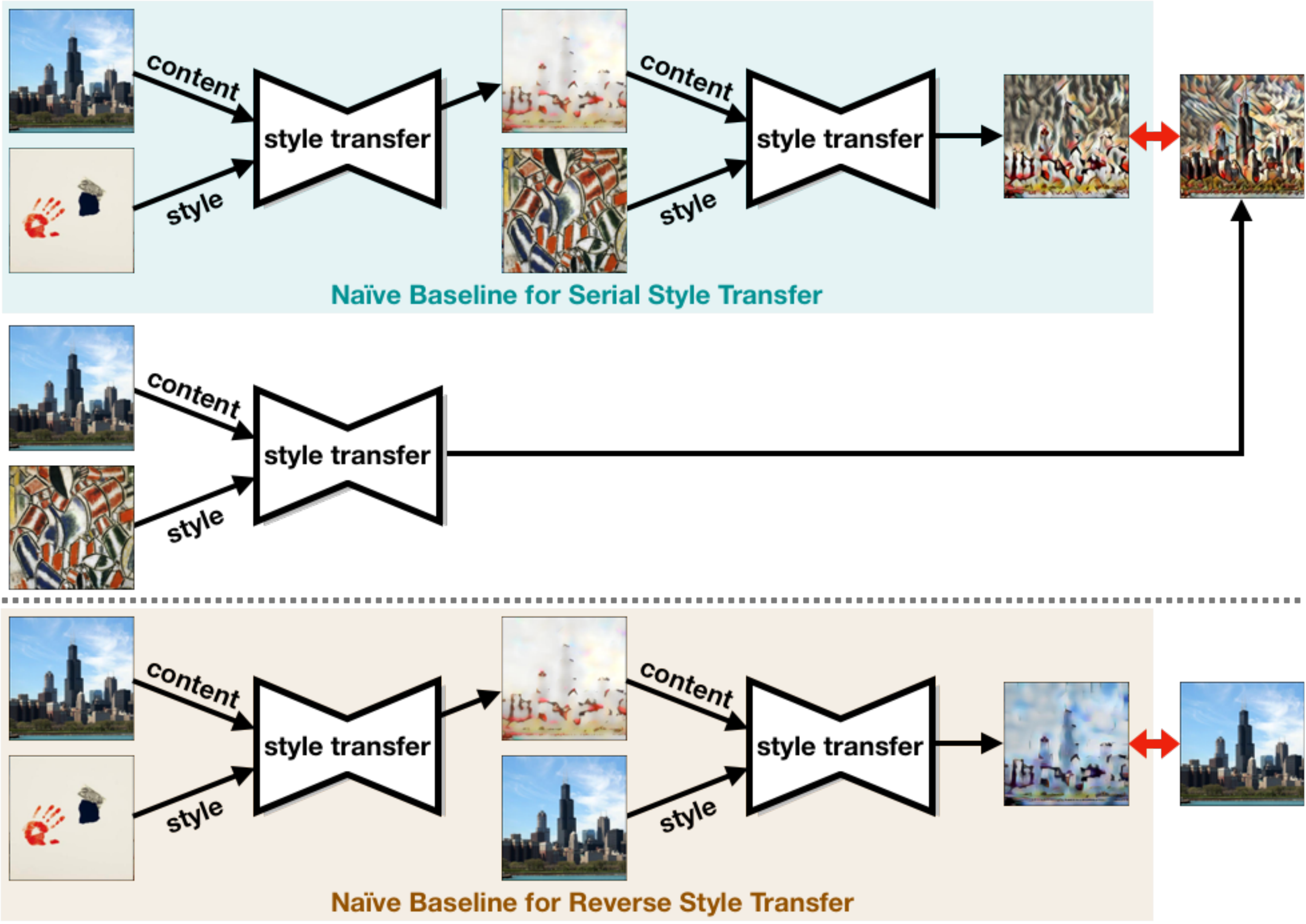}
	\caption{Illustration of issues for serial and reverse style transfer. The colored regions present the na\"ive baselines for both tasks based on typical style transfer approach, while the rightmost figures are the expected results accrodingly.}
    \label{fig:issue}
\end{figure}

Although these two problems intuitively seem easy to solve by performing style transfer again on the stylized image with taking the image of another artistic style or the original photo as the source of stylization respectively, the results are usually not visually satisfying and lose the content consistency. For instance, when two style transfer operations are performed in series, such characteristic brings artifacts to the final output and makes it significantly distinct from the result obtained by applying the second style transfer to the original input, as shown in the upper part of Figure~\ref{fig:issue}. Similarly, upon taking a stylized image and its corresponding original photo as sources of content and style respectively, we are not able to achieve reverse stylization of reconstructing the original input, as shown in the lower part of Figure~\ref{fig:issue}. Furthermore, there could exist a potential argument that both reverse and serial style transfer are simple once the original photo is always transmitted with the stylized image. However, this na\"ive solution doubles the bit-rates for transmission thus being quite inefficient for sharing stylized images on the internet or social networks.

To tackle the aforementioned issues, it calls for a framework which can not only generate visually appealing stylized images as typical style-transfer approaches do, but also maintain the important representations related to the content feature of input photo, so that the content inconsistency between the stylized image and the original photo can be compensated afterwards. In this paper we propose to achieve so by integrating the power of \textbf{steganography} \cite{Gupta2012LSBSteganography, Baluja2017DeepSteganography, HiDDeN} into style transfer approaches, where the content information of input photo is hidden into the style-transferred output with steganographic method. 
With a decoder trained to extract the hidden information from the stylized image produced by our proposed method, the issue of having severe artifacts while doing reverse or serial transfer could be resolved. 
As the content information is self-contained in the stylized image via the use of steganography, in the scenario of Figure~\ref{fig:teaser}, the serial and reverse style transfer are now naturally achievable without any additional cost of transmitting the original photo or any other forms of attaching data. It is also worth noting that, with a simple extension on our approach such as a gate to control whether the content information of original photo is provided for the steganography component or not, the users can easily control the usage right of their stylized images, i.e. allowing or forbidding the images to be further style-transferred or de-stylized.

We implement the idea with two different deep-learning architectures, where one is a two-stage model and the other one is an end-to-end model, as going to be detailed in Section \ref{sec:method}. 
The two-stage model needs to hide a bigger amount of information into the image, but can be coupled with various style transfer methods, leading to a better adaptability; the end-to-end model is highly dependent on the traits of AdaIN \cite{Huang2017AdaIN}, but it only needs to encrypt a small amount of information into the image, being more robust to the potential error accumulation during multiple serial style transfers. 
We conduct extensive experimental validation comparing to several baselines and demonstrate the efficacy of our proposed method to advance serial and reverse style transfer.

%% file: related.tex
\section{Related Works}\label{sec:related}
\paragraph{Style transfer.} 
Giving images a new artistic style or texture has long been a topic that attracts researchers' attention. Some of the early research works prior to the renaissance of deep learning tackle the style transfer by simply matching the characteristic in color, or searching for the correspondences across source and style images \cite{ImageQuilting, ImageAnalogies}. Instead of using low-level feature cues as early works, Gatys \etal \cite{Gatys2015TextureSynthesisUsingCNN, Gatys2016ImageStyleTransferUsingCNN} utilize representations obtained from the pre-trained convolutional neural network (CNN) to extract more semantic description on the content and style features of images. Their methods can generate visually appealing results; however, it is extremely slow due to iterative optimization for matching style features between the output and style image. 

In order to speed up the process of image style transfer, several feed-forward approaches (e.g. \cite{Johnson2016Perceptual, Ulyanov2016TextureNetworks}) are proposed, which directly learn feed-forward networks to approximate the iterative optimization procedure with respect to the same objectives. The style transfer now can be carried out in real time, however, there usually exists a trade-off between the the processing speed and the image quality of the stylized output. For example, the result of \cite{Johnson2016Perceptual} suffers from repetitive patterns in plain area. Fortunately, Ulyanov \etal~\cite{Ulyanov2016TextureNetworks} uncover that the image quality produced by the network of \cite{Johnson2016Perceptual} could be greatly improved through replacing its batch normalization layers (BN) with instance normalization (IN) ones, while \cite{Dumoulin2017ConditionalIN} steps further to introduce conditional instance normalization and learns to perform real-time style transfer upon multiple styles that have been seen during training. Nevertheless, all these feed-forward models are typically constrained to particular styles and hardly generalizable to arbitrary stylization. That's where adaptive instance normalization (AdaIN) \cite{Huang2017AdaIN} comes into play.

AdaIN could be roughly seen as IN with a twist. It basically follows the IN steps, except now the content feature of input photo is first normalized then affine-transformed by using the mean and standard deviation of the style features of style image. This operation matches the statistics of content and style features in order to transfer the input photo into an arbitrary style, since the parameter applied in AdaIN is dependent on the target style. Given a content feature $x$ and a style feature $y$, the procedure of AdaIN is: 
\begin{equation} \label{eq:AdaIN}
\text{AdaIN}(x,y)=\sigma(y)\left(\frac{x-\mu(x)}{\sigma(x)}\right)+\mu(y)
\end{equation}
where $\mu$ and $\sigma$ denote the mean and standard deviation respectively.
There are other research works \cite{Li2017WCT, Sheng2018Avatar-Net} sharing the similar idea with AdaIN, where various manners are introduced for adaptively transforming the content feature of input photo in accordance to the style image. Since the simplicity of AdaIN and its ability for universal style transfer, we utilize AdaIN as the base model in our proposed method for style transfer and make extensions for handling issues of serial and reverse stylizations.

\paragraph{Image de-stylization / Reverse style transfer.}
As image style transfer typically applies artistic styles to the input images, image de-stylization or reverse style transfer attempts to remove those styles from the stylized images and recover them back to their original appearance. To the best of our knowledge, only a handful of research works tackle this topic. Shiri \etal \cite{FaceDestylization,FaceRecoveryfromPortraits} explore the field of image de-stylization with a particular focus on human faces. Their methods learn a style removal network to recover the photo-realistic face images from the stylized portraits and retain the identity information. However, they rely on the specific properties of human faces, so it can hardly be generalized to other object categories. \cite{tomei2019art2real} proposes to translate artworks to photo-realistic images, which is similar to de-stylization but is limited to only few artistic styles.
The na\"ive approach of having the original input as the style image and other methods from the image-to-image translation area (e.g.  CycleGAN~\cite{CycleGAN2017} or Pix2Pix~\cite{isola2017image}) are also incapable of achieving image de-stylization or only applicable to the seen styles (thus not universal), as already shown in \cite{FaceRecoveryfromPortraits}.

\paragraph{Image steganography.}
Image steganography is a way to deliver a message secretly by hiding it into in an irrelevant cover image while minimizing the perturbations within the cover, and has been studied for a long period in the area of image processing \cite{kessler2011overview, cheddad2010digital}. In general, traditional approaches rely on carefully and manually designed algorithms to achieve both message hiding and retrieval from the cover image. Some examples of such methods would be HUGO \cite{Pevny2010HUGO} and least significant bit steganography \cite{Gupta2012LSBSteganography}. 

After the application of deep learning has grown popular, few research works \cite{hayes2017generating,Baluja2017DeepSteganography, HiDDeN} explore the possibility of having deep neural networks perform steganography on images, where the hiding and revealing processes are learned together in the manner of end-to-end training. \cite{hayes2017generating} proposes to train the steganographic algorithm and steganalyzer jointly via an adversarial scheme between three-players. In comparison to handling lower bit rates of \cite{hayes2017generating}, \cite{Baluja2017DeepSteganography} intends to hide an image entirely into another image of the same size, but has a potential drawback of being detectable. For the method proposed in \cite{HiDDeN}, it hides relatively smaller amount of message into an irrelevant cover image but specifically tackles the problem of making the hidden message robust to noises.

%% file: method.tex
\section{Proposed Methods}\label{sec:method}

\input{method_2_stage.tex}

\input{method_e2e.tex}

%% file: method_2_stage.tex
\subsection{Two-Stage Model}\label{sec:2stage}
\begin{figure*}[t]
    \centering
    \subfigure[Visualization of the architecture and training objectives for our proposed two-stage model, which is composed of a style transfer stage (shaded in purple) and a steganography stage (shaded in green).
    \label{fig:2st_training}]{\includegraphics[width=450pt]{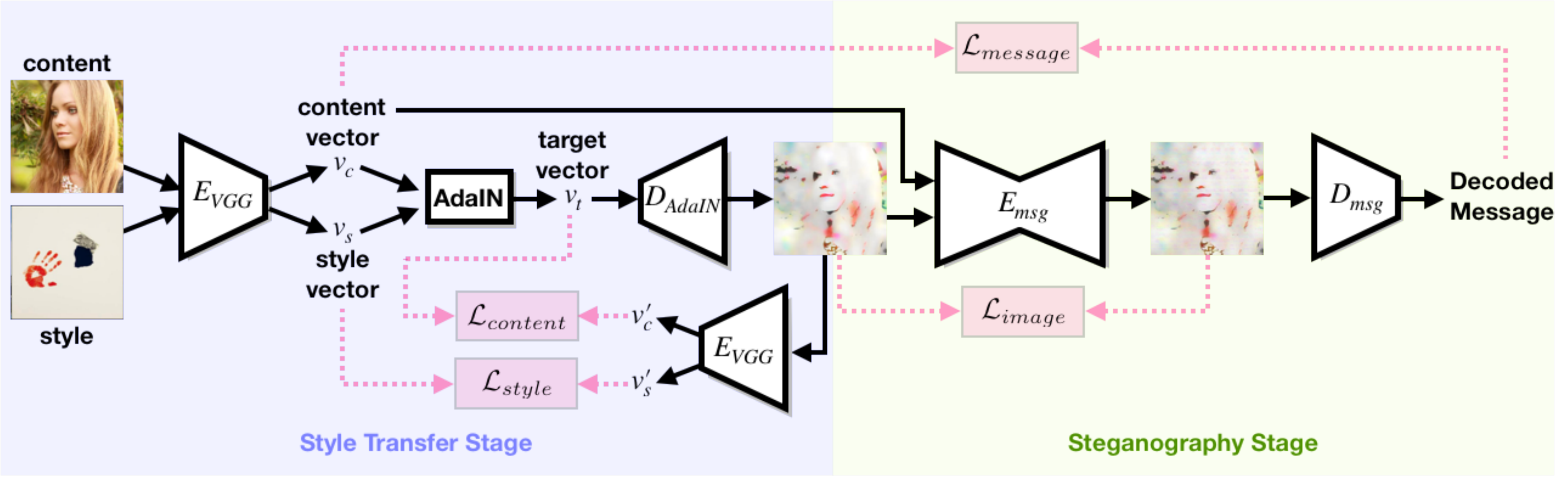}}
    \subfigure[Illustrations of how to apply our two-stage model in the tasks of reverse and serial style transfer respectively.
    \label{fig:2st_resolve_issues}]{\includegraphics[width=450pt]{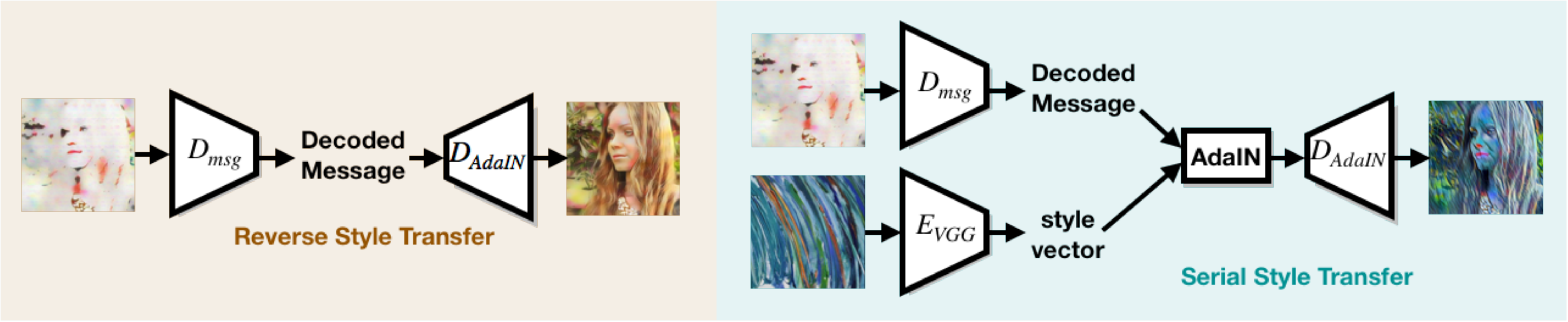}}
    \caption{Overview of the training and testing procedure of our two-stage model. Please refer to Section~\ref{sec:2stage} for details.}
\end{figure*}

Our two-stage model is a pipeline built upon a straightforward integration of style transfer and steganography networks, as shown in Figure~\ref{fig:2st_training}.
In the first stage, we stylize the content image $I_c$ according to the style image $I_s$ based on a style transfer model. Afterward in the second stage, the steganography network learns an encoder to hide the content information of $I_c$ into the stylized image $I_t$ from the previous stage, as well as a paired decoder which is able to retrieve the hidden information from the encoded image.

\subsubsection{Style Transfer Stage}\label{sec:2stage_st}
We adopt AdaIN ~\cite{Huang2017AdaIN} as our primary reference method while our two-stage model is capable of incorporating with other style transfer algorithms (e.g. WCT~\cite{Li2017WCT} or~\cite{Ulyanov2016TextureNetworks}, please refer to the supplementary materials for more details). The architecture is composed of a pre-trained VGG19~\cite{VGG} encoder $E_{VGG}$ and a decoder $D_{AdaIN}$. The concept of the training procedure can be briefly summarized as follows:

1) the encoder extracts the content feature $v_c = E_{VGG}(I_c)$ and style feature $v_s = E_{VGG}(I_s)$  from content image $I_c$ and style image $I_s$ respectively; 
2) based on Eq.~\ref{eq:AdaIN}, the content feature $v_c$ is adaptively redistributed according to the statistics of style feature $v_s$ to obtain the target feature $v_t$; 3) the stylized output $I_t$ is finally produced by $D_{AdaIN}(v_t)$. 
While encoder $E_{VGG}$ is pre-trained and fixed (based on first few layers of VGG19 up to \texttt{relu4\_1}), the learning of AdaIN style transfer focuses on training $D_{AdaIN}$ with respect to the objective (same as in AdaIN~\cite{Huang2017AdaIN}):
\begin{equation}
\begin{aligned}
\mathcal{L}_{style-transfer} = \mathcal{L}_{content} + \lambda_{style} \mathcal{L}_{style}
\label{eq:L_first_stage}
\end{aligned}
\end{equation}
in which the content loss  $\mathcal{L}_{content}$ and style loss $\mathcal{L}_{style}$ are defined as follows with their balance controlled by $\lambda_{style}$ (which is set to $10$ in all our experiments, as used in~\cite{Huang2017AdaIN}):
\begin{equation}
\begin{aligned}
\mathcal{L}_{content} = &\left \|E_{VGG}(I_t)-v_t \right\|_2 \label{eq:L_content_two_stage}
\end{aligned}
\end{equation}
\begin{equation}
\begin{aligned}
\mathcal{L}_{style} = &\sum_{i}^{L}\left \|\mu(l_i(I_t))- \mu(l_i(I_s))\right \|_2 +\\&\sum_{i}^{L}\left \|\sigma(l_i(I_t))- \sigma(l_i(I_s))\right \|_2 \label{eq:L_style_two_stage}
\end{aligned}
\end{equation}
where each $l_i$ denotes the feature map obtained from a layer in VGG19, and $L=\{$ \texttt{relu1\_1}, \texttt{relu2\_1}, \texttt{relu3\_1}, \texttt{relu4\_1} $\}$ in our experiments.

In addition to the objective function above which encourages $D_{AdaIN}(v_t)$ to output the stylized image $I_t$ with its content feature $E_{VGG}(I_t)$ close to target $v_t$ and similar style as $I_t$, we also train $D_{AdaIN}$ with \textit{identity mapping}, i.e. reconstructing content image $\tilde{I}_c$ solely from its content feature $v_c$, for the purpose of better dealing with reverse style transfer later on. To achieve identity mapping, we occasionally place the same photo for both content and style images during the training of $D_{AdaIN}$, so that the content feature $v_c$ and target feature $v_t$ are identical. Thus, the output $I_t$ of $D_{AdaIN}$ is similar to $I_c$ by the same objectives as Eq.~\ref{eq:L_content_two_stage}.

\subsubsection{Steganography Stage}
The steganography stage in our model contains a message encoder $E_{msg}$ and a corresponding message decoder $D_{msg}$. 
The message encoder $E_{msg}$ aims to hide content feature $v_c$ into stylized image $I_t$ and produce the encoded image $I_e = E_{msg}(I_t, v_c)$, which is exactly the output of our two-stage model, while the message decoder $D_{msg}$ tries to decode $v_c$ out from $I_e$.
As the typical scheme of steganography, the difference between the encoded image $I_e$ and stylized image $I_t$ should be visually undetectable, therefore the $E_{msg}$ is trained to minimize the objective defined as: 
\begin{equation}
\label{eq:L_image_two_stage}
\mathcal{L}_{image} = \left\|I_e-I_t\right\|_2
\end{equation}
On the other hand, the message decoder $D_{msg}$ is optimized to well retrieve the message $v_c$ hidden in $I_e$, with respect to the objective:
\begin{equation}
\label{eq:L_msg_two_stage}
\mathcal{L}_{message} = \left\|D_{msg}(I_e)-v_c\right\|_2
\end{equation}
where the architecture designs of both $E_{msg}$ and $D_{msg}$ follow the ones used in the recent steganography paper~\cite{HiDDeN}. 

\noindent The objective for the steganography stage is summarized as:
\begin{equation}
\label{eq:L_steganography}
 \mathcal{L}_{steganography} = \lambda_{img}\mathcal{L}_{image} + \lambda_{msg}\mathcal{L}_{message}
\end{equation}
where $\lambda_{img}$ and $\lambda_{msg}$ are used to balance $\mathcal{L}_{image}$ and $\mathcal{L}_{message}$ respectively.

\begin{figure*}[t]
    \centering
    \includegraphics[width=450pt]{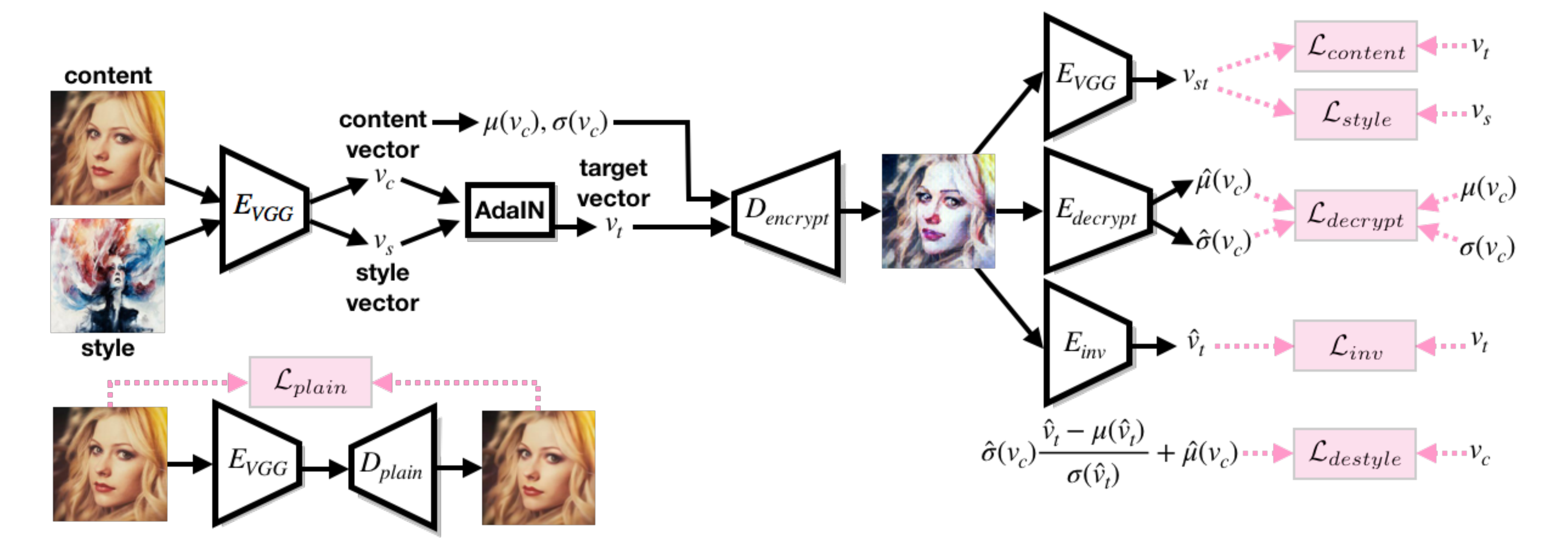}
    \caption{Overview of our end-to-end model and its training objectives, where the image stylization and content information encryption are now particularly performed jointly in a single network $D_{encrypt}$. Please refer to Section~\ref{subsec:e2eModel} for more details.}
    \label{fig:e2e_train_model}
\end{figure*}

\begin{figure*}[t]
    \centering
    \includegraphics[width=450pt]{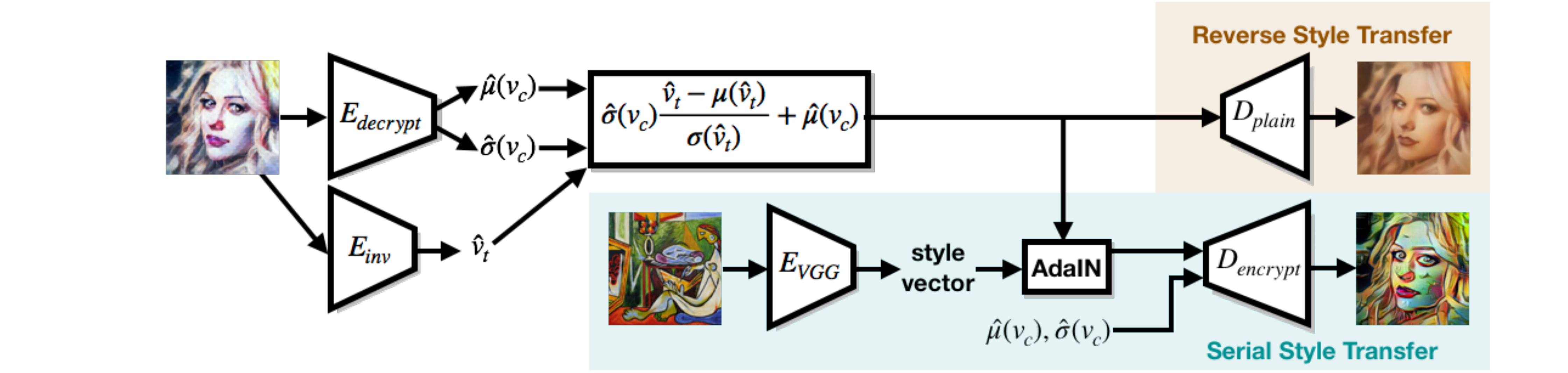}
    \caption{Illustration of applying proposed end-to-end model, especially the reconstructed content feature (cf. Eq.~\ref{eq:v_c_reconstruction}), for both tasks of reverse and serial style transfer network, denoted by different colors respectively.}
    \label{fig:e2e_test_model}
\end{figure*}

\subsubsection{Reverse \& Serial Stylization by Two-Stage Model}
\noindent\textbf{Reverse style transfer.}
As shown in the left portion of Figure~\ref{fig:2st_resolve_issues}, by using the decoder $D_{AdaIN}$, which is capable of performing identity mapping, 
to decode the content feature $v_c$ from a given encoded image $I_e$,
the original content image $I_c$ can now be recovered by $D_{AdaIN}(D_{msg}(I_e))$. \\
\noindent\textbf{Serial style transfer.}
To transfer the encoded image $I_e$ (which is already stylized) into another style given by ${I}'_s$, as shown in the right portion of Figure~\ref{fig:2st_resolve_issues}, the content feature $v_c=D_{msg}(I_e)$ decoded from $I_e$ and the style feature ${v}'_s=E_{VGG}({I}'_s)$ extracted from ${I}'_s$ are taken as inputs for AdaIN transformation, then the serial style transfer is achieved by computing ${I}'_t=D_{AdaIN}(\text{AdaIN}(D_{msg}(I_e),{v}'_s))$. In addition, performing multiple times of style transfer in series is naturally doable when the steganography is applied for encoding $v_c$ into ${I}'_t$ again.

%% file: method_e2e.tex
\subsection{End-to-End Model} \label{subsec:e2eModel}

\newcommand\theLength{5em}
\begin{figure*}[htbp]
\centering
\begin{tabular}{cc||cccc}
\parbox[c]{\theLength{}}{\includegraphics[width=44pt]{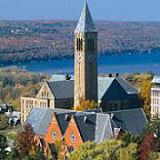}}&
\parbox[c]{\theLength{}}{\includegraphics[width=44pt]{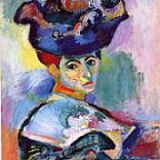}}&
\parbox[c]{\theLength{}}{\includegraphics[width=44pt]{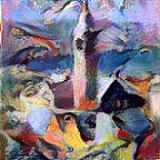}}&
\parbox[c]{\theLength{}}{\includegraphics[width=44pt]{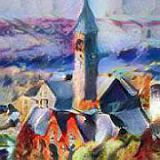}}&
\parbox[c]{\theLength{}}{\includegraphics[width=44pt]{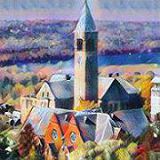}}&
\parbox[c]{\theLength{}}{\includegraphics[width=44pt]{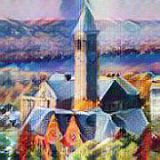}}\\
\parbox[c]{\theLength{}}{\includegraphics[width=44pt]{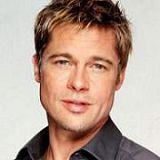}}&
\parbox[c]{\theLength{}}{\includegraphics[width=44pt]{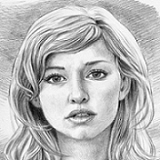}}&
\parbox[c]{\theLength{}}{\includegraphics[width=44pt]{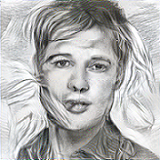}}&
\parbox[c]{\theLength{}}{\includegraphics[width=44pt]{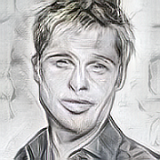}}&
\parbox[c]{\theLength{}}{\includegraphics[width=44pt]{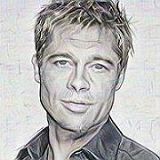}}&
\parbox[c]{\theLength{}}{\includegraphics[width=44pt]{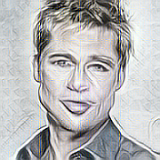}}\\
\parbox[c]{\theLength{}}{\includegraphics[width=44pt]{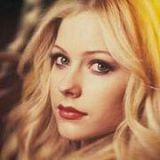}}&
\parbox[c]{\theLength{}}{\includegraphics[width=44pt]{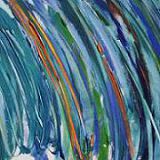}}&
\parbox[c]{\theLength{}}{\includegraphics[width=44pt]{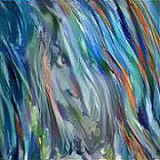}}&
\parbox[c]{\theLength{}}{\includegraphics[width=44pt]{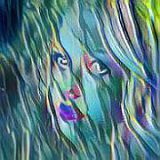}}&
\parbox[c]{\theLength{}}{\includegraphics[width=44pt]{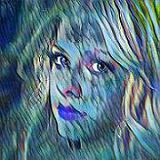}}&
\parbox[c]{\theLength{}}{\includegraphics[width=44pt]{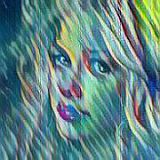}}\\
 Content & Style & Gatys~\cite{Gatys2016ImageStyleTransferUsingCNN} & AdaIN~\cite{Huang2017AdaIN} & Two-stage & End-to-end\\
\end{tabular}
\caption{Example results of regular style transfer produced by different methods. First two columns show the pairs of content/style images; third to last columns present the results from Gatys~\cite{Gatys2016ImageStyleTransferUsingCNN}, AdaIN~\cite{Huang2017AdaIN}, our two-stage model, and our end-to-end model respectively. We observe that our models are able to generate results with quality comparable to the baselines.}
\label{fig:result_regular_compare}
\end{figure*}

Aside from the two-stage model which can take several style transfer methods as its base (e.g. WCT~\cite{Li2017WCT} or~\cite{Ulyanov2016TextureNetworks}, please refer to our supplementary material), our end-to-end model digs deeply into the characteristic of AdaIN for enabling image stylization and content information encryption simultaneously in a single network.
As we know, the procedure of AdaIN (cf. Eq.~\ref{eq:AdaIN}) produce a target feature $v_t$ by transforming the content feature $v_c$ to match the statistics of the style feature $v_s$, i.e. mean $\mu(v_s)$ and standard deviation $\sigma(v_s)$. Assume there exists an inverse function which can estimate the corresponding target feature $v_t$ of a stylized image $I_{st}$, we hypothesize that the content feature $v_c$ is derivable from $v_t$ by $\sigma(v_c)\frac{v_t - \mu(v_t)}{\sigma(v_t)}+\mu(v_c)$ once its original statistics $\left\{ \mu(v_c), \sigma(v_c)\right\}$ is available. 

Based on this hypothesis, our end-to-end model is designed to have several key components as shown in Figure~\ref{fig:e2e_train_model}: 1) a encrypted image decoder $D_{encrypt}$ which takes $v_t, \mu(v_c), \sigma(v_c)$ as input and produce a stylized image $I_{st}$ with $\left\{ \mu(v_c), \sigma(v_c)\right\}$ 
being encrypted into it; 2) a decrypter $E_{decrypt}$ which is able to decrypt $\left\{ \mu(v_c), \sigma(v_c)\right\}$ out from $I_{st}$; and 3) an inverse target encoder $E_{inv}$ which is capable of estimating $v_t$ from the given stylized image $I_{st}$. 
In the following and Figure~\ref{fig:e2e_train_model} we detail the overall computation of our model and the objectives for training. 

First, the output image $I_{st}$ of the encrypted image decoder $D_{encrypt}$, which is simultaneously encrypted and stylized, should still have the similar content/style feature as the one in the content/style image respectively (i.e. $\left\{ v_c, v_s\right\}$). The same objective functions, $\mathcal{L}_{content}$ and $\mathcal{L}_{style}$, defined in Eq.~\ref{eq:L_content_two_stage}, can then be adopted to optimize $D_{encrypt}$ by simply replacing $I_t$ with $I_{st}$ here.

Second, we see that the $\left\{ \mu(v_c), \sigma(v_c)\right\}$ encrypted into $I_{st}$ with $D_{encrypt}$ should be retrievable by using the corresponding decrypter $E_{decrypt}$. Therefore, the output of $E_{decrypt}$, $\left\{ \hat{\mu}(v_c), \hat{\sigma}(v_c)\right\}$, is compared to the original $\left\{ \mu(v_c), \sigma(v_c)\right\}$, leading to the decryption loss $L_{decrypt}$ for jointly optimizing $D_{encrypt}$ and $E_{decrypt}$:
\begin{equation}
\mathcal{L}_{decrypt} = \left \| \hat{\mu}(v_c) - \mu(v_c) \right\|_2 + \left \| \hat{\sigma}(v_c) - \sigma(v_c) \right\|_2
\label{eq:e2eDecryptLoss}
\end{equation}
 
 Third, as motivated in our hypothesis, there should be an inverse target encoder $E_{inv}$ which is able to recover the target vector $v_t$ used for generating $I_{st}$. It is worth noting that, here we design $E_{inv}$ to have the same architecture as $E_{VGG}$, but it is trained to ignore the influence caused by the encrypted information in $I_{st}$ and focus on retrieving the target vector $v_t$. With denoting the feature vector estimated by $E_{inv}$ as $\hat{v_t} = E_{inv}(I_{st})$, the objective function for training $E_{inv}$ is then defined as 
\begin{equation}
\mathcal{L}_{inv} = \left \| \hat{v_t} - v_t \right\|_2
\label{eq:e2eInverseLoss}
\end{equation}

Fourth, with having $\left \{ \hat{\mu}(v_c), \hat{\sigma}(v_c)\right \}$ and $\hat{v_t}$ obtained from $E_{decrypt}$ and $E_{inv}(I_{st})$ respectively, we can reconstruct the content feature $\hat{v}_c$ according to:
\begin{equation}
\hat{v}_c = \hat{\sigma}(v_c) \frac{\hat{v_t} - \mu(\hat{v_t})}{\sigma(\hat{v_t})} + \hat{\mu}(v_c)
\label{eq:v_c_reconstruction}
\end{equation}
Then an objective is defined based on the difference between $\hat{v}_c$ and the original $v_c$:
\begin{equation}
\mathcal{L}_{destyle} = \left \| \hat{v}_c - v_c \right\|_2
\label{eq:e2eDestyleLoss}
\end{equation}
where it can update $D_{encrypt}$, $E_{decrypt}$, and $E_{inv}$ jointly.

Fifth, as a similar idea of having identity mapping in our two-stage model, here we learn a plain image decoder $D_{plain}$ which can map a content feature back to the corresponding content image $I_c$.
Its training is simply done by: 
\begin{equation}
\mathcal{L}_{plain} = \left \| D_{plain}(E_{VGG}(I_c)) - I_c \right\|_2
\label{eq:e2ePlainLoss}
\end{equation}

\noindent The overall objective function $\mathcal{L}_{end2end}$ for our end-to-end model training is then summarized as below, where $\lambda$ parameters are used to balance weights of different losses:

\begin{equation}
\begin{aligned}
\label{eq:L_end_to-end}
 \mathcal{L}_{end2end} &= \lambda_{c}\mathcal{L}_{content} + \lambda_{s}\mathcal{L}_{style}+ \lambda_{dec}\mathcal{L}_{decrypt}\\&  + \lambda_{inv}\mathcal{L}_{inv} + \lambda_{des}\mathcal{L}_{destyle}+\lambda_{p}\mathcal{L}_{plain}
 \end{aligned}
\end{equation}

\subsubsection{Reverse \& Serial Stylization by End-to-End Model}
After our end-to-end model are properly trained, since the content vector of the original content image can be reconstructed by using Eq.~\ref{eq:v_c_reconstruction}, the reverse and serial style transfer are now straightforwardly achievable, as shown in Figure~\ref{fig:e2e_test_model}. \\
\textbf{Reverse style transfer.}
The reverse style transfer, which recovers the original image based on a stylized image $I_{st}$, is done by having the decrypted content feature $\hat{v}_c$ go through the plain image decoder $D_{plain}$.\\

\textbf{Serial style transfer.}
Given a stylized image $I_{st}$, by decrypting content feature $\hat{v}_t$ from $I_{st}$ and extracting style feature ${v}'_s$ from a new style image, the serial style transfer is then produced based on $D_{plain}(\text{AdaIN}(\hat{v}_c, {v}'_s))$. Please note here we can encrypt the statistics of content feature into the output again, as shown in the lower part of Figure~\ref{fig:e2e_test_model}, for enabling multiple times of style transfer in series.

%% file: experiment.tex
\section{Experiment}\label{sec:exp}
\noindent \textbf{Dataset}~ We follow the similar setting in~\cite{Huang2017AdaIN} to build up the training set for our models. We randomly sample 10K content and 20K style images respectively from the training set of MS-COCO~\cite{lin2014microsoft} and the training set of WikiArt~\cite{wiki_art}. These training images are first resized to have the smallest dimension be 512 while the aspect ratio is kept, then randomly cropped to the size of 256$\times$256. 

\newcommand\shorterLength{4.8em}
\newcommand\imgLen{34pt}
\begin{figure*}[htbp]
\footnotesize
\centering
\begin{tabular}{p{33pt}p{3pt}p{33pt}p{33pt}p{33pt}p{43pt}| p{33pt}p{3pt}p{33pt}p{33pt}p{33pt}p{33pt}}
\parbox[c]{\imgLen{}}{\includegraphics[width=44pt]{figures/content_img/avril.jpg}}&
\begin{tabular}{r}
\scriptsize (1) 
\end{tabular}&
\parbox[c]{\imgLen{}}{\includegraphics[width=44pt]{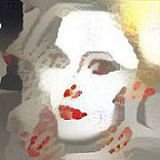}}&
\parbox[c]{\imgLen{}}{\includegraphics[width=44pt]{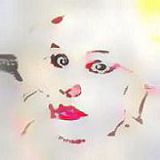}}&
\parbox[c]{\imgLen{}}{\includegraphics[width=44pt]{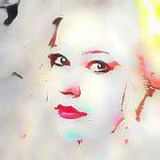}}&
\parbox[c]{\imgLen{}}{\includegraphics[width=44pt]{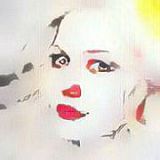}}&
\parbox[c]{\imgLen{}}{\includegraphics[width=44pt]{figures/content_img/cornell.jpg}}&
\begin{tabular}{r}
\scriptsize (1) 
\end{tabular}&
\parbox[c]{\imgLen{}}{\includegraphics[width=44pt]{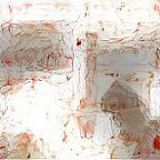}}&
\parbox[c]{\imgLen{}}{\includegraphics[width=44pt]{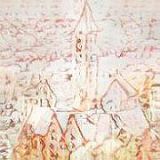}}&
\parbox[c]{\imgLen{}}{\includegraphics[width=44pt]{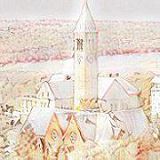}}&
\parbox[c]{\imgLen{}}{\includegraphics[width=44pt]{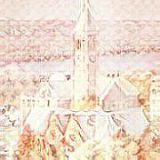}}\\
\raisebox{17pt}{
\begin{tabular}{c}
Content
\end{tabular}}&
\begin{tabular}{r}
{\scriptsize (2) }
\end{tabular}&
\parbox[c]{\imgLen{}}{\includegraphics[width=44pt]{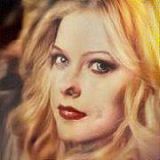}}&
\parbox[c]{\imgLen{}}{\includegraphics[width=44pt]{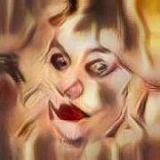}}&
\parbox[c]{\imgLen{}}{\includegraphics[width=44pt]{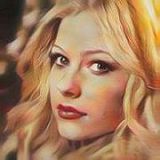}}&
\parbox[c]{\imgLen{}}{\includegraphics[width=44pt]{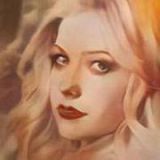}}&
\raisebox{17pt}{
\begin{tabular}{c}
Content
\end{tabular}}&
\begin{tabular}{r}
{\scriptsize (2) }
\end{tabular}&
\parbox[c]{\imgLen{}}{\includegraphics[width=44pt]{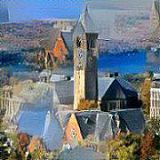}}&
\parbox[c]{\imgLen{}}{\includegraphics[width=44pt]{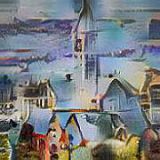}}&
\parbox[c]{\imgLen{}}{\includegraphics[width=44pt]{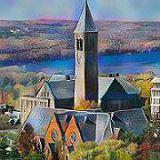}}&
\parbox[c]{\imgLen{}}{\includegraphics[width=44pt]{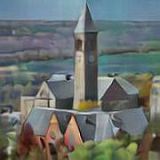}}\\

\parbox[c]{\imgLen{}}{}&
\parbox[r]{0.05pt}{} &
\begin{tabular}{c}
Gatys~\cite{Gatys2016ImageStyleTransferUsingCNN}
\end{tabular}&
\begin{tabular}{c}
AdaIN~\cite{Huang2017AdaIN}
\end{tabular}&
\begin{tabular}{c}
{\scriptsize Two-Stage}
\end{tabular}&
\begin{tabular}{c}
{\scriptsize End-to-End}
\end{tabular}&
\parbox[c]{\imgLen{}}{}&
\parbox[r]{0.05pt}{} &
\begin{tabular}{c}
Gatys~\cite{Gatys2016ImageStyleTransferUsingCNN}
\end{tabular}&
\begin{tabular}{c}
AdaIN~\cite{Huang2017AdaIN}
\end{tabular}&
\begin{tabular}{c}
{\scriptsize Two-Stage}
\end{tabular}&
\begin{tabular}{c}
{\scriptsize End-to-End}
\end{tabular}\\
\multicolumn{6}{c}{\normalsize (a)} & \multicolumn{6}{c}{\normalsize (b)} \\
\end{tabular}
\caption{Two sets of example results for reverse style transfer. The rows show (1) stylized images and (2) the de-stylized results. The corresponding content image is provided in the left of each set. Our models better reconstruct the overall structure of original content images.}
\label{fig:reslut_reverse_comapre}
\end{figure*}

\begin{figure*}[htbp]
\footnotesize
\centering
\begin{tabular}{p{33pt}p{3pt}p{33pt}p{33pt}p{33pt}p{43pt}| p{33pt}p{3pt}p{33pt}p{33pt}p{33pt}p{33pt}}
\parbox[c]{\imgLen{}}{\includegraphics[width=44pt]{figures/content_img/avril.jpg}}&
\begin{tabular}{r}
{\scriptsize (1) }
\end{tabular}&
\parbox[c]{\imgLen{}}{\includegraphics[width=44pt]{figures/Gatys_result/avril__impronte_d_artista.jpg}}&
\parbox[c]{\imgLen{}}{\includegraphics[width=44pt]{figures/AdaIN_result/avril__impronte_d_artista.jpg}}&
\parbox[c]{\imgLen{}}{\includegraphics[width=44pt]{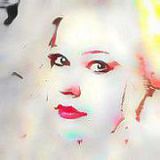}}&
\parbox[c]{\imgLen{}}{\includegraphics[width=44pt]{figures/e2e_result/avril__impronte_d_artista.jpg}}&
\parbox[c]{\imgLen{}}{\includegraphics[width=44pt]{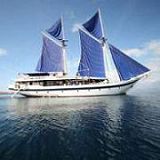}}&
\begin{tabular}{r}
{\scriptsize (1) }
\end{tabular}&
\parbox[c]{\imgLen{}}{\includegraphics[width=44pt]{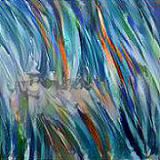}}&
\parbox[c]{\imgLen{}}{\includegraphics[width=44pt]{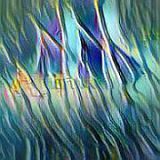}}&
\parbox[c]{\imgLen{}}{\includegraphics[width=44pt]{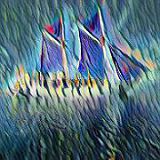}}&
\parbox[c]{\imgLen{}}{\includegraphics[width=44pt]{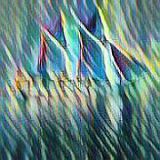}}\\
\raisebox{17pt}{
\begin{tabular}{c}
Content
\end{tabular}}&
\begin{tabular}{r}
{\scriptsize (2) }
\end{tabular}&
\parbox[c]{\imgLen{}}{\includegraphics[width=44pt]{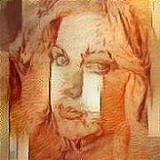}}&
\parbox[c]{\imgLen{}}{\includegraphics[width=44pt]{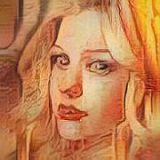}}&
\parbox[c]{\imgLen{}}{\includegraphics[width=44pt]{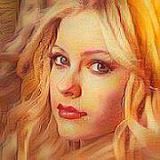}}&
\parbox[c]{\imgLen{}}{\includegraphics[width=44pt]{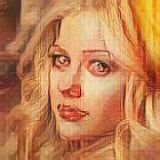}}&
\raisebox{17pt}{
\begin{tabular}{c}
Content
\end{tabular}}&
\begin{tabular}{r}
{\scriptsize (2) }
\end{tabular}&
\parbox[c]{\imgLen{}}{\includegraphics[width=44pt]{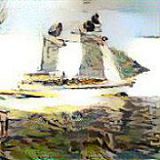}}&
\parbox[c]{\imgLen{}}{\includegraphics[width=44pt]{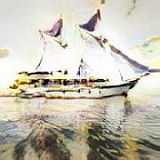}}&
\parbox[c]{\imgLen{}}{\includegraphics[width=44pt]{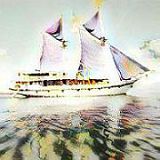}}&
\parbox[c]{\imgLen{}}{\includegraphics[width=44pt]{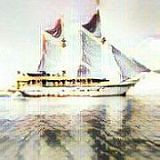}}\\
\parbox[c]{\imgLen{}}{\includegraphics[width=44pt]{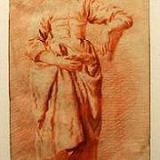}}&
\begin{tabular}{r}
{\scriptsize (3) }
\end{tabular}&
\parbox[c]{\imgLen{}}{\includegraphics[width=44pt]{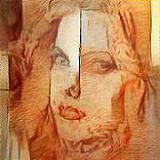}}&
\parbox[c]{\imgLen{}}{\includegraphics[width=44pt]{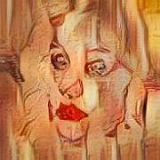}}&
\parbox[c]{\imgLen{}}{\includegraphics[width=44pt]{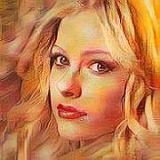}}&
\parbox[c]{\imgLen{}}{\includegraphics[width=44pt]{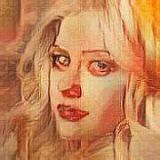}}&
\parbox[c]{\imgLen{}}{\includegraphics[width=44pt]{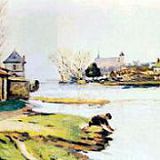}}&
\begin{tabular}{r}
{\scriptsize (3) }
\end{tabular}&
\parbox[c]{\imgLen{}}{\includegraphics[width=44pt]{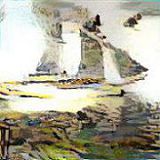}}&
\parbox[c]{\imgLen{}}{\includegraphics[width=44pt]{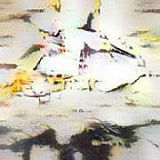}}&
\parbox[c]{\imgLen{}}{\includegraphics[width=44pt]{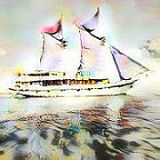}}&
\parbox[c]{\imgLen{}}{\includegraphics[width=44pt]{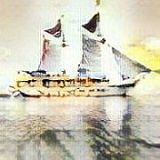}}\\
\begin{tabular}{c}
New Style
\end{tabular}&
\parbox[r]{0.05pt}{} &
\begin{tabular}{c}
Gatys~\cite{Gatys2016ImageStyleTransferUsingCNN}
\end{tabular}&
\begin{tabular}{c}
AdaIN~\cite{Huang2017AdaIN}
\end{tabular}&
\begin{tabular}{c}
{\scriptsize Two-Stage}
\end{tabular}&
\begin{tabular}{c}
{\scriptsize End-to-End}
\end{tabular}&
\begin{tabular}{c}
New Style
\end{tabular}&
\parbox[r]{0.05pt}{} &
\begin{tabular}{c}
Gatys~\cite{Gatys2016ImageStyleTransferUsingCNN}
\end{tabular}&
\begin{tabular}{c}
AdaIN~\cite{Huang2017AdaIN}
\end{tabular}&
\begin{tabular}{c}
{\scriptsize Two-Stage}
\end{tabular}&
\begin{tabular}{c}
{\scriptsize End-to-End}
\end{tabular}\\
\multicolumn{6}{c}{\normalsize (a)} & \multicolumn{6}{c}{\normalsize (b)} \\
\end{tabular}
\caption{Two sets of example results of serial style transfer. The rows from top to bottom sequentially show (1) stylized images based on the first style, (2) the expectation of new stylization on the content image, and (3) results of serial stylization produced by different methods. The corresponding content image and the new style going to be applied on it are provided in the left of each set. Our models have results better aligned w.r.t. the corresponding expectations.}
\label{fig:reslut_serial_comapre}
\end{figure*}

\subsection{Qualitative Evaluation}\label{sec:qual_results}

We compare our proposed models to the baselines from Gatys \etal~\cite{Gatys2016ImageStyleTransferUsingCNN} and AdaIN~\cite{Huang2017AdaIN} (more baselines, e.g. ~\cite{Li2017WCT,Ulyanov2016TextureNetworks}, in the appendix), based on the qualitative results for the tasks of regular, reverse and serial style transfer.
In particular, we apply two times of  stylization sequentially in the serial style transfer experiments, for both qualitative and quantitative evaluations (subsection~\ref{sec:qual_results} and~\ref{sec:quan_results} respectively). Please note that all the style images used in qualitative evaluation have never been seen during our training.

\subsubsection{Regular Style Transfer}
As the goal of our proposed models is not aiming to improve the quality of regular style transfer, we simply examine whether the stylization produced by our models is reasonable in comparison to the baselines. Figure.~\ref{fig:result_regular_compare} provides example results of regular style transfer generated by using different methods. We can see that although both our two-stage and end-to-end models have different stylized results w.r.t their base AdaIN approach, they retain comparable quality where the global structure of content image is maintained and the stylization is effective. 

\subsubsection{Reverse Style Transfer}
The goal of reverse style transfer is performing de-stylization on a stylized image, such that the content image can be reconstructed as close to its original appearance as possible. As the baselines, Gatys~\etal~\cite{Gatys2016ImageStyleTransferUsingCNN} and AdaIN~\cite{Huang2017AdaIN}, have no corresponding procedures for reverse style transfer, we thus utilize a na\"ive solution for them, where the stylization is applied to a given stylized image with having the original content image as source of target style. 
Please note here that this na\"ive solution of reverse style transfer for baselines needs the access to original content image, while our proposed models can perform  de-stylization solely with the given stylized image. 
Two sets of example results for the task of reverse style transfer are shown in Figure~\ref{fig:reslut_reverse_comapre}. From set (a), both baselines, especially AdaIN, fail to preserve the contour of the face. Although the results of our two-stage and end-to-end models have some mild color patches and slight color shift respectively, they both well reconstruct the overall structure of the content image. Similar observation also exists in set (b). The results of both our models are unaffected by the fuzzy patterns in stylized images, and have clear boundaries between objects, while the baselines could not discriminate the actual contours from the edges caused by stylization, which leads to the results with severe artifacts. These experimental results verify the capability of our models toward resolving the issue of reverse style transfer.

\subsubsection{Serial Style Transfer}
Serial style transfer attempts to transfer a stylized image into another different style, while keeping the result minimally affected by the previous stylization. Ideally, the result of serial style transfer is expected to be close to the one obtained by stylizing the original content image with the new style image.
Two sets of example results of serial style transfer are shown in Figure~\ref{fig:reslut_serial_comapre}. It is obvious that the results produced by our proposed method are more similar to their respective expectations than the ones from baselines which are deeply influenced by the previous stylization. Therefore our proposed models are successfully verified for their competence on dealing with serial style transfer.

\begin{table}[ht]
    \centering
    \small
\begin{tabular}{cc||cc|cc}
 & & 
 \begin{tabular}[c]{@{}c@{}}Gatys\\\cite{Gatys2016ImageStyleTransferUsingCNN}\end{tabular} &
 \begin{tabular}[c]{@{}c@{}}AdaIN\\\cite{Huang2017AdaIN}\end{tabular}
  & \begin{tabular}[c]{@{}c@{}}Two-\\Stage\end{tabular} & 
  \begin{tabular}[c]{@{}c@{}}End-to-\\End\end{tabular} \\ \hline\hline
\multicolumn{1}{l|}{\multirow{3}{*}{\begin{tabular}[c]{@{}c@{}}Reverse\\ Style\\ Transfer\end{tabular}}} & L2    & 4.4331 & 0.0368 & \textbf{0.0187} & 0.0193          \\
\cline{2-6}
\multicolumn{1}{l|}{}                                                                                    & SSIM  & 0.2033 & 0.3818 & 0.4796          & \textbf{0.5945} \\
\cline{2-6}
\multicolumn{1}{l|}{}                                                                                    & LPIPS & 0.3684 & 0.4614 & \textbf{0.3323} & 0.3802          \\
\hline
\multicolumn{1}{l|}{\multirow{3}{*}{\begin{tabular}[c]{@{}c@{}}Serial\\ Style\\ Trasfer\end{tabular}}}   & L2    & 7.5239 & 0.0213 & 0.0148          & \textbf{0.0104} \\
\cline{2-6}
\multicolumn{1}{l|}{} & SSIM  & 0.0472 & 0.5470 & 0.7143          & \textbf{0.8523} \\
\cline{2-6}
\multicolumn{1}{l|}{} & LPIPS & 0.4317 & 0.3637 & 0.2437          & \textbf{0.1487}
\end{tabular}
\caption{
    The averaged L2 distance, SSIM, and LPIPS~\cite{zhang2018perceptual} between the results and their corresponding expectations.}
    \label{tab:quantEval}
\end{table}

\subsection{Quantitative Evaluation}\label{sec:quan_results}
We conduct experiments to quantitatively evaluate the performance of our proposed models in both reverse and serial style transfer. A test set is built upon 1000 content images randomly sampled from the testing set of MS-COCO, with each of them transferred into 5 random styles that have never been used in the training phase. We perform reverse and serial style transfer with different models and compare the outputs with respect to their corresponding expectations. 
The averaged L2 distance, structural similarity (SSIM), and learned perceptual image patch similarity (LPIPS~\cite{zhang2018perceptual}) are used to measure the difference and the results are shown in Table~\ref{tab:quantEval}. 
Both our models perform better than the baselines. Particularly, our two-stage model performs the best for reverse style transfer while the end-to-end model does so for serial style transfer. We believe that our two stage model benefits from its larger amount of encrypted information and the design of identity mapping, leading to the better result in reverse style transfer, and the end-to-end model shows its advantage in having less information to hide, making it more robust to the propagated error caused by serial style transfer.

\subsection{Ablation Study}\label{sec:ablation_study}
Here we perform ablation studies to verify the benefits of some design choices in our proposed models. Due to page limit, please refer to our appendix for more studies.\\

\begin{figure}[t]
\centering

\begin{tabular}{ccc}
\parbox[c]{\shorterLength{}}{\includegraphics[width=0.2\columnwidth]{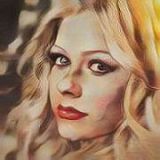}}&
\parbox[c]{\shorterLength{}}{\includegraphics[width=0.2\columnwidth]{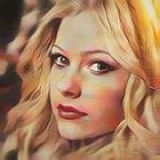}}&
\parbox[c]{\shorterLength{}}{\includegraphics[width=0.2\columnwidth]{figures/content_img/avril.jpg}}\\
\parbox[c]{\shorterLength{}}{\includegraphics[width=0.2\columnwidth]{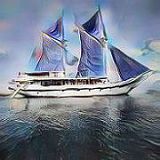}}&
\parbox[c]{\shorterLength{}}{\includegraphics[width=0.2\columnwidth]{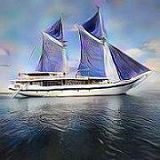}}&
\parbox[c]{\shorterLength{}}{\includegraphics[width=0.2\columnwidth]{figures/content_img/sailboat.jpg}}\\
AdaIN & Identity mapping & Ground truth \\
\end{tabular}
\caption{Comparison between results of using the AdaIN decoder trained w/o and w/ identity mapping in the task of reverse style transfer. It shows the AdaIN decoder trained w/ identity mapping generates results with less artifacts.}
\label{fig:comapre_ideneity}
\end{figure}

\begin{figure}[t]
\begin{tabular}{cccc}
\parbox[c]{\theLength{}}{\includegraphics[width=0.195\columnwidth]{figures/content_img/brad_pitt.jpg}}&
\parbox[c]{\theLength{}}{\includegraphics[width=0.195\columnwidth]{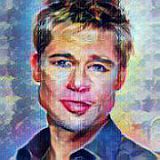}}&
\parbox[c]{\theLength{}}{\includegraphics[width=0.195\columnwidth]{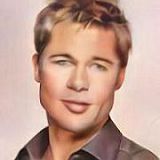}}&
\parbox[c]{\theLength{}}{\includegraphics[width=0.195\columnwidth]{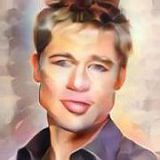}}\\
\parbox[c]{\theLength{}}{\includegraphics[width=0.195\columnwidth]{figures/content_img/cornell.jpg}}&
\parbox[c]{\theLength{}}{\includegraphics[width=0.195\columnwidth]{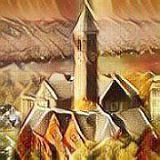}}&
\parbox[c]{\theLength{}}{\includegraphics[width=0.195\columnwidth]{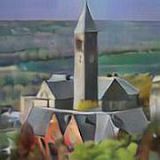}}&
\parbox[c]{\theLength{}}{\includegraphics[width=0.195\columnwidth]{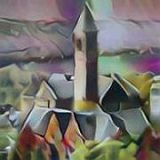}}\\
Content $I_c$ & Stylized $I_{st}$ & $E_{inv}$ & $E_{VGG}$\\
\end{tabular}
\caption{
Comparison between using \(E_{inv}\) and \(E_{VGG}\) to extract the target feature $v_t$ in reverse style transfer. The ones of using $E_{inv}$ show less influence from the stylization.}
\label{fig:ablation_Einv}
\end{figure}

\textbf{Identity mapping of two-stage model}
As described in Sec.~\ref{sec:2stage_st}, for the decoder $D_{AdaIN}$, we have an additional objective based on identity mapping. From the example results provided in Figure~\ref{fig:comapre_ideneity}, we can see the ones produced by our $D_{AdaIN}$ have less artifacts, which clearly demonstrate the benefits to the task of reverse style transfer brought by using identity mapping in our proposed model, in comparison to the decoder used in the typical style transfer method.\\
\textbf{Using $E_{inv}$ to recover $v_t$ from $I_{st}$ in end-to-end model}
There is a potential argument that we could replace $E_{inv}$ with $E_{VGG}$ due to the similarity between $\mathcal{L}_{inv}$ and $\mathcal{L}_{content}$ in our end-to-end model (please note that $E_{VGG}$ is pretrained and kept fixed). Hence we perform experiments accordingly in the task of reverse style transfer, and observe that the results of using $E_{inv}$ preserve the overall content structure better, while the ones of using $E_{VGG}$ tend to have severe interference from stylization as shown in Figure~\ref{fig:ablation_Einv}. The benefit of having $E_{inv}$ in our model is thus verified. Please note that more results and videos are available in
the appendix. All the source code and datasets (or
trained models) will be made available to the public.

%% file: conclusion.tex
\section{Conclusion}\label{sec:conclude}

In this paper, we introduce the issues and artifacts that are inevitably introduced by typical style transfer methods in the scenarios of serial and reverse style transfer. We successfully address these problems by proposing a two-stage and an end-to-end approach while retaining the image quality of stylized output comparable to the state-of-the-art style transfer method simultaneously. Our methods are novel on uniquely integrating the steganography technique into style transfer for preserving the important characteristic of content features extracted from input photo, and the extensive experiments clearly verify the capability of our networks. \\
~\\
\noindent\textbf{Acknowledgements}~
This project is supported by MediaTek Inc., MOST-108-2636-E-009-001, MOST-108-2634-F-009 -007, and MOST-108-2634-F-009-013.
We are grateful to the National Center for Highperformance Computing for computer time and facilities. 

%% file: appendix.tex
\appendix
{\LARGE \textbf{Appendix}}

\input{appendix_result.tex}
\input{appendix_ablation_2_stage.tex}

\input{appendix_ablation_e2e.tex}

\input{appendix_baseline.tex}
\input{appendix_limitations.tex}
\input{appendix_implementation.tex}
\input{full_page_tables.tex}

%% file: appendix_result.tex
    \begin{table*}[bht]
    \centering
    \begin{tabular}{c|ccc|ccc}
    \multirow{2}{*}{} & \multicolumn{3}{c}{ Reverse Style Transfer} & \multicolumn{3}{c}{Serial Style Transfer}  \\
    \cline{2-4} \cline{5-7} 
     &  L2 &  SSIM & LPIPS & L2 & SSIM & LPIPS\\
    \hline
    \hline
    Gatys \etal \cite{Gatys2016ImageStyleTransferUsingCNN}&
    4.4331 & 0.2033 & 0.3684 &7.5239 & 0.0472 & 0.4317\\
    AdaIN \cite{Huang2017AdaIN} &
    0.0368 &0.3818 & 0.4614 &0.0213 &0.5477 & 0.3637\\
        WCT \cite{Li2017WCT} &
    0.0597 &0.3042 & 0.5534 &0.0568 &0.3318 & 0.5048 \\\hline
    Extended baseline (AdaIN w/ cycle consistency)&
    0.0502 & 0.2931 & 0.5809 & 0.0273 & 0.4140 & 0.4314\\
\hline
    Our two-stage &
    \textbf{0.0187} &0.4796 & \textbf{0.3323} &0.0148 &0.7143& 0.2437\\
    Our end-to-end &
    0.0193 &\textbf{0.5945} & 0.3802 &\textbf{0.0104} &\textbf{0.8523} & \textbf{0.1487}\\
    \end{tabular}
    \caption{The average L2 distance, structural similarity (SSIM) and learned perceptual image patch similarity (LPIPS~\cite{zhang2018perceptual}) between the results produced by different models and their corresponding expectations. Regarding extended baseline (AdaIN with cycle consistency), please refer to the Section~\ref{sec:baseline} for more detailed description.}
    \label{tab:MorequantEval}
    \end{table*}
     
\section{More Results}
\subsection{Regular, Reverse and Serial Style Transfer}
\subsubsection{Qualitative Evaluation}
First, we provide three more sets of results in Figure~\ref{fig:more_results}, demonstrating the differences between the results of regular, reverse, and serial style transfer performed by different methods. 
Moreover, we provide in the Figure~\ref{fig:more_test_results} and~\ref{fig:more_test_results_1} more qualitative results, based on diverse sets of content and style images from MS-COCO~\cite{lin2014microsoft} and WikiArt~\cite{wiki_art} datasets respectively. In which these results show that our proposed methods are working fine to perform regular, reverse, and serial style transfer on various images.

\subsubsection{Quantitative Evaluation}
As mentioned in the Section 4.3 of our main paper, here we provide more quantitative evaluations in Table~\ref{tab:MorequantEval}, based on L2 distance, structural similarity (SSIM), and LPIPS~\cite{zhang2018perceptual}. Both our methods in the tasks of reverse and serial stylization perform better than the baselines in terms of different metrics. Please note that although Gatys \etal~\cite{Gatys2016ImageStyleTransferUsingCNN} can obtain also good performance for the task of reverse style transfer in terms of LPIPS metric (based on the similarity in semantic feature representation), it needs to use the original image as the style reference to perform the reverse style transfer, which is actually impractical.

\subsection{Serial Style Transfer for Multiple Times}
To further demonstrate the ability of preserving the content information of our models, we perform serial style transfer on an image for multiple times. There are three sets of results in Figure~\ref{fig:result_multi_serial} for comparing the results generated by different methods. It can be seen that Gatys \etal~\cite{Gatys2016ImageStyleTransferUsingCNN} and AdaIN~\cite{Huang2017AdaIN} fail to distinguish the contour of the content objects from the edges caused by the stylization, thus the results deviate further from the original content when serial style transfer is applied. As for our two-stage and end-to-end model, the content is still nicely preserved even in the final results after a series of style transfer. It clearly indicates that our models provide better solutions to the issue of serial style transfer.

%% file: appendix_ablation_2_stage.tex
\section{More Ablation Study}\label{sec:appendix_ablation_study}

\subsection{Two-Stage Model}

\begin{table*}[ht]
    \centering
    \begin{tabular}{c|ccc|ccc}
    \multirow{2}{*}{} & \multicolumn{3}{c}{ Reverse Style Transfer} & \multicolumn{3}{c}{Serial Style Transfer}  \\
    \cline{2-4} \cline{5-7} 
     &  L2 &  SSIM & LPIPS & L2 & SSIM & LPIPS\\
    \hline
    \hline
    Our two-stage (w/ identity mapping)&
    \textbf{0.0187} & \textbf{0.4796} & \textbf{0.3323} & \textbf{0.0148} & \textbf{0.7143} & \textbf{0.2437}\\
    Our two-stage (w/o identity mapping) &
    0.0226 & 0.4596 & 0.3637 & 0.0152 & 0.6990 & 0.2560\\
    Our two-stage (w/ adversarial learning)  &
    0.0271 & 0.4292 & 0.3878 & 0.0168 & 0.5946 & 0.3236\\
    \end{tabular}
    \caption{
    The average L2 distance, structural similarity (SSIM) and learned perceptual image patch similarity (LPIPS~\cite{zhang2018perceptual}) between the expected results and the ones which are obtained by our two-stage model and its variants of having identity mapping AdaIN decoder or adversarial learning. 
    }
    \label{tab:quantEval_ablation_2st}
\end{table*}

\begin{figure*}[t]
    \centering
    \includegraphics[width=495pt]{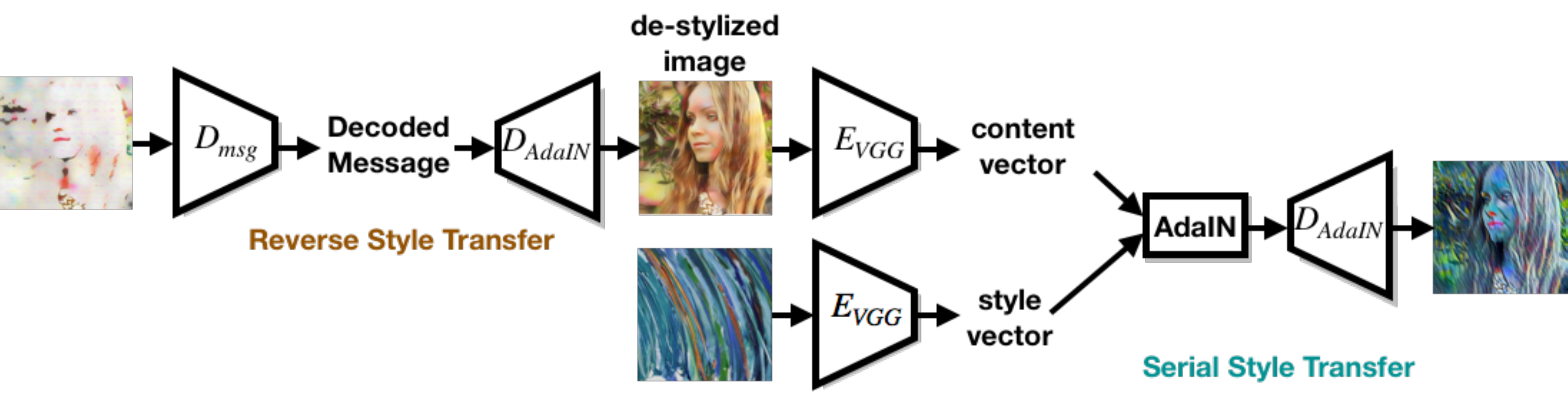}
    \caption{Illustrations of how to apply our two-stage model in the task of serial style transfer with de-stylized image.}
    \label{fig:sst_destylized}
\end{figure*}
 
\subsubsection{Quantitative Evaluation of Identity Mapping}
We evaluate the effect of having \textit{identity mapping} (Section 3.1.1 in the main paper) in our proposed two-stage model based on the average L2 distance, structural similarity (SSIM), and learned perceptual image patch similarity (LPIPS~\cite{zhang2018perceptual}).
The results are provided in Table \ref{tab:quantEval_ablation_2st}. It clearly shows that adding identity mapping in the training of AdaIN decoder $D_{AdaIN}$ enhances the performance of reverse and serial style transfer.
 
\subsubsection{Training with and without Adversarial Learning}
As mentioned in Section 3.1.3 of the main paper, the architectures of our message encoder $E_{msg}$ and decoder $D_{msg}$ in the steganography stage are the same as the ones used in HiDDeN~\cite{HiDDeN}, while HiDDeN~\cite{HiDDeN} additionally utilizes adversarial learning to improve the performance of encoding. Here we experiment to train our steganography stage with adversarial learning as well, where two losses $\left\{ \mathcal{L}_{discriminator}, \mathcal{L}_{generator}\right \}$ are added to our object function as follows. 
 
\begin{equation}
\begin{aligned}
\mathcal{L}_{\text{discriminator}} =& \mathbb{E}\left[\left(Dis\left(I_t\right)-\mathbb{E}\left(Dis\left(I_e\right)\right)-1\right)^2\right] + \\&\mathbb{E}\left[\left(Dis\left(I_e\right)-\mathbb{E}\left(Dis\left(I_t\right)\right)+1\right)^2\right]
\label{eq:L_discriminator}
\end{aligned}
\end{equation}

\begin{equation}
\begin{aligned}
\mathcal{L}_{\text{generator}} =& \mathbb{E}\left[\left(Dis\left(I_e\right)-\mathbb{E}\left(Dis\left(I_t\right)\right)-1\right)^2\right] + \\&\mathbb{E}\left[\left(Dis\left(I_t\right)-\mathbb{E}\left(Dis\left(I_e\right)\right)+1\right)^2\right]
\label{eq:L_generator}
\end{aligned}
\end{equation}
where $Dis$ denotes the discriminator used in adversarial learning. Here in our experiment, the architecture of the discriminator is identical to the one used in HiDDeN~\cite{HiDDeN}, and we adopt the optimization procedure proposed in~\cite{jolicoeur2018relativistic} for adversarial learning.
 
Afterward, we perform qualitative and quantitative evaluations on the results, as shown in Figure~\ref{fig:compare_adversarial} and Table~\ref{tab:quantEval_ablation_2st} respectively. We observe that adding adversarial learning does not enhance the quantitative performance. Similarly, we remark that the results are visually similar according to the qualitative examples as shown in Figure~\ref{fig:compare_adversarial}.

\subsubsection{Serial Style Transfer with De-Stylized Image}
As mentioned in the main paper (cf. Section 3.1.3), we stylize the image generated from the decoded message to perform serial style transfer. However, we can also resolve the issue of serial style transfer in a different way. Figure \ref{fig:sst_destylized} shows that we can implement serial style transfer by stylizing the de-stylized image from the result of reserve style transfer. For comparison, we qualitatively evaluate the results generated with the de-stylized image and the decoded message. Figure \ref{fig:comapre_msg_img} shows that the results of these two methods are nearly identical. Since the model using decoded message (as in the main paper) is simpler than the other, we choose to adopt it in our proposed method. 
The quantitative evaluation is also provided in the Table~\ref{tab:2st_ablation_destyle_quant}, based on the metrics of average L2 distance, structural similarity (SSIM) and learned perceptual image patch similarity (LPIPS~\cite{zhang2018perceptual}). We can see that our model of using decoded message performs better than the one of using de-stylized image, in which this observation thus verifies our design choice.

\newcommand\theLengthMsgImg{5em}
 \begin{figure}[htbp]
 \centering
 \begin{tabular}{cccc}
 \parbox[c]{\theLengthMsgImg{}}{\includegraphics[width=0.195\columnwidth]{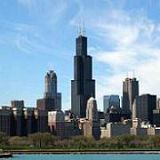}}&
 \parbox[c]{\theLengthMsgImg{}}{\includegraphics[width=0.195\columnwidth]{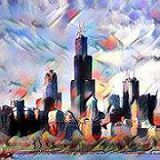}}&
 \parbox[c]{\theLengthMsgImg{}}{\includegraphics[width=0.195\columnwidth]{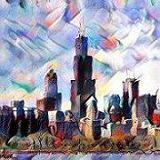}}&
 \parbox[c]{\theLengthMsgImg{}}{\includegraphics[width=0.195\columnwidth]{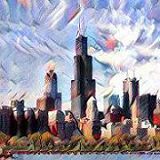}}\\
 \parbox[c]{\theLengthMsgImg{}}{\includegraphics[width=0.195\columnwidth]{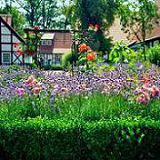}}&
 \parbox[c]{\theLengthMsgImg{}}{\includegraphics[width=0.195\columnwidth]{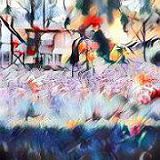}}&
 \parbox[c]{\theLengthMsgImg{}}{\includegraphics[width=0.195\columnwidth]{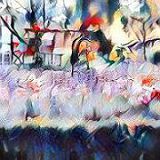}}&
 \parbox[c]{\theLengthMsgImg{}}{\includegraphics[width=0.195\columnwidth]{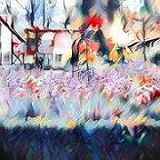}}\\
 Content & Message & De-stylized & Ground Truth \\
 \end{tabular}
 \caption{Comparison between the results of serial style transfer generated with decoded messages and the de-stylized images.}
 \label{fig:comapre_msg_img}
 \end{figure}
 
 \begin{table}[ht]
     \centering
     \begin{tabular}{c|ccc}
          & L2 & SSIM & LPIPS \\
          \hline
          w/ de-stylized image & 0.02558 & 0.48694 & 0.40362\\
          w/ decoded message & \textbf{0.01480} & \textbf{0.71430} & \textbf{0.24370}\\
     \end{tabular}
     \caption{
     The average L2 distance, structural similarity (SSIM) and learned perceptual image patch similarity (LPIPS~\cite{zhang2018perceptual}) between expected results and the ones which are produced by our two-stage model with performing serial style transfer w/ de-stylized image or w/ decoded message.}
     \label{tab:2st_ablation_destyle_quant}
 \end{table}

%% file: appendix_ablation_e2e.tex
\begin{table*}[t]
    \centering
    \begin{tabular}{c|ccc|ccc}
    \multirow{2}{*}{} & \multicolumn{3}{c}{ 
    \begin{tabular}[c]{@{}c@{}} Reverse \\ Style Transfer  \end{tabular}
    } & \multicolumn{3}{c}{
   \begin{tabular}[c]{@{}c@{}} Serial \\ Style Transfer  \end{tabular}
    }  \\
    \cline{2-4} \cline{5-7} 
     &  L2 &  SSIM & LPIPS & L2 & SSIM & LPIPS\\
    \hline
    \hline
    $E_{inv}$ &
    \textbf{0.0193} &\textbf{0.5945} & \textbf{0.3802} &\textbf{0.0104} &\textbf{0.8523} & \textbf{0.1487} \\
    $E_{VGG}$ &
    0.0241 & 0.5190 & 0.4727 & 0.0149 & 0.7525 & 0.2362\\
    \end{tabular}
    \caption{
    The average L2 distance, structural similarity (SSIM) and learned perceptual image patch similarity (LPIPS~\cite{zhang2018perceptual}) between expected results and the ones which are obtained by our end-to-end model of using $E_{inv}$ or $E_{VGG}$. 
    }
    \label{tab:e2e_E_inv_quant}
\end{table*}

\begin{figure*}[ht]
\centering
\begin{tabular}{cccccccccccc}
&&&&&&&
\parbox[c]{4.2em}{\includegraphics[width=0.195\columnwidth]{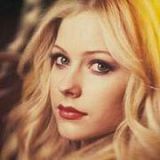}}&
\parbox[c]{4.2em}{\includegraphics[width=0.195\columnwidth]{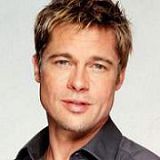}}&
\parbox[c]{4.2em}{\includegraphics[width=0.195\columnwidth]{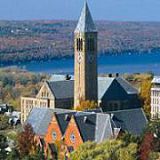}}&
\parbox[c]{4.2em}{\includegraphics[width=0.195\columnwidth]{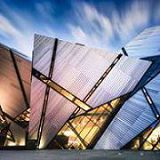}}\\
&
\parbox[c]{4.2em}{\includegraphics[width=0.195\columnwidth]{appendix_figures/content_img/avril.jpg}}&
\parbox[c]{4.2em}{\includegraphics[width=0.195\columnwidth]{appendix_figures/content_img/chicago.jpg}}&
\parbox[c]{4.2em}{\includegraphics[width=0.195\columnwidth]{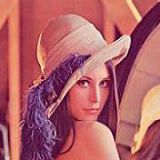}}&
\parbox[c]{4.2em}{\includegraphics[width=0.195\columnwidth]{appendix_figures/content_img/modern.jpg}}&
\parbox[c]{0.3em}&
\parbox[c]{0.6em}&
\parbox[c]{4.2em}{\includegraphics[width=0.195\columnwidth]{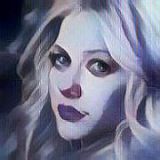}}&
\parbox[c]{4.2em}{\includegraphics[width=0.195\columnwidth]{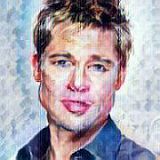}}&
\parbox[c]{4.2em}{\includegraphics[width=0.195\columnwidth]{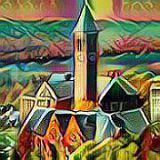}}&
\parbox[c]{4.2em}{\includegraphics[width=0.195\columnwidth]{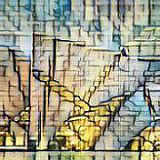}}\\

\parbox[c]{0.6em}{(1)}&
\parbox[c]{4.2em}{\includegraphics[width=0.195\columnwidth]{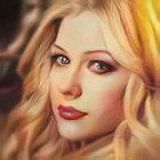}}&
\parbox[c]{4.2em}{\includegraphics[width=0.195\columnwidth]{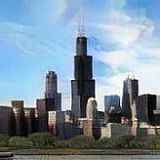}}&
\parbox[c]{4.2em}{\includegraphics[width=0.195\columnwidth]{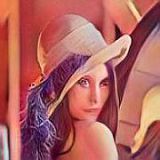}}&
\parbox[c]{4.2em}{\includegraphics[width=0.195\columnwidth]{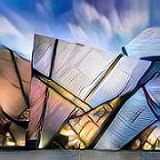}}&
\parbox[c]{0.3em}&
\parbox[c]{0.6em}{(1)}&
\parbox[c]{4.2em}{\includegraphics[width=0.195\columnwidth]{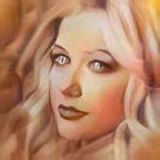}}&
\parbox[c]{4.2em}{\includegraphics[width=0.195\columnwidth]{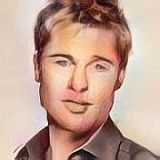}}&
\parbox[c]{4.2em}{\includegraphics[width=0.195\columnwidth]{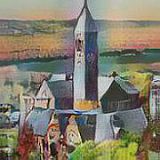}}&
\parbox[c]{4.2em}{\includegraphics[width=0.195\columnwidth]{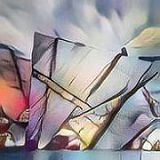}}\\
\parbox[c]{0.6em}{(2)}&
\parbox[c]{4.2em}{\includegraphics[width=0.195\columnwidth]{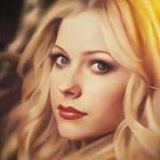}}&
\parbox[c]{4.2em}{\includegraphics[width=0.195\columnwidth]{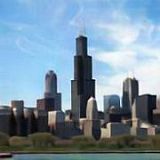}}&
\parbox[c]{4.2em}{\includegraphics[width=0.195\columnwidth]{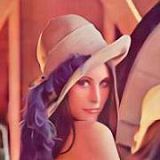}}&
\parbox[c]{4.2em}{\includegraphics[width=0.195\columnwidth]{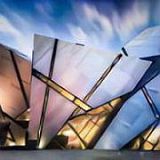}}&
\parbox[c]{0.3em}&
\parbox[c]{0.6em}{(2)}&
\parbox[c]{4.2em}{\includegraphics[width=0.195\columnwidth]{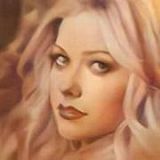}}&
\parbox[c]{4.2em}{\includegraphics[width=0.195\columnwidth]{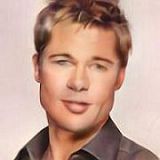}}&
\parbox[c]{4.2em}{\includegraphics[width=0.195\columnwidth]{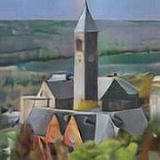}}&
\parbox[c]{4.2em}{\includegraphics[width=0.195\columnwidth]{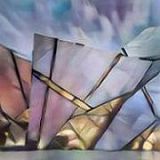}}\\
&\multicolumn{4}{c}{(a)}&&&\multicolumn{4}{c}{(b)}\\
\end{tabular}
\caption{Comparison between the images decoded from the same feature vectors by (1) the AdaIN decoder $D_{AdaIN}$ in two-stage model and (2) the plain image decoder $D_{plain}$ in end-to-end model. In set (a), the features given to the decoders are the content features extracted from the images in the top row by pre-trained VGG19~\cite{VGG}, which is $E_{VGG}(I_c)$. As for set (b), the given feature vectors are the ones derived from the stylized images (the second row) with the end-to-end model, i.e $\hat{v}_c$.}
\label{fig:ablation_e2e_decoders}
\end{figure*}

\subsection{End-to-End Model}

\subsubsection{Quantitative evaluation of using $E_{inv}$ to recover $v_t$ from $I_{st}$ in end-to-end model}

We evaluate the effect of having $E_{inv}$ (please refer to the Section 4.4 in the main paper) in our proposed end-to-end model based on the metrics of average L2 distance, structural similarity (SSIM), and learned perceptual image patch similarity (LPIPS~\cite{zhang2018perceptual}). The results are provided in Table \ref{tab:e2e_E_inv_quant}. It clearly shows that using $E_{inv}$ instead of $E_{VGG}$ enhances the performance of reverse and serial style transfer, which thus verifies our design choice of having $E_{inv}$ in our end-to-end model.

\subsubsection{Decoding with Plain Image Decoder or AdaIN Decoder for Reverse Style Transfer}
It is mentioned in Section 3.2 of the main paper that the training of a plain image decoder $D_{plain}$ in the end-to-end model shares the same idea with the identity mapping, which is used in learning AdaIN decoder $D_{AdaIN}$ of the two-stage model. However, although they both are trained to reconstruct the image $I_c$ with its own feature $E_{VGG}(I_c)$, these two decoders accentuate different parts of the given feature during the reconstruction. The AdaIN decoder is trained to decode the results of regular and reverse style transfer simultaneously, but with an emphasis on the stylization, considering that identity mapping is only activated occasionally during the training. It is optimized toward both content and style features based on the perceptual loss in order to evaluate the effect of the stylization. As for the plain image decoder, it is solely trained for reconstructing the image with the given content feature, and optimized with the L2 distance to the original image. Such distinction brings differences to the images decoded from the same feature by these two decoders, as shown in Figure~\ref{fig:ablation_e2e_decoders} and Table~\ref{tab:e2e_ablation_plain_quant}.

Comparing to the results generated by the plain image decoder, the images decoded by the AdaIN decoder have sharper edges and more fine-grained details, but sometimes the straight lines are distorted and the contours of the objects are not in the same place as they are in the original image, harming the consistency of the overall content structure. Examples can be found in Figure~\ref{fig:ablation_e2e_decoders}, especially on the boundaries of the buildings. 
The quantitative evaluation provided in Table \ref{tab:e2e_ablation_plain_quant} also shows that using plain image decoder could provide better performance than adopting AdaIN decoder in terms of different metrics.
The benefit of introducing the plain image decoder for reverse style transfer of end-to-end model is therefore verified. 
 \begin{table}[ht]
     \centering
     \begin{tabular}{c|ccc}
          & L2 & SSIM & LPIPS \\
          \hline
          Plain image decoder & \textbf{0.0193} & \textbf{0.5945} & \textbf{0.3802}\\
          AdaIN decoder & 0.0349 & 0.4261 & 0.4141\\
     \end{tabular}
     \caption{
     The average L2 distance, structural similarity (SSIM) and learned perceptual image patch similarity (LPIPS~\cite{zhang2018perceptual}) between expected results and the ones which are obtained by our end-to-end model of using plain image decoder $D_{plain}$ or VGG decoder for reverse style transfer. 
     }
     \label{tab:e2e_ablation_plain_quant}
 \end{table}

%% file: appendix_baseline.tex
\begin{figure}[ht]
    \centering
    \includegraphics[width=\linewidth]{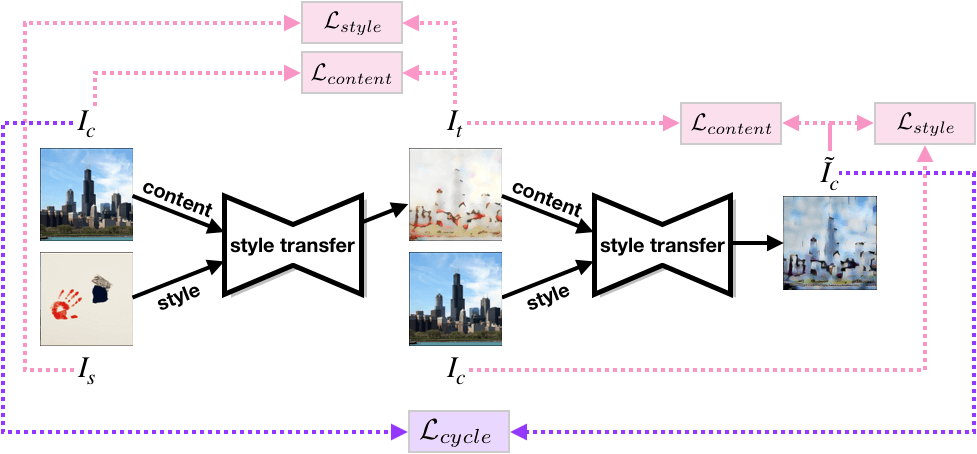}
    \caption{Illustration of the framework and training objectives for the extended baseline for reverse style transfer, which is based on a typical style transfer approach (i.e. AdaIN) and the cycle consistency objective $\mathcal{L}_{cycle}$.}
    \label{fig:baseline_method}
\end{figure}

\section{Extended Baseline}\label{sec:baseline}
\noindent \textbf{Typical Style Transfer Approach Extended with Cycle Consistency.}
As mentioned in the Section 4.2.2. and Figure.2 of our main paper, the na\"ive baselines built upon the typical style transfer approaches (i.e. Gatys \etal~\cite{Gatys2016ImageStyleTransferUsingCNN} and AdaIN~\cite{Huang2017AdaIN}) are not able to resolve the task of reverse style transfer, which is analogous to perform de-stylization on a stylized image back to its original photo. For further exploration of the capacity of the na\"ive baselines for reverse style transfer, here we provide another extended baseline for comparison. 

The framework of this extended baseline is illustrated in Figure~\ref{fig:baseline_method}, where the AdaIN style transfer component is composed of a pre-trained VGG19 encoder and a decoder $D_{AdaIN}$.  First, given a content photo $I_c$ and a style image $I_s$, $D_{AdaIN}$ is trained for making the stylized image $I_t$ to have similar content and style as $I_c$ and $I_s$ respectively, where the content loss $\mathcal{L}_{content}(I_c, I_t)$ and the style loss $\mathcal{L}_{style}(I_s, I_t)$ are used (please refer to Equation.3 and 4 in the main paper). Second, $I_t$ and $I_c$ are taken as the source of content and style respectively to produce a de-stylized output $\tilde{I_c}$, where $D_{AdaIN}$ is now trained to minimize $\mathcal{L}_{content}(I_t, \tilde{I_c})$ and $\mathcal{L}_{style}(I_c, \tilde{I_c})$. Last, the cycle consistency objective $\mathcal{L}_{cycle} = \left \| \tilde{I_c} - I_c\right\|$ is introduced for updating $D_{AdaIN}$ in order to encourage $\tilde{I_c}$ and $I_c$ to be identical, i.e. reverse style transfer or de-stylization.

\begin{figure}[ht]
\centering
\begin{tabular}{ccc}
\parbox[c]{\theLength{}}{\includegraphics[width=0.2\columnwidth]{appendix_figures/content_img/brad_pitt.jpg}}&
\parbox[c]{\theLength{}}{\includegraphics[width=0.2\columnwidth]{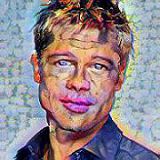}}&
\parbox[c]{\theLength{}}{\includegraphics[width=0.2\columnwidth]{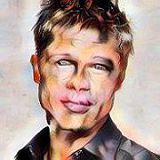}}\\

\parbox[c]{\theLength{}}{\includegraphics[width=0.2\columnwidth]{appendix_figures/content_img/cornell.jpg}}&
\parbox[c]{\theLength{}}{\includegraphics[width=0.2\columnwidth]{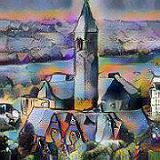}}&
\parbox[c]{\theLength{}}{\includegraphics[width=0.2\columnwidth]{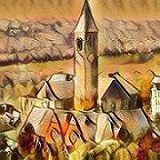}}\\

\parbox[c]{\theLength{}}{\includegraphics[width=0.2\columnwidth]{appendix_figures/content_img/avril.jpg}}&
\parbox[c]{\theLength{}}{\includegraphics[width=0.2\columnwidth]{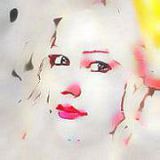}}&
\parbox[c]{\theLength{}}{\includegraphics[width=0.2\columnwidth]{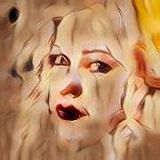}}\\

\parbox[c]{\theLength{}}{\includegraphics[width=0.2\columnwidth]{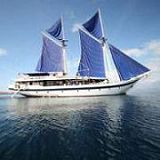}}&
\parbox[c]{\theLength{}}{\includegraphics[width=0.2\columnwidth]{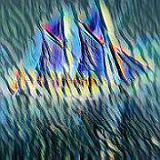}}&
\parbox[c]{\theLength{}}{\includegraphics[width=0.2\columnwidth]{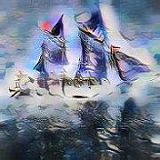}}\\

Content & Stylized & De-stylized \\
\end{tabular}
\caption{The results produced by the extended baseline of reverse style transfer which is trained with cycle consistency loss.}
\label{fig:result_improved_baseline}
\end{figure}

As shown in Figure \ref{fig:result_improved_baseline}, even if the extended baseline is trained with the cycle consistency objective, it is still not able to resolve the task of reverse style transfer. 
The quantitative results provided in the Table \ref{tab:MorequantEval} also indicate the inferior performance of this extended baseline. 
These results also demonstrate that the content information is lost during the procedure of the typical style transfer and can not be easily recovered, which further emphasize the contribution and the novelty of our proposed models based on the steganography idea. Please also note that the extended baseline needs to take the original content photo $I_c$ as the source of style for performing de-stylization, while our proposed models are self-contained without any additional input.

\section{Replacing AdaIN with Other Stylization Methods for Two-Stage Model}
To verify the adaptability of our two-stage model, we replace AdaIN, which is originally adopted in the style transfer stage, with WCT~\cite{Li2017WCT} and instance  normalization~\cite{Ulyanov2017InstanceNormalization}, and compare their results to the ones of the original implementation. Denote the selected style transfer method (e.g.WCT~\cite{Li2017WCT}) by $f$, we can get the stylized image $I_t = f(I_c, I_s)$ in the style transfer stage. Meanwhile, the content feature $v_c = D_{VGG}(I_c)$ remains to be \texttt{relu4\_1} extracted by VGG19 from the content image, and is encrypted into the stylized image $I_t$ by $I_e = E_{message}(I_t, v_c)$ later in the steganography stage.

As the content feature encrypted in $I_e$ is retrievable by $v'_c = D_{msg}(I_e)$ just like the original implementation, reverse style transfer can be intuitively done by $I'_c = D_{AdaIN}(v'_c)$. When it comes to serial style transfer, we simply need to further stylize the reconstructed image $I'_c$ with another style $I'_s$ by computing $I'_t = f(I'_c, I'_s)$. As the results shown in Figure \ref{fig:two-stage_other_method}, our two-stage model still performs well when adapted to WCT~\cite{Li2017WCT} and instance normalization~\cite{Ulyanov2017InstanceNormalization}. Their results are closer to the corresponding expectations, have less artifacts, and preserve more content structure and detail than the ones of na\"ive approaches, as the original implementation with AdaIN does. Please note that all these replacements are done without the need of any additional training. The encoders/decoders trained with the original implementation can be directly inherited without further modification.

%% file: appendix_limitations.tex
\section{Limitations}

The main limitation of our proposed methods, which stem from the idea of steganography, is being unavoidable to have errors in the decrypted message through the procedure of encryption and decryption. In our two-stage model, since it needs to hide the whole content feature of the original image into its stylized output, the errors in the decrypted message would cause inconsistent color patches in the results of reverse style transfer. For instance in Figure~\ref{fig:more_results}, as can be seen from the reverse style transfer results of the sailboat image produced by our two-stage model, there are different color patches in the sky which ideally should be homogeneous. While for our end-to-end model, it aims to encrypt the statistic (i.e., mean and variance) of the content feature of the original image into the stylized output, the errors in the decrypted message now lead to the color shift issue when performing reverse style transfer, which is also observable in the Figure~\ref{fig:more_results}. We would seek for other network architecture designs or training techniques (e.g. add random noise during network training, as used in~\cite{HiDDeN}) in order to have better robustness of our models against the errors caused by encryption and decryption.

%% file: appendix_implementation.tex
\section{Implementation Details}
Here we provide some implementation details of our two-stage and end-to-end model. We use PyTorch~\cite{PyTorch} 0.4.1 as our environment of developing deep learning framework.

All the source code and trained models will be publicly available for reproductivity once the paper is accepted.

\subsection{Two-Stage Model}

\noindent \textbf{Architectures.}
$D_{AdaIN}$ has the same architecture as the decoder used in the original implementation of AdaIN~\cite{Huang2017AdaIN}. It consists of 3 nearest up-sampling layers, 9 convolutional layers with the kernels of size $3\times3$, and ReLU activations after each conv-layer except the last one. Our $E_{msg}$ and $D_{msg}$ also inherit the architecture of the encoder and decoder in the implementation of HiDDeN~\cite{HiDDeN}.
$E_{msg}$ has 4 convolution blocks. Each convolution block includes a convolutional layer with a $3\times3$ kernel, a batch normalization layer and a ReLU activation (except the last block). 
The message to encrypt is first reshaped, then concatenated to the output of the first convolution block.
$D_{msg}$ has 8 convolution blocks. Each convolution block includes a convolutional layer with a $3\times3$ kernel, a batch normalization layer and a ReLU activation (except the first and the last block). 
The dimension of the decrypted message is recovered by adaptive average pooling and reshaping after the final block.

\noindent \textbf{Hyperparameters.}
The learning rate used in our model training is $10^{-4}$. We adopt Adam optimizer~\cite{kingma2014adam} with hyper-parameters $\{\beta_1=0.5, \beta_2=0.999\}$. The batch-size is set to $8$. The $\lambda$ parameters for the objective function $\mathcal{L}_{steganography}$ in the steganography stage are set as $\{\lambda_{img}=2000, \lambda_{msg}=10^-5\}$.

\subsection{End-to-End Model}
\noindent \textbf{Architectures.}
$D_{encrypt}$ is a deeper version of the decoder used in the original AdaIN implementation. It consists of 3 nearest up-sampling layers, 13 convolutional layers with kernels of size $3\times 3$, and ReLU activations after each conv-layer except the last one. $E_{decrypt}$ stacks up 8 building blocks, where each building block contains a convolutional layer with kernels of size $3\times 3$, a batch normalization layer, a ReLU activation, and a max-pooling layer with kernel of size $3\times 3$. 

\noindent \textbf{Hyperparameters.}
The learning rate used in our model training is $10^{-4}$. We adopt Adam optimizer~\cite{kingma2014adam} with hyper-parameters $\{\beta_1=0.5, \beta_2=0.999\}$. The batch-size is set to $8$. The $\lambda$ parameters for the objective function $\mathcal{L}_{end2end}$ are set as $\{\lambda_{c}=2, \lambda_{s}=10, \lambda_{dec}=30, \lambda_{inv}=5, \lambda_{des}=5, \lambda_{p}=1\}$.

%% file: full_page_tables.tex

\newcommand\FullPageTableLength{0.18\columnwidth}
\begin{figure*}
\centering

\begin{tabular}{cccccccc}

\parbox[c]{4em}{Gatys~\cite{Gatys2016ImageStyleTransferUsingCNN}}&
\parbox[c]{\FullPageTableLength{}}{\includegraphics[width=\FullPageTableLength{}]{appendix_figures/content_img/lenna.jpg}}&
\parbox[c]{\FullPageTableLength{}}{\includegraphics[width=\FullPageTableLength{}]{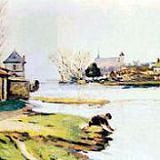}}&
\parbox[c]{\FullPageTableLength{}}{\includegraphics[width=\FullPageTableLength{}]{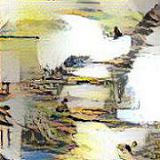}}&
\parbox[c]{\FullPageTableLength{}}{\includegraphics[width=\FullPageTableLength{}]{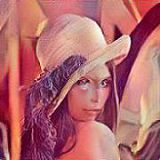}}&
\parbox[c]{\FullPageTableLength{}}{\includegraphics[width=\FullPageTableLength{}]{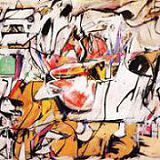}}&
\parbox[c]{\FullPageTableLength{}}{\includegraphics[width=\FullPageTableLength{}]{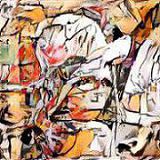}}&
\parbox[c]{\FullPageTableLength{}}{\includegraphics[width=\FullPageTableLength{}]{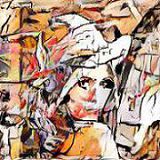}}\\
\parbox[c]{4em}{AdaIN~\cite{Huang2017AdaIN}}&
\parbox[c]{\FullPageTableLength{}}{\includegraphics[width=\FullPageTableLength{}]{appendix_figures/content_img/lenna.jpg}}&
\parbox[c]{\FullPageTableLength{}}{\includegraphics[width=\FullPageTableLength{}]{appendix_figures/style_img/the_resevoir_at_poitiers.jpg}}&
\parbox[c]{\FullPageTableLength{}}{\includegraphics[width=\FullPageTableLength{}]{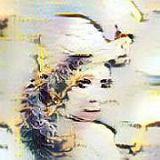}}&
\parbox[c]{\FullPageTableLength{}}{\includegraphics[width=\FullPageTableLength{}]{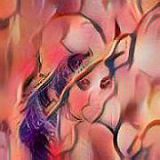}}&
\parbox[c]{\FullPageTableLength{}}{\includegraphics[width=\FullPageTableLength{}]{appendix_figures/style_img/asheville.jpg}}&
\parbox[c]{\FullPageTableLength{}}{\includegraphics[width=\FullPageTableLength{}]{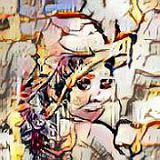}}&
\parbox[c]{\FullPageTableLength{}}{\includegraphics[width=\FullPageTableLength{}]{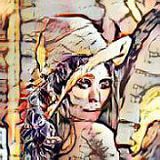}}\\
\parbox[c]{4em}{\begin{tabular}[c]{@{}c@{}}Our\\Two-Stage\end{tabular}}&
\parbox[c]{\FullPageTableLength{}}{\includegraphics[width=\FullPageTableLength{}]{appendix_figures/content_img/lenna.jpg}}&
\parbox[c]{\FullPageTableLength{}}{\includegraphics[width=\FullPageTableLength{}]{appendix_figures/style_img/the_resevoir_at_poitiers.jpg}}&
\parbox[c]{\FullPageTableLength{}}{\includegraphics[width=\FullPageTableLength{}]{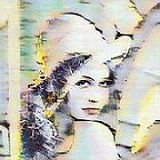}}&
\parbox[c]{\FullPageTableLength{}}{\includegraphics[width=\FullPageTableLength{}]{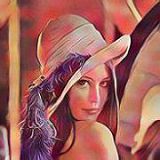}}&
\parbox[c]{\FullPageTableLength{}}{\includegraphics[width=\FullPageTableLength{}]{appendix_figures/style_img/asheville.jpg}}&
\parbox[c]{\FullPageTableLength{}}{\includegraphics[width=\FullPageTableLength{}]{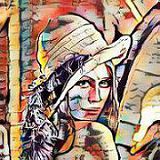}}&
\parbox[c]{\FullPageTableLength{}}{\includegraphics[width=\FullPageTableLength{}]{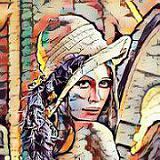}}\\
\parbox[c]{4em}{\begin{tabular}[c]{@{}c@{}}Our\\End-to-End\end{tabular}}&
\parbox[c]{\FullPageTableLength{}}{\includegraphics[width=\FullPageTableLength{}]{appendix_figures/content_img/lenna.jpg}}&
\parbox[c]{\FullPageTableLength{}}{\includegraphics[width=\FullPageTableLength{}]{appendix_figures/style_img/the_resevoir_at_poitiers.jpg}}&
\parbox[c]{\FullPageTableLength{}}{\includegraphics[width=\FullPageTableLength{}]{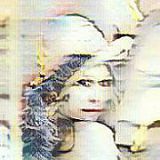}}&
\parbox[c]{\FullPageTableLength{}}{\includegraphics[width=\FullPageTableLength{}]{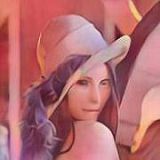}}&
\parbox[c]{\FullPageTableLength{}}{\includegraphics[width=\FullPageTableLength{}]{appendix_figures/style_img/asheville.jpg}}&
\parbox[c]{\FullPageTableLength{}}{\includegraphics[width=\FullPageTableLength{}]{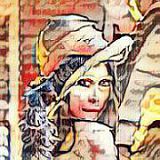}}&
\parbox[c]{\FullPageTableLength{}}{\includegraphics[width=\FullPageTableLength{}]{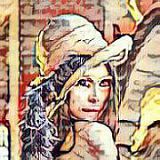}}\\
\hline
\parbox[c]{4em}{Gatys~\cite{Gatys2016ImageStyleTransferUsingCNN}}&
\parbox[c]{\FullPageTableLength{}}{\includegraphics[width=\FullPageTableLength{}]{appendix_figures/content_img/chicago.jpg}}&
\parbox[c]{\FullPageTableLength{}}{\includegraphics[width=\FullPageTableLength{}]{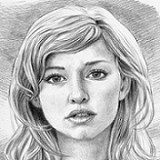}}&
\parbox[c]{\FullPageTableLength{}}{\includegraphics[width=\FullPageTableLength{}]{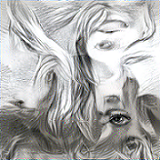}}&
\parbox[c]{\FullPageTableLength{}}{\includegraphics[width=\FullPageTableLength{}]{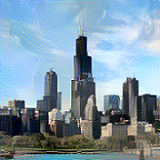}}&
\parbox[c]{\FullPageTableLength{}}{\includegraphics[width=\FullPageTableLength{}]{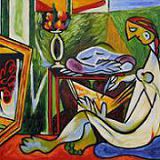}}&
\parbox[c]{\FullPageTableLength{}}{\includegraphics[width=\FullPageTableLength{}]{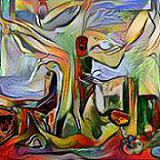}}&
\parbox[c]{\FullPageTableLength{}}{\includegraphics[width=\FullPageTableLength{}]{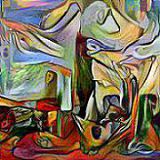}}\\
\parbox[c]{4em}{AdaIN~\cite{Huang2017AdaIN}}&
\parbox[c]{\FullPageTableLength{}}{\includegraphics[width=\FullPageTableLength{}]{appendix_figures/content_img/chicago.jpg}}&
\parbox[c]{\FullPageTableLength{}}{\includegraphics[width=\FullPageTableLength{}]{appendix_figures/style_img/sketch.png}}&
\parbox[c]{\FullPageTableLength{}}{\includegraphics[width=\FullPageTableLength{}]{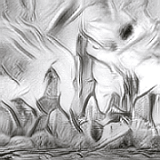}}&
\parbox[c]{\FullPageTableLength{}}{\includegraphics[width=\FullPageTableLength{}]{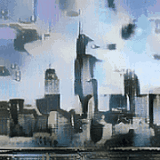}}&
\parbox[c]{\FullPageTableLength{}}{\includegraphics[width=\FullPageTableLength{}]{appendix_figures/style_img/la_muse.jpg}}&
\parbox[c]{\FullPageTableLength{}}{\includegraphics[width=\FullPageTableLength{}]{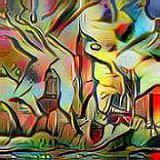}}&
\parbox[c]{\FullPageTableLength{}}{\includegraphics[width=\FullPageTableLength{}]{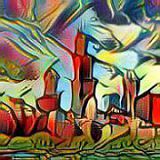}}\\
\parbox[c]{4em}{\begin{tabular}[c]{@{}c@{}}Our\\Two-Stage\end{tabular}}&
\parbox[c]{\FullPageTableLength{}}{\includegraphics[width=\FullPageTableLength{}]{appendix_figures/content_img/chicago.jpg}}&
\parbox[c]{\FullPageTableLength{}}{\includegraphics[width=\FullPageTableLength{}]{appendix_figures/style_img/sketch.png}}&
\parbox[c]{\FullPageTableLength{}}{\includegraphics[width=\FullPageTableLength{}]{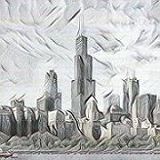}}&
\parbox[c]{\FullPageTableLength{}}{\includegraphics[width=\FullPageTableLength{}]{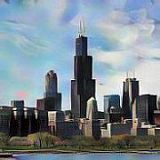}}&
\parbox[c]{\FullPageTableLength{}}{\includegraphics[width=\FullPageTableLength{}]{appendix_figures/style_img/la_muse.jpg}}&
\parbox[c]{\FullPageTableLength{}}{\includegraphics[width=\FullPageTableLength{}]{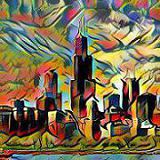}}&
\parbox[c]{\FullPageTableLength{}}{\includegraphics[width=\FullPageTableLength{}]{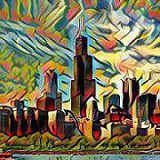}}\\
\parbox[c]{4em}{\begin{tabular}[c]{@{}c@{}}Our\\End-to-End\end{tabular}}&
\parbox[c]{\FullPageTableLength{}}{\includegraphics[width=\FullPageTableLength{}]{appendix_figures/content_img/chicago.jpg}}&
\parbox[c]{\FullPageTableLength{}}{\includegraphics[width=\FullPageTableLength{}]{appendix_figures/style_img/sketch.png}}&
\parbox[c]{\FullPageTableLength{}}{\includegraphics[width=\FullPageTableLength{}]{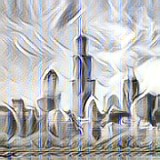}}&
\parbox[c]{\FullPageTableLength{}}{\includegraphics[width=\FullPageTableLength{}]{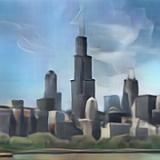}}&
\parbox[c]{\FullPageTableLength{}}{\includegraphics[width=\FullPageTableLength{}]{appendix_figures/style_img/la_muse.jpg}}&
\parbox[c]{\FullPageTableLength{}}{\includegraphics[width=\FullPageTableLength{}]{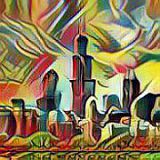}}&
\parbox[c]{\FullPageTableLength{}}{\includegraphics[width=\FullPageTableLength{}]{appendix_figures/e2e_result/serial/E2E_another-chicago__sketch__la_muse.jpg}}\\
\hline
\parbox[c]{4em}{Gatys~\cite{Gatys2016ImageStyleTransferUsingCNN}}&
\parbox[c]{\FullPageTableLength{}}{\includegraphics[width=\FullPageTableLength{}]{appendix_figures/content_img/sailboat.jpg}}&
\parbox[c]{\FullPageTableLength{}}{\includegraphics[width=\FullPageTableLength{}]{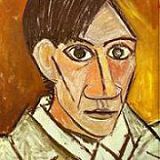}}&
\parbox[c]{\FullPageTableLength{}}{\includegraphics[width=\FullPageTableLength{}]{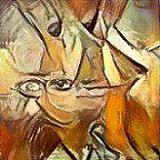}}&
\parbox[c]{\FullPageTableLength{}}{\includegraphics[width=\FullPageTableLength{}]{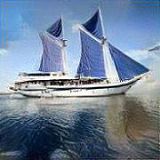}}&
\parbox[c]{\FullPageTableLength{}}{\includegraphics[width=\FullPageTableLength{}]{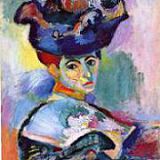}}&
\parbox[c]{\FullPageTableLength{}}{\includegraphics[width=\FullPageTableLength{}]{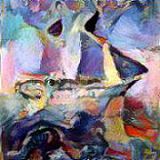}}&
\parbox[c]{\FullPageTableLength{}}{\includegraphics[width=\FullPageTableLength{}]{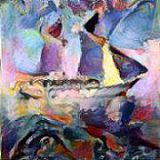}}\\
\parbox[c]{4em}{AdaIN~\cite{Huang2017AdaIN}}&
\parbox[c]{\FullPageTableLength{}}{\includegraphics[width=\FullPageTableLength{}]{appendix_figures/content_img/sailboat.jpg}}&
\parbox[c]{\FullPageTableLength{}}{\includegraphics[width=\FullPageTableLength{}]{appendix_figures/style_img/picasso_self_portrait.jpg}}&
\parbox[c]{\FullPageTableLength{}}{\includegraphics[width=\FullPageTableLength{}]{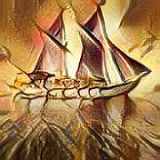}}&
\parbox[c]{\FullPageTableLength{}}{\includegraphics[width=\FullPageTableLength{}]{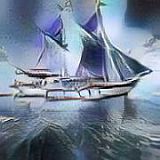}}&
\parbox[c]{\FullPageTableLength{}}{\includegraphics[width=\FullPageTableLength{}]{appendix_figures/style_img/woman_with_hat_matisse.jpg}}&
\parbox[c]{\FullPageTableLength{}}{\includegraphics[width=\FullPageTableLength{}]{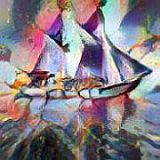}}&
\parbox[c]{\FullPageTableLength{}}{\includegraphics[width=\FullPageTableLength{}]{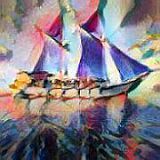}}\\
\parbox[c]{4em}{\begin{tabular}[c]{@{}c@{}}Our\\Two-Stage\end{tabular}}&
\parbox[c]{\FullPageTableLength{}}{\includegraphics[width=\FullPageTableLength{}]{appendix_figures/content_img/sailboat.jpg}}&
\parbox[c]{\FullPageTableLength{}}{\includegraphics[width=\FullPageTableLength{}]{appendix_figures/style_img/picasso_self_portrait.jpg}}&
\parbox[c]{\FullPageTableLength{}}{\includegraphics[width=\FullPageTableLength{}]{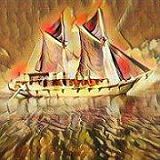}}&
\parbox[c]{\FullPageTableLength{}}{\includegraphics[width=\FullPageTableLength{}]{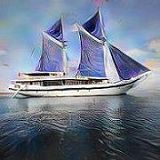}}&
\parbox[c]{\FullPageTableLength{}}{\includegraphics[width=\FullPageTableLength{}]{appendix_figures/style_img/woman_with_hat_matisse.jpg}}&
\parbox[c]{\FullPageTableLength{}}{\includegraphics[width=\FullPageTableLength{}]{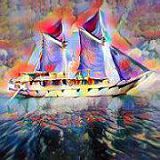}}&
\parbox[c]{\FullPageTableLength{}}{\includegraphics[width=\FullPageTableLength{}]{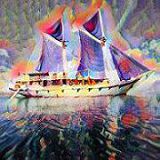}}\\
\parbox[c]{4em}{\begin{tabular}[c]{@{}c@{}}Our\\End-to-End\end{tabular}}&
\parbox[c]{\FullPageTableLength{}}{\includegraphics[width=\FullPageTableLength{}]{appendix_figures/content_img/sailboat.jpg}}&
\parbox[c]{\FullPageTableLength{}}{\includegraphics[width=\FullPageTableLength{}]{appendix_figures/style_img/picasso_self_portrait.jpg}}&
\parbox[c]{\FullPageTableLength{}}{\includegraphics[width=\FullPageTableLength{}]{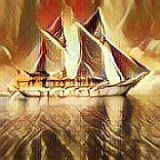}}&
\parbox[c]{\FullPageTableLength{}}{\includegraphics[width=\FullPageTableLength{}]{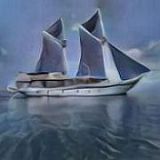}}&
\parbox[c]{\FullPageTableLength{}}{\includegraphics[width=\FullPageTableLength{}]{appendix_figures/style_img/woman_with_hat_matisse.jpg}}&
\parbox[c]{\FullPageTableLength{}}{\includegraphics[width=\FullPageTableLength{}]{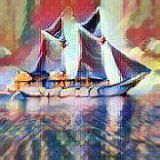}}&
\parbox[c]{\FullPageTableLength{}}{\includegraphics[width=\FullPageTableLength{}]{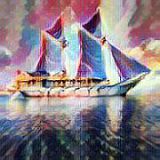}}\\
& Content & Style & Regular & Reverse & \begin{tabular}[c]{@{}c@{}}Second\\Style\end{tabular} & Serial & \begin{tabular}[c]{@{}c@{}}Expected\\Serial\end{tabular}\\
\end{tabular}
\caption{Three sets of additional results to demonstrate the comparison between different methods for regular, reverse, and serial style transfer. The rows in each set sequentially show the results generated by (1) Gatys \etal~\cite{Gatys2016ImageStyleTransferUsingCNN}, (2) AdaIN~\cite{Huang2017AdaIN}, (3) our two-stage model, and (4) our end-to-end model. }
\label{fig:more_results}
\end{figure*}

\begin{figure*}
\centering

\begin{tabular}{cccccccc}

\parbox[c]{4em}{\begin{tabular}[c]{@{}c@{}}Our\\Two-Stage\end{tabular}}&
\parbox[c]{\FullPageTableLength{}}{\includegraphics[width=\FullPageTableLength{}]{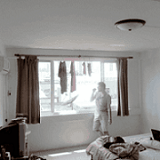}}&
\parbox[c]{\FullPageTableLength{}}{\includegraphics[width=\FullPageTableLength{}]{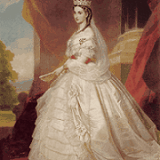}}&
\parbox[c]{\FullPageTableLength{}}{\includegraphics[width=\FullPageTableLength{}]{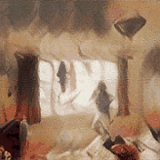}}&
\parbox[c]{\FullPageTableLength{}}{\includegraphics[width=\FullPageTableLength{}]{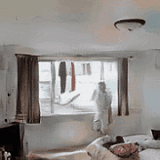}}&
\parbox[c]{\FullPageTableLength{}}{\includegraphics[width=\FullPageTableLength{}]{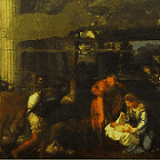}}&
\parbox[c]{\FullPageTableLength{}}{\includegraphics[width=\FullPageTableLength{}]{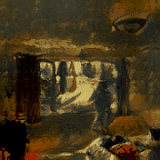}}&
\parbox[c]{\FullPageTableLength{}}{\includegraphics[width=\FullPageTableLength{}]{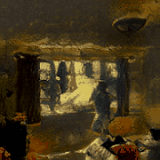}}\\
\parbox[c]{4em}{\begin{tabular}[c]{@{}c@{}}Our\\End-to-End\end{tabular}}&
\parbox[c]{\FullPageTableLength{}}{\includegraphics[width=\FullPageTableLength{}]{appendix_figures/testset_result/r0/content.png}}&
\parbox[c]{\FullPageTableLength{}}{\includegraphics[width=\FullPageTableLength{}]{appendix_figures/testset_result/r0/style.png}}&
\parbox[c]{\FullPageTableLength{}}{\includegraphics[width=\FullPageTableLength{}]{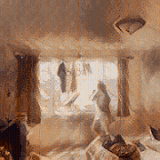}}&
\parbox[c]{\FullPageTableLength{}}{\includegraphics[width=\FullPageTableLength{}]{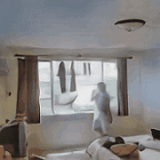}}&
\parbox[c]{\FullPageTableLength{}}{\includegraphics[width=\FullPageTableLength{}]{appendix_figures/testset_result/r0/serial_style.png}}&
\parbox[c]{\FullPageTableLength{}}{\includegraphics[width=\FullPageTableLength{}]{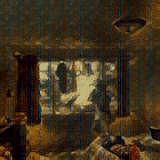}}&
\parbox[c]{\FullPageTableLength{}}{\includegraphics[width=\FullPageTableLength{}]{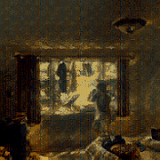}}\\
\hline
\parbox[c]{4em}{\begin{tabular}[c]{@{}c@{}}Our\\Two-Stage\end{tabular}}&
\parbox[c]{\FullPageTableLength{}}{\includegraphics[width=\FullPageTableLength{}]{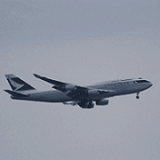}}&
\parbox[c]{\FullPageTableLength{}}{\includegraphics[width=\FullPageTableLength{}]{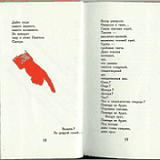}}&
\parbox[c]{\FullPageTableLength{}}{\includegraphics[width=\FullPageTableLength{}]{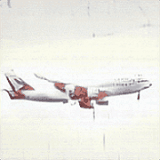}}&
\parbox[c]{\FullPageTableLength{}}{\includegraphics[width=\FullPageTableLength{}]{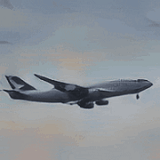}}&
\parbox[c]{\FullPageTableLength{}}{\includegraphics[width=\FullPageTableLength{}]{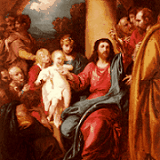}}&
\parbox[c]{\FullPageTableLength{}}{\includegraphics[width=\FullPageTableLength{}]{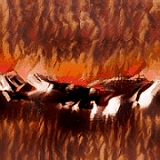}}&
\parbox[c]{\FullPageTableLength{}}{\includegraphics[width=\FullPageTableLength{}]{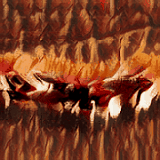}}\\
\parbox[c]{4em}{\begin{tabular}[c]{@{}c@{}}Our\\End-to-End\end{tabular}}&
\parbox[c]{\FullPageTableLength{}}{\includegraphics[width=\FullPageTableLength{}]{appendix_figures/testset_result/r1/content.png}}&
\parbox[c]{\FullPageTableLength{}}{\includegraphics[width=\FullPageTableLength{}]{appendix_figures/testset_result/r1/style.png}}&
\parbox[c]{\FullPageTableLength{}}{\includegraphics[width=\FullPageTableLength{}]{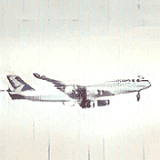}}&
\parbox[c]{\FullPageTableLength{}}{\includegraphics[width=\FullPageTableLength{}]{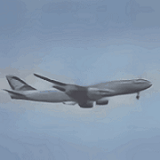}}&
\parbox[c]{\FullPageTableLength{}}{\includegraphics[width=\FullPageTableLength{}]{appendix_figures/testset_result/r1/serial_style.png}}&
\parbox[c]{\FullPageTableLength{}}{\includegraphics[width=\FullPageTableLength{}]{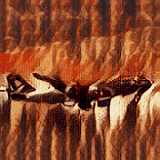}}&
\parbox[c]{\FullPageTableLength{}}{\includegraphics[width=\FullPageTableLength{}]{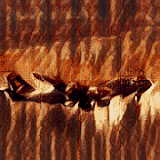}}\\
\hline
\parbox[c]{4em}{\begin{tabular}[c]{@{}c@{}}Our\\Two-Stage\end{tabular}}&
\parbox[c]{\FullPageTableLength{}}{\includegraphics[width=\FullPageTableLength{}]{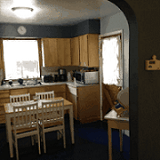}}&
\parbox[c]{\FullPageTableLength{}}{\includegraphics[width=\FullPageTableLength{}]{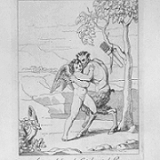}}&
\parbox[c]{\FullPageTableLength{}}{\includegraphics[width=\FullPageTableLength{}]{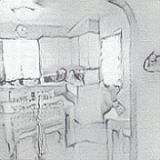}}&
\parbox[c]{\FullPageTableLength{}}{\includegraphics[width=\FullPageTableLength{}]{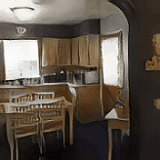}}&
\parbox[c]{\FullPageTableLength{}}{\includegraphics[width=\FullPageTableLength{}]{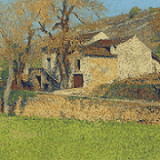}}&
\parbox[c]{\FullPageTableLength{}}{\includegraphics[width=\FullPageTableLength{}]{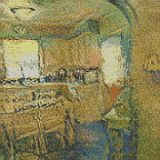}}&
\parbox[c]{\FullPageTableLength{}}{\includegraphics[width=\FullPageTableLength{}]{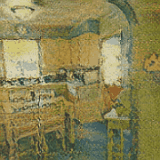}}\\
\parbox[c]{4em}{\begin{tabular}[c]{@{}c@{}}Our\\End-to-End\end{tabular}}&
\parbox[c]{\FullPageTableLength{}}{\includegraphics[width=\FullPageTableLength{}]{appendix_figures/testset_result/r2/content.png}}&
\parbox[c]{\FullPageTableLength{}}{\includegraphics[width=\FullPageTableLength{}]{appendix_figures/testset_result/r2/style.png}}&
\parbox[c]{\FullPageTableLength{}}{\includegraphics[width=\FullPageTableLength{}]{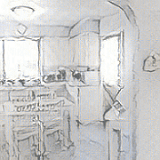}}&
\parbox[c]{\FullPageTableLength{}}{\includegraphics[width=\FullPageTableLength{}]{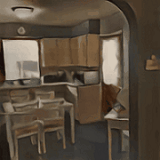}}&
\parbox[c]{\FullPageTableLength{}}{\includegraphics[width=\FullPageTableLength{}]{appendix_figures/testset_result/r2/serial_style.png}}&
\parbox[c]{\FullPageTableLength{}}{\includegraphics[width=\FullPageTableLength{}]{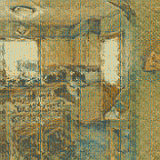}}&
\parbox[c]{\FullPageTableLength{}}{\includegraphics[width=\FullPageTableLength{}]{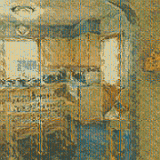}}\\
\hline
\parbox[c]{4em}{\begin{tabular}[c]{@{}c@{}}Our\\Two-Stage\end{tabular}}&
\parbox[c]{\FullPageTableLength{}}{\includegraphics[width=\FullPageTableLength{}]{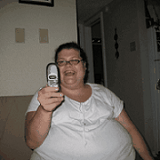}}&
\parbox[c]{\FullPageTableLength{}}{\includegraphics[width=\FullPageTableLength{}]{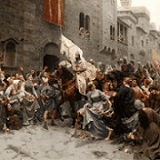}}&
\parbox[c]{\FullPageTableLength{}}{\includegraphics[width=\FullPageTableLength{}]{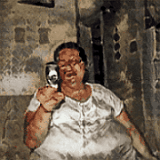}}&
\parbox[c]{\FullPageTableLength{}}{\includegraphics[width=\FullPageTableLength{}]{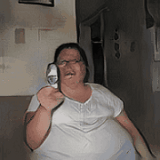}}&
\parbox[c]{\FullPageTableLength{}}{\includegraphics[width=\FullPageTableLength{}]{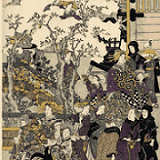}}&
\parbox[c]{\FullPageTableLength{}}{\includegraphics[width=\FullPageTableLength{}]{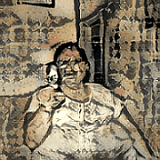}}&
\parbox[c]{\FullPageTableLength{}}{\includegraphics[width=\FullPageTableLength{}]{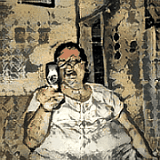}}\\
\parbox[c]{4em}{\begin{tabular}[c]{@{}c@{}}Our\\End-to-End\end{tabular}}&
\parbox[c]{\FullPageTableLength{}}{\includegraphics[width=\FullPageTableLength{}]{appendix_figures/testset_result/r3/content.png}}&
\parbox[c]{\FullPageTableLength{}}{\includegraphics[width=\FullPageTableLength{}]{appendix_figures/testset_result/r3/style.png}}&
\parbox[c]{\FullPageTableLength{}}{\includegraphics[width=\FullPageTableLength{}]{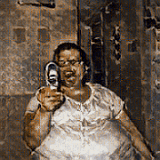}}&
\parbox[c]{\FullPageTableLength{}}{\includegraphics[width=\FullPageTableLength{}]{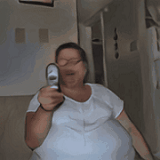}}&
\parbox[c]{\FullPageTableLength{}}{\includegraphics[width=\FullPageTableLength{}]{appendix_figures/testset_result/r3/serial_style.png}}&
\parbox[c]{\FullPageTableLength{}}{\includegraphics[width=\FullPageTableLength{}]{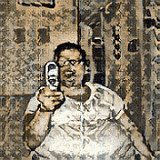}}&
\parbox[c]{\FullPageTableLength{}}{\includegraphics[width=\FullPageTableLength{}]{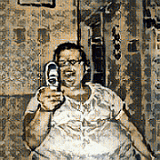}}\\
\hline
\parbox[c]{4em}{\begin{tabular}[c]{@{}c@{}}Our\\Two-Stage\end{tabular}}&
\parbox[c]{\FullPageTableLength{}}{\includegraphics[width=\FullPageTableLength{}]{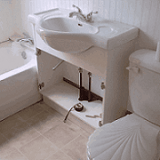}}&
\parbox[c]{\FullPageTableLength{}}{\includegraphics[width=\FullPageTableLength{}]{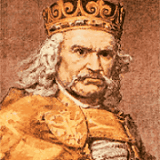}}&
\parbox[c]{\FullPageTableLength{}}{\includegraphics[width=\FullPageTableLength{}]{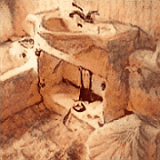}}&
\parbox[c]{\FullPageTableLength{}}{\includegraphics[width=\FullPageTableLength{}]{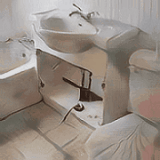}}&
\parbox[c]{\FullPageTableLength{}}{\includegraphics[width=\FullPageTableLength{}]{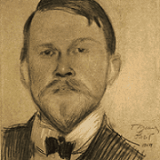}}&
\parbox[c]{\FullPageTableLength{}}{\includegraphics[width=\FullPageTableLength{}]{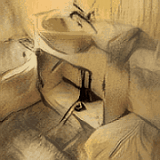}}&
\parbox[c]{\FullPageTableLength{}}{\includegraphics[width=\FullPageTableLength{}]{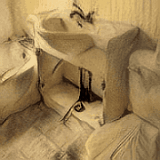}}\\
\parbox[c]{4em}{\begin{tabular}[c]{@{}c@{}}Our\\End-to-End\end{tabular}}&
\parbox[c]{\FullPageTableLength{}}{\includegraphics[width=\FullPageTableLength{}]{appendix_figures/testset_result/r4/content.png}}&
\parbox[c]{\FullPageTableLength{}}{\includegraphics[width=\FullPageTableLength{}]{appendix_figures/testset_result/r4/style.png}}&
\parbox[c]{\FullPageTableLength{}}{\includegraphics[width=\FullPageTableLength{}]{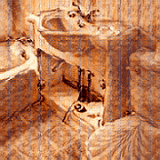}}&
\parbox[c]{\FullPageTableLength{}}{\includegraphics[width=\FullPageTableLength{}]{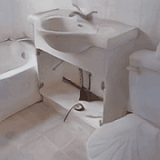}}&
\parbox[c]{\FullPageTableLength{}}{\includegraphics[width=\FullPageTableLength{}]{appendix_figures/testset_result/r4/serial_style.png}}&
\parbox[c]{\FullPageTableLength{}}{\includegraphics[width=\FullPageTableLength{}]{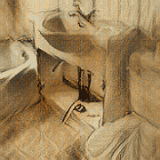}}&
\parbox[c]{\FullPageTableLength{}}{\includegraphics[width=\FullPageTableLength{}]{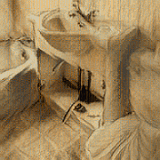}}\\
\hline
\parbox[c]{4em}{\begin{tabular}[c]{@{}c@{}}Our\\Two-Stage\end{tabular}}&
\parbox[c]{\FullPageTableLength{}}{\includegraphics[width=\FullPageTableLength{}]{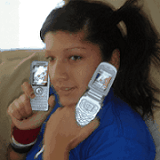}}&
\parbox[c]{\FullPageTableLength{}}{\includegraphics[width=\FullPageTableLength{}]{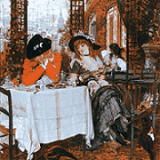}}&
\parbox[c]{\FullPageTableLength{}}{\includegraphics[width=\FullPageTableLength{}]{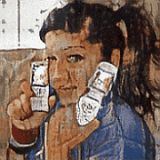}}&
\parbox[c]{\FullPageTableLength{}}{\includegraphics[width=\FullPageTableLength{}]{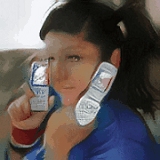}}&
\parbox[c]{\FullPageTableLength{}}{\includegraphics[width=\FullPageTableLength{}]{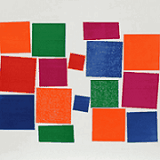}}&
\parbox[c]{\FullPageTableLength{}}{\includegraphics[width=\FullPageTableLength{}]{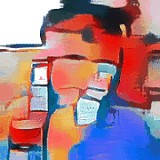}}&
\parbox[c]{\FullPageTableLength{}}{\includegraphics[width=\FullPageTableLength{}]{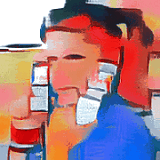}}\\
\parbox[c]{4em}{\begin{tabular}[c]{@{}c@{}}Our\\End-to-End\end{tabular}}&
\parbox[c]{\FullPageTableLength{}}{\includegraphics[width=\FullPageTableLength{}]{appendix_figures/testset_result/r5/content.png}}&
\parbox[c]{\FullPageTableLength{}}{\includegraphics[width=\FullPageTableLength{}]{appendix_figures/testset_result/r5/style.png}}&
\parbox[c]{\FullPageTableLength{}}{\includegraphics[width=\FullPageTableLength{}]{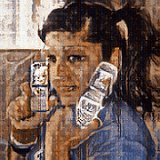}}&
\parbox[c]{\FullPageTableLength{}}{\includegraphics[width=\FullPageTableLength{}]{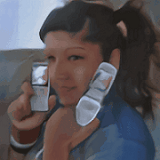}}&
\parbox[c]{\FullPageTableLength{}}{\includegraphics[width=\FullPageTableLength{}]{appendix_figures/testset_result/r5/serial_style.png}}&
\parbox[c]{\FullPageTableLength{}}{\includegraphics[width=\FullPageTableLength{}]{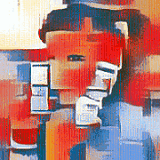}}&
\parbox[c]{\FullPageTableLength{}}{\includegraphics[width=\FullPageTableLength{}]{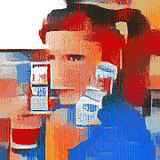}}\\

& Content & Style & Regular & Reverse & \begin{tabular}[c]{@{}c@{}}Second\\Style\end{tabular} & Serial & \begin{tabular}[c]{@{}c@{}}Expected\\Serial\end{tabular}\\
\end{tabular}
\caption{
Example results of our proposed models in regular, reverse, and serial style transfer, based on diverse sets of content and style images from MS-COCO \cite{lin2014microsoft} and WikiArt \cite{wiki_art} datasets respectively.
}
\label{fig:more_test_results}
\end{figure*}

\begin{figure*}
\centering

\begin{tabular}{cccccccc}

\parbox[c]{4em}{\begin{tabular}[c]{@{}c@{}}Our\\Two-Stage\end{tabular}}&
\parbox[c]{\FullPageTableLength{}}{\includegraphics[width=\FullPageTableLength{}]{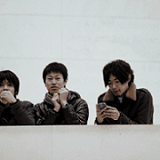}}&
\parbox[c]{\FullPageTableLength{}}{\includegraphics[width=\FullPageTableLength{}]{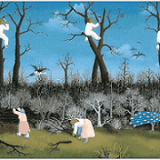}}&
\parbox[c]{\FullPageTableLength{}}{\includegraphics[width=\FullPageTableLength{}]{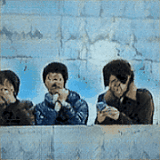}}&
\parbox[c]{\FullPageTableLength{}}{\includegraphics[width=\FullPageTableLength{}]{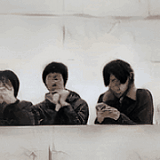}}&
\parbox[c]{\FullPageTableLength{}}{\includegraphics[width=\FullPageTableLength{}]{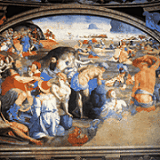}}&
\parbox[c]{\FullPageTableLength{}}{\includegraphics[width=\FullPageTableLength{}]{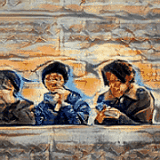}}&
\parbox[c]{\FullPageTableLength{}}{\includegraphics[width=\FullPageTableLength{}]{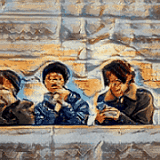}}\\
\parbox[c]{4em}{\begin{tabular}[c]{@{}c@{}}Our\\End-to-End\end{tabular}}&
\parbox[c]{\FullPageTableLength{}}{\includegraphics[width=\FullPageTableLength{}]{appendix_figures/testset_result/r6/content.png}}&
\parbox[c]{\FullPageTableLength{}}{\includegraphics[width=\FullPageTableLength{}]{appendix_figures/testset_result/r6/style.png}}&
\parbox[c]{\FullPageTableLength{}}{\includegraphics[width=\FullPageTableLength{}]{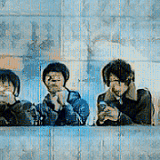}}&
\parbox[c]{\FullPageTableLength{}}{\includegraphics[width=\FullPageTableLength{}]{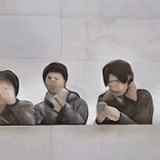}}&
\parbox[c]{\FullPageTableLength{}}{\includegraphics[width=\FullPageTableLength{}]{appendix_figures/testset_result/r6/serial_style.png}}&
\parbox[c]{\FullPageTableLength{}}{\includegraphics[width=\FullPageTableLength{}]{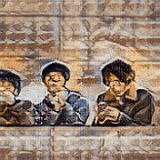}}&
\parbox[c]{\FullPageTableLength{}}{\includegraphics[width=\FullPageTableLength{}]{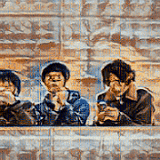}}\\
\hline
\parbox[c]{4em}{\begin{tabular}[c]{@{}c@{}}Our\\Two-Stage\end{tabular}}&
\parbox[c]{\FullPageTableLength{}}{\includegraphics[width=\FullPageTableLength{}]{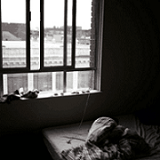}}&
\parbox[c]{\FullPageTableLength{}}{\includegraphics[width=\FullPageTableLength{}]{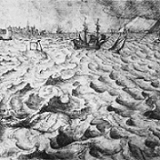}}&
\parbox[c]{\FullPageTableLength{}}{\includegraphics[width=\FullPageTableLength{}]{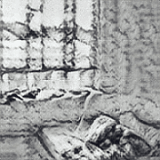}}&
\parbox[c]{\FullPageTableLength{}}{\includegraphics[width=\FullPageTableLength{}]{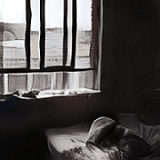}}&
\parbox[c]{\FullPageTableLength{}}{\includegraphics[width=\FullPageTableLength{}]{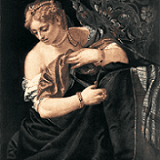}}&
\parbox[c]{\FullPageTableLength{}}{\includegraphics[width=\FullPageTableLength{}]{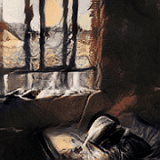}}&
\parbox[c]{\FullPageTableLength{}}{\includegraphics[width=\FullPageTableLength{}]{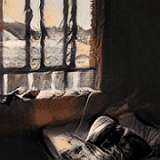}}\\
\parbox[c]{4em}{\begin{tabular}[c]{@{}c@{}}Our\\End-to-End\end{tabular}}&
\parbox[c]{\FullPageTableLength{}}{\includegraphics[width=\FullPageTableLength{}]{appendix_figures/testset_result/r7/content.png}}&
\parbox[c]{\FullPageTableLength{}}{\includegraphics[width=\FullPageTableLength{}]{appendix_figures/testset_result/r7/style.png}}&
\parbox[c]{\FullPageTableLength{}}{\includegraphics[width=\FullPageTableLength{}]{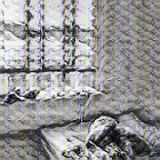}}&
\parbox[c]{\FullPageTableLength{}}{\includegraphics[width=\FullPageTableLength{}]{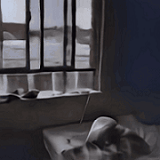}}&
\parbox[c]{\FullPageTableLength{}}{\includegraphics[width=\FullPageTableLength{}]{appendix_figures/testset_result/r7/serial_style.png}}&
\parbox[c]{\FullPageTableLength{}}{\includegraphics[width=\FullPageTableLength{}]{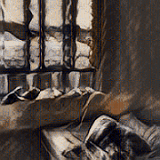}}&
\parbox[c]{\FullPageTableLength{}}{\includegraphics[width=\FullPageTableLength{}]{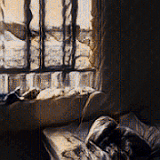}}\\
\hline
\parbox[c]{4em}{\begin{tabular}[c]{@{}c@{}}Our\\Two-Stage\end{tabular}}&
\parbox[c]{\FullPageTableLength{}}{\includegraphics[width=\FullPageTableLength{}]{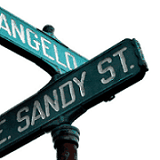}}&
\parbox[c]{\FullPageTableLength{}}{\includegraphics[width=\FullPageTableLength{}]{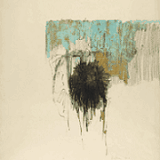}}&
\parbox[c]{\FullPageTableLength{}}{\includegraphics[width=\FullPageTableLength{}]{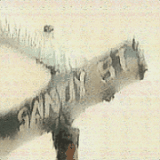}}&
\parbox[c]{\FullPageTableLength{}}{\includegraphics[width=\FullPageTableLength{}]{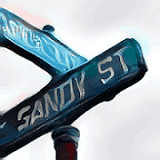}}&
\parbox[c]{\FullPageTableLength{}}{\includegraphics[width=\FullPageTableLength{}]{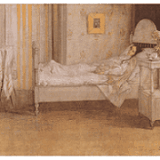}}&
\parbox[c]{\FullPageTableLength{}}{\includegraphics[width=\FullPageTableLength{}]{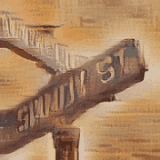}}&
\parbox[c]{\FullPageTableLength{}}{\includegraphics[width=\FullPageTableLength{}]{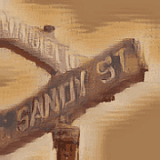}}\\
\parbox[c]{4em}{\begin{tabular}[c]{@{}c@{}}Our\\End-to-End\end{tabular}}&
\parbox[c]{\FullPageTableLength{}}{\includegraphics[width=\FullPageTableLength{}]{appendix_figures/testset_result/r8/content.png}}&
\parbox[c]{\FullPageTableLength{}}{\includegraphics[width=\FullPageTableLength{}]{appendix_figures/testset_result/r8/style.png}}&
\parbox[c]{\FullPageTableLength{}}{\includegraphics[width=\FullPageTableLength{}]{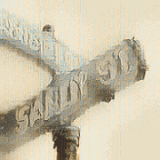}}&
\parbox[c]{\FullPageTableLength{}}{\includegraphics[width=\FullPageTableLength{}]{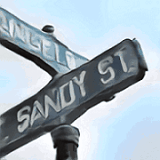}}&
\parbox[c]{\FullPageTableLength{}}{\includegraphics[width=\FullPageTableLength{}]{appendix_figures/testset_result/r8/serial_style.png}}&
\parbox[c]{\FullPageTableLength{}}{\includegraphics[width=\FullPageTableLength{}]{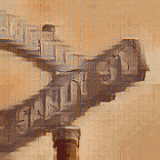}}&
\parbox[c]{\FullPageTableLength{}}{\includegraphics[width=\FullPageTableLength{}]{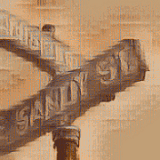}}\\
\hline
\parbox[c]{4em}{\begin{tabular}[c]{@{}c@{}}Our\\Two-Stage\end{tabular}}&
\parbox[c]{\FullPageTableLength{}}{\includegraphics[width=\FullPageTableLength{}]{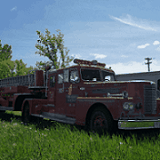}}&
\parbox[c]{\FullPageTableLength{}}{\includegraphics[width=\FullPageTableLength{}]{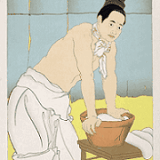}}&
\parbox[c]{\FullPageTableLength{}}{\includegraphics[width=\FullPageTableLength{}]{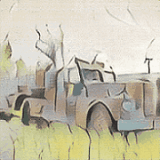}}&
\parbox[c]{\FullPageTableLength{}}{\includegraphics[width=\FullPageTableLength{}]{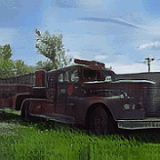}}&
\parbox[c]{\FullPageTableLength{}}{\includegraphics[width=\FullPageTableLength{}]{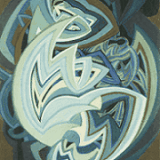}}&
\parbox[c]{\FullPageTableLength{}}{\includegraphics[width=\FullPageTableLength{}]{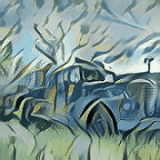}}&
\parbox[c]{\FullPageTableLength{}}{\includegraphics[width=\FullPageTableLength{}]{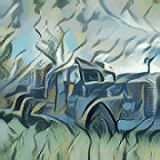}}\\
\parbox[c]{4em}{\begin{tabular}[c]{@{}c@{}}Our\\End-to-End\end{tabular}}&
\parbox[c]{\FullPageTableLength{}}{\includegraphics[width=\FullPageTableLength{}]{appendix_figures/testset_result/r9/content.png}}&
\parbox[c]{\FullPageTableLength{}}{\includegraphics[width=\FullPageTableLength{}]{appendix_figures/testset_result/r9/style.png}}&
\parbox[c]{\FullPageTableLength{}}{\includegraphics[width=\FullPageTableLength{}]{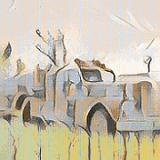}}&
\parbox[c]{\FullPageTableLength{}}{\includegraphics[width=\FullPageTableLength{}]{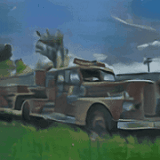}}&
\parbox[c]{\FullPageTableLength{}}{\includegraphics[width=\FullPageTableLength{}]{appendix_figures/testset_result/r9/serial_style.png}}&
\parbox[c]{\FullPageTableLength{}}{\includegraphics[width=\FullPageTableLength{}]{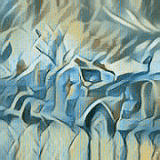}}&
\parbox[c]{\FullPageTableLength{}}{\includegraphics[width=\FullPageTableLength{}]{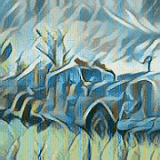}}\\
\hline
\parbox[c]{4em}{\begin{tabular}[c]{@{}c@{}}Our\\Two-Stage\end{tabular}}&
\parbox[c]{\FullPageTableLength{}}{\includegraphics[width=\FullPageTableLength{}]{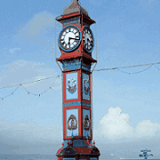}}&
\parbox[c]{\FullPageTableLength{}}{\includegraphics[width=\FullPageTableLength{}]{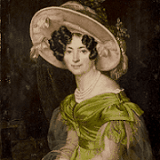}}&
\parbox[c]{\FullPageTableLength{}}{\includegraphics[width=\FullPageTableLength{}]{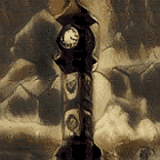}}&
\parbox[c]{\FullPageTableLength{}}{\includegraphics[width=\FullPageTableLength{}]{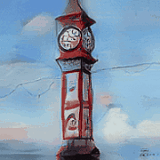}}&
\parbox[c]{\FullPageTableLength{}}{\includegraphics[width=\FullPageTableLength{}]{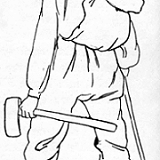}}&
\parbox[c]{\FullPageTableLength{}}{\includegraphics[width=\FullPageTableLength{}]{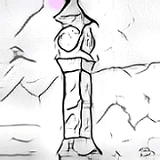}}&
\parbox[c]{\FullPageTableLength{}}{\includegraphics[width=\FullPageTableLength{}]{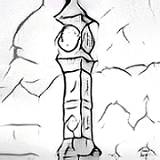}}\\
\parbox[c]{4em}{\begin{tabular}[c]{@{}c@{}}Our\\End-to-End\end{tabular}}&
\parbox[c]{\FullPageTableLength{}}{\includegraphics[width=\FullPageTableLength{}]{appendix_figures/testset_result/r10/content.png}}&
\parbox[c]{\FullPageTableLength{}}{\includegraphics[width=\FullPageTableLength{}]{appendix_figures/testset_result/r10/style.png}}&
\parbox[c]{\FullPageTableLength{}}{\includegraphics[width=\FullPageTableLength{}]{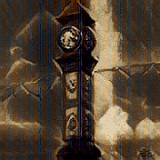}}&
\parbox[c]{\FullPageTableLength{}}{\includegraphics[width=\FullPageTableLength{}]{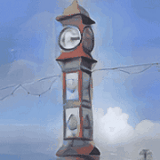}}&
\parbox[c]{\FullPageTableLength{}}{\includegraphics[width=\FullPageTableLength{}]{appendix_figures/testset_result/r10/serial_style.png}}&
\parbox[c]{\FullPageTableLength{}}{\includegraphics[width=\FullPageTableLength{}]{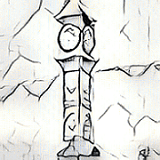}}&
\parbox[c]{\FullPageTableLength{}}{\includegraphics[width=\FullPageTableLength{}]{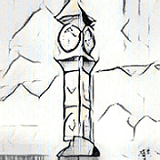}}\\
\hline
\parbox[c]{4em}{\begin{tabular}[c]{@{}c@{}}Our\\Two-Stage\end{tabular}}&
\parbox[c]{\FullPageTableLength{}}{\includegraphics[width=\FullPageTableLength{}]{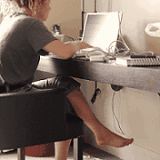}}&
\parbox[c]{\FullPageTableLength{}}{\includegraphics[width=\FullPageTableLength{}]{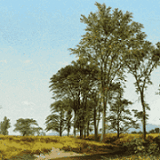}}&
\parbox[c]{\FullPageTableLength{}}{\includegraphics[width=\FullPageTableLength{}]{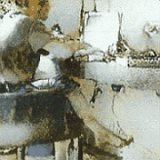}}&
\parbox[c]{\FullPageTableLength{}}{\includegraphics[width=\FullPageTableLength{}]{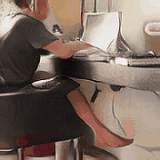}}&
\parbox[c]{\FullPageTableLength{}}{\includegraphics[width=\FullPageTableLength{}]{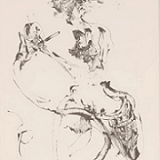}}&
\parbox[c]{\FullPageTableLength{}}{\includegraphics[width=\FullPageTableLength{}]{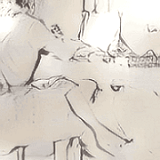}}&
\parbox[c]{\FullPageTableLength{}}{\includegraphics[width=\FullPageTableLength{}]{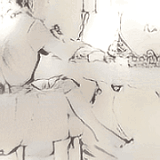}}\\
\parbox[c]{4em}{\begin{tabular}[c]{@{}c@{}}Our\\End-to-End\end{tabular}}&
\parbox[c]{\FullPageTableLength{}}{\includegraphics[width=\FullPageTableLength{}]{appendix_figures/testset_result/r11/content.png}}&
\parbox[c]{\FullPageTableLength{}}{\includegraphics[width=\FullPageTableLength{}]{appendix_figures/testset_result/r11/style.png}}&
\parbox[c]{\FullPageTableLength{}}{\includegraphics[width=\FullPageTableLength{}]{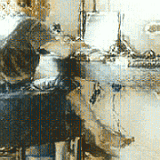}}&
\parbox[c]{\FullPageTableLength{}}{\includegraphics[width=\FullPageTableLength{}]{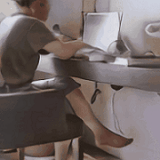}}&
\parbox[c]{\FullPageTableLength{}}{\includegraphics[width=\FullPageTableLength{}]{appendix_figures/testset_result/r11/serial_style.png}}&
\parbox[c]{\FullPageTableLength{}}{\includegraphics[width=\FullPageTableLength{}]{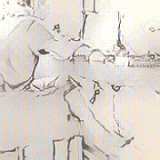}}&
\parbox[c]{\FullPageTableLength{}}{\includegraphics[width=\FullPageTableLength{}]{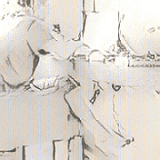}}\\

& Content & Style & Regular & Reverse & \begin{tabular}[c]{@{}c@{}}Second\\Style\end{tabular} & Serial & \begin{tabular}[c]{@{}c@{}}Expected\\Serial\end{tabular}\\
\end{tabular}
\caption{
Example results of our proposed models in regular, reverse, and serial style transfer, based on diverse sets of content and style images from MS-COCO \cite{lin2014microsoft} and WikiArt \cite{wiki_art} datasets respectively.
}
\label{fig:more_test_results_1}
\end{figure*}


\newcommand\MultipleSerialLength{0.18\columnwidth}
\begin{figure*}
\centering

\begin{tabular}{cc||ccccccc}

&  &
\parbox[c]{\MultipleSerialLength{}}{\includegraphics[width=\MultipleSerialLength{}]{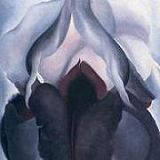}}&
\parbox[c]{\MultipleSerialLength{}}{\includegraphics[width=\MultipleSerialLength{}]{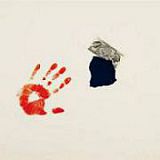}}&
\parbox[c]{\MultipleSerialLength{}}{\includegraphics[width=\MultipleSerialLength{}]{appendix_figures/style_img/asheville.jpg}}&
\parbox[c]{\MultipleSerialLength{}}{\includegraphics[width=\MultipleSerialLength{}]{appendix_figures/style_img/la_muse.jpg}}&
\parbox[c]{\MultipleSerialLength{}}{\includegraphics[width=\MultipleSerialLength{}]{appendix_figures/style_img/woman_with_hat_matisse.jpg}}&
\parbox[c]{\MultipleSerialLength{}}{\includegraphics[width=\MultipleSerialLength{}]{appendix_figures/style_img/sketch.png}}&
\parbox[c]{\MultipleSerialLength{}}{\includegraphics[width=\MultipleSerialLength{}]{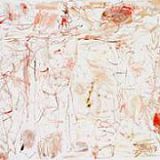}}\\ \hline

\parbox[c]{0.5em}{(1)}&
\parbox[c]{\MultipleSerialLength{}}{\includegraphics[width=\MultipleSerialLength{}]{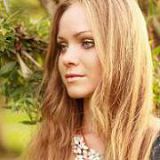}}&
\parbox[c]{\MultipleSerialLength{}}{\includegraphics[width=\MultipleSerialLength{}]{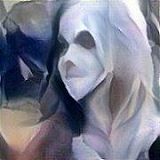}}&
\parbox[c]{\MultipleSerialLength{}}{\includegraphics[width=\MultipleSerialLength{}]{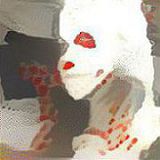}}&
\parbox[c]{\MultipleSerialLength{}}{\includegraphics[width=\MultipleSerialLength{}]{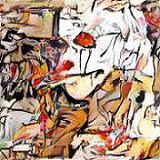}}&
\parbox[c]{\MultipleSerialLength{}}{\includegraphics[width=\MultipleSerialLength{}]{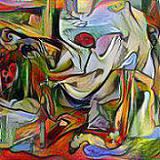}}&
\parbox[c]{\MultipleSerialLength{}}{\includegraphics[width=\MultipleSerialLength{}]{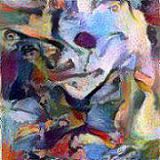}}&
\parbox[c]{\MultipleSerialLength{}}{\includegraphics[width=\MultipleSerialLength{}]{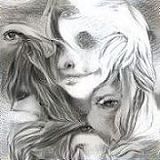}}&
\parbox[c]{\MultipleSerialLength{}}{\includegraphics[width=\MultipleSerialLength{}]{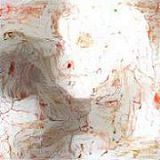}}\\

\parbox[c]{0.5em}{(2)}&
\parbox[c]{\MultipleSerialLength{}}{\includegraphics[width=\MultipleSerialLength{}]{appendix_figures/content_img/blonde_girl.jpg}}&
\parbox[c]{\MultipleSerialLength{}}{\includegraphics[width=\MultipleSerialLength{}]{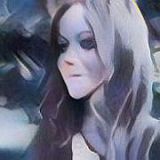}}&
\parbox[c]{\MultipleSerialLength{}}{\includegraphics[width=\MultipleSerialLength{}]{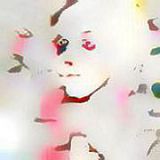}}&
\parbox[c]{\MultipleSerialLength{}}{\includegraphics[width=\MultipleSerialLength{}]{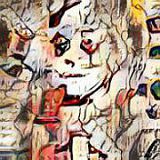}}&
\parbox[c]{\MultipleSerialLength{}}{\includegraphics[width=\MultipleSerialLength{}]{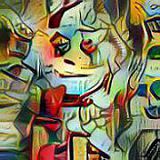}}&
\parbox[c]{\MultipleSerialLength{}}{\includegraphics[width=\MultipleSerialLength{}]{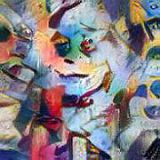}}&
\parbox[c]{\MultipleSerialLength{}}{\includegraphics[width=\MultipleSerialLength{}]{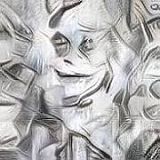}}&
\parbox[c]{\MultipleSerialLength{}}{\includegraphics[width=\MultipleSerialLength{}]{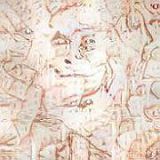}}\\

\parbox[c]{0.5em}{(3)}&
\parbox[c]{\MultipleSerialLength{}}{\includegraphics[width=\MultipleSerialLength{}]{appendix_figures/content_img/blonde_girl.jpg}}&
\parbox[c]{\MultipleSerialLength{}}{\includegraphics[width=\MultipleSerialLength{}]{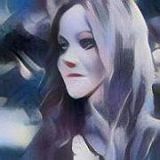}}&
\parbox[c]{\MultipleSerialLength{}}{\includegraphics[width=\MultipleSerialLength{}]{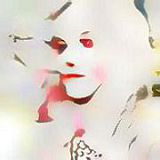}}&
\parbox[c]{\MultipleSerialLength{}}{\includegraphics[width=\MultipleSerialLength{}]{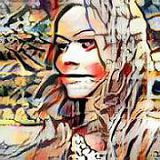}}&
\parbox[c]{\MultipleSerialLength{}}{\includegraphics[width=\MultipleSerialLength{}]{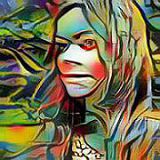}}&
\parbox[c]{\MultipleSerialLength{}}{\includegraphics[width=\MultipleSerialLength{}]{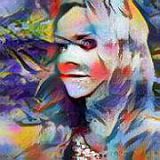}}&
\parbox[c]{\MultipleSerialLength{}}{\includegraphics[width=\MultipleSerialLength{}]{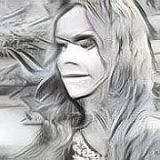}}&
\parbox[c]{\MultipleSerialLength{}}{\includegraphics[width=\MultipleSerialLength{}]{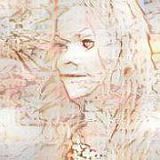}}\\

\parbox[c]{0.5em}{(4)}&
\parbox[c]{\MultipleSerialLength{}}{\includegraphics[width=\MultipleSerialLength{}]{appendix_figures/content_img/blonde_girl.jpg}}&
\parbox[c]{\MultipleSerialLength{}}{\includegraphics[width=\MultipleSerialLength{}]{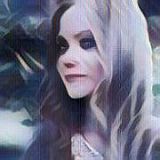}}&
\parbox[c]{\MultipleSerialLength{}}{\includegraphics[width=\MultipleSerialLength{}]{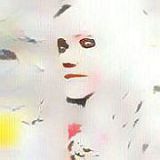}}&
\parbox[c]{\MultipleSerialLength{}}{\includegraphics[width=\MultipleSerialLength{}]{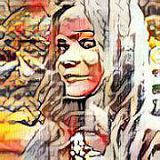}}&
\parbox[c]{\MultipleSerialLength{}}{\includegraphics[width=\MultipleSerialLength{}]{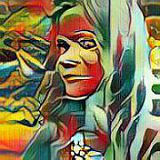}}&
\parbox[c]{\MultipleSerialLength{}}{\includegraphics[width=\MultipleSerialLength{}]{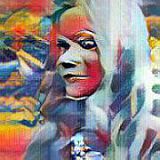}}&
\parbox[c]{\MultipleSerialLength{}}{\includegraphics[width=\MultipleSerialLength{}]{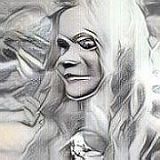}}&
\parbox[c]{\MultipleSerialLength{}}{\includegraphics[width=\MultipleSerialLength{}]{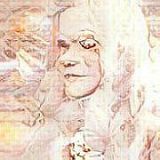}}\\
\hline

\parbox[c]{0.5em}{(1)}&
\parbox[c]{\MultipleSerialLength{}}{\includegraphics[width=\MultipleSerialLength{}]{appendix_figures/content_img/chicago.jpg}}&
\parbox[c]{\MultipleSerialLength{}}{\includegraphics[width=\MultipleSerialLength{}]{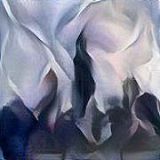}}&
\parbox[c]{\MultipleSerialLength{}}{\includegraphics[width=\MultipleSerialLength{}]{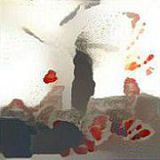}}&
\parbox[c]{\MultipleSerialLength{}}{\includegraphics[width=\MultipleSerialLength{}]{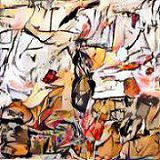}}&
\parbox[c]{\MultipleSerialLength{}}{\includegraphics[width=\MultipleSerialLength{}]{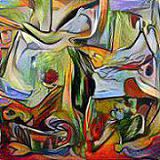}}&
\parbox[c]{\MultipleSerialLength{}}{\includegraphics[width=\MultipleSerialLength{}]{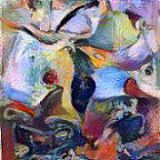}}&
\parbox[c]{\MultipleSerialLength{}}{\includegraphics[width=\MultipleSerialLength{}]{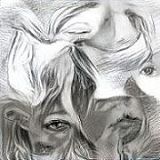}}&
\parbox[c]{\MultipleSerialLength{}}{\includegraphics[width=\MultipleSerialLength{}]{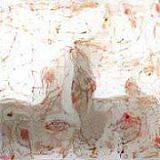}}\\

\parbox[c]{0.5em}{(2)}&
\parbox[c]{\MultipleSerialLength{}}{\includegraphics[width=\MultipleSerialLength{}]{appendix_figures/content_img/chicago.jpg}}&
\parbox[c]{\MultipleSerialLength{}}{\includegraphics[width=\MultipleSerialLength{}]{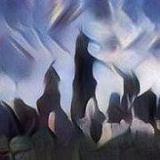}}&
\parbox[c]{\MultipleSerialLength{}}{\includegraphics[width=\MultipleSerialLength{}]{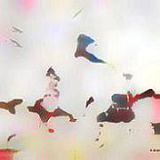}}&
\parbox[c]{\MultipleSerialLength{}}{\includegraphics[width=\MultipleSerialLength{}]{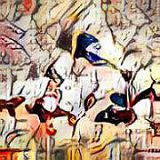}}&
\parbox[c]{\MultipleSerialLength{}}{\includegraphics[width=\MultipleSerialLength{}]{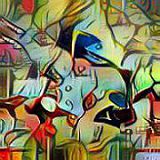}}&
\parbox[c]{\MultipleSerialLength{}}{\includegraphics[width=\MultipleSerialLength{}]{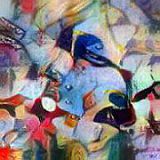}}&
\parbox[c]{\MultipleSerialLength{}}{\includegraphics[width=\MultipleSerialLength{}]{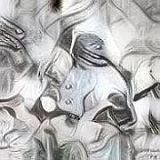}}&
\parbox[c]{\MultipleSerialLength{}}{\includegraphics[width=\MultipleSerialLength{}]{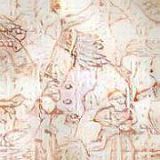}}\\

\parbox[c]{0.5em}{(3)}&
\parbox[c]{\MultipleSerialLength{}}{\includegraphics[width=\MultipleSerialLength{}]{appendix_figures/content_img/chicago.jpg}}&
\parbox[c]{\MultipleSerialLength{}}{\includegraphics[width=\MultipleSerialLength{}]{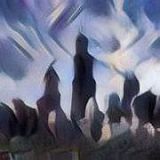}}&
\parbox[c]{\MultipleSerialLength{}}{\includegraphics[width=\MultipleSerialLength{}]{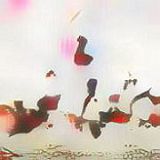}}&
\parbox[c]{\MultipleSerialLength{}}{\includegraphics[width=\MultipleSerialLength{}]{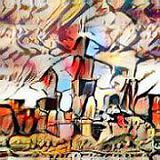}}&
\parbox[c]{\MultipleSerialLength{}}{\includegraphics[width=\MultipleSerialLength{}]{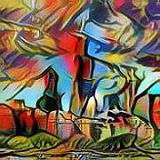}}&
\parbox[c]{\MultipleSerialLength{}}{\includegraphics[width=\MultipleSerialLength{}]{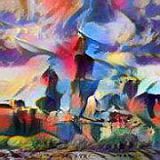}}&
\parbox[c]{\MultipleSerialLength{}}{\includegraphics[width=\MultipleSerialLength{}]{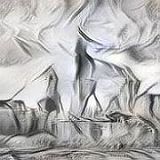}}&
\parbox[c]{\MultipleSerialLength{}}{\includegraphics[width=\MultipleSerialLength{}]{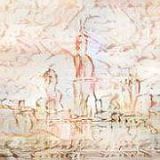}}\\

\parbox[c]{0.5em}{(4)}&
\parbox[c]{\MultipleSerialLength{}}{\includegraphics[width=\MultipleSerialLength{}]{appendix_figures/content_img/chicago.jpg}}&
\parbox[c]{\MultipleSerialLength{}}{\includegraphics[width=\MultipleSerialLength{}]{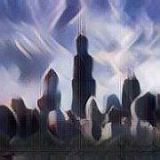}}&
\parbox[c]{\MultipleSerialLength{}}{\includegraphics[width=\MultipleSerialLength{}]{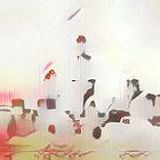}}&
\parbox[c]{\MultipleSerialLength{}}{\includegraphics[width=\MultipleSerialLength{}]{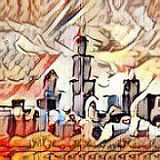}}&
\parbox[c]{\MultipleSerialLength{}}{\includegraphics[width=\MultipleSerialLength{}]{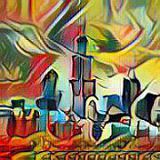}}&
\parbox[c]{\MultipleSerialLength{}}{\includegraphics[width=\MultipleSerialLength{}]{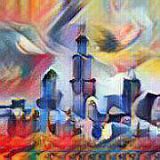}}&
\parbox[c]{\MultipleSerialLength{}}{\includegraphics[width=\MultipleSerialLength{}]{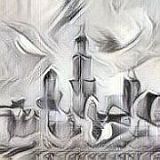}}&
\parbox[c]{\MultipleSerialLength{}}{\includegraphics[width=\MultipleSerialLength{}]{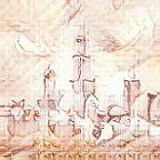}}\\
\hline

\parbox[c]{0.5em}{(1)}&
\parbox[c]{\MultipleSerialLength{}}{\includegraphics[width=\MultipleSerialLength{}]{appendix_figures/content_img/sailboat.jpg}}&
\parbox[c]{\MultipleSerialLength{}}{\includegraphics[width=\MultipleSerialLength{}]{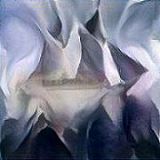}}&
\parbox[c]{\MultipleSerialLength{}}{\includegraphics[width=\MultipleSerialLength{}]{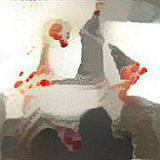}}&
\parbox[c]{\MultipleSerialLength{}}{\includegraphics[width=\MultipleSerialLength{}]{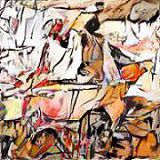}}&
\parbox[c]{\MultipleSerialLength{}}{\includegraphics[width=\MultipleSerialLength{}]{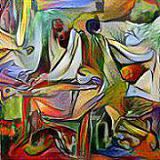}}&
\parbox[c]{\MultipleSerialLength{}}{\includegraphics[width=\MultipleSerialLength{}]{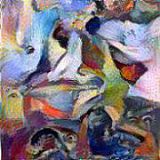}}&
\parbox[c]{\MultipleSerialLength{}}{\includegraphics[width=\MultipleSerialLength{}]{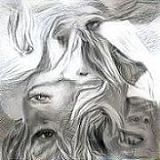}}&
\parbox[c]{\MultipleSerialLength{}}{\includegraphics[width=\MultipleSerialLength{}]{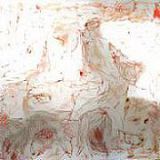}}\\

\parbox[c]{0.5em}{(2)}&
\parbox[c]{\MultipleSerialLength{}}{\includegraphics[width=\MultipleSerialLength{}]{appendix_figures/content_img/sailboat.jpg}}&
\parbox[c]{\MultipleSerialLength{}}{\includegraphics[width=\MultipleSerialLength{}]{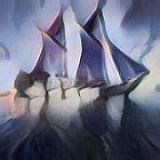}}&
\parbox[c]{\MultipleSerialLength{}}{\includegraphics[width=\MultipleSerialLength{}]{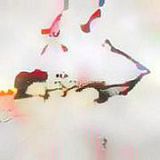}}&
\parbox[c]{\MultipleSerialLength{}}{\includegraphics[width=\MultipleSerialLength{}]{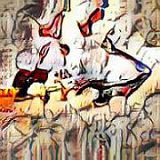}}&
\parbox[c]{\MultipleSerialLength{}}{\includegraphics[width=\MultipleSerialLength{}]{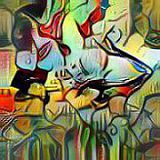}}&
\parbox[c]{\MultipleSerialLength{}}{\includegraphics[width=\MultipleSerialLength{}]{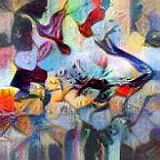}}&
\parbox[c]{\MultipleSerialLength{}}{\includegraphics[width=\MultipleSerialLength{}]{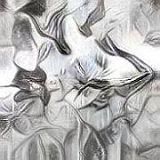}}&
\parbox[c]{\MultipleSerialLength{}}{\includegraphics[width=\MultipleSerialLength{}]{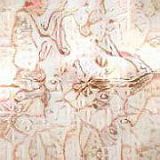}}\\

\parbox[c]{0.5em}{(3)}&
\parbox[c]{\MultipleSerialLength{}}{\includegraphics[width=\MultipleSerialLength{}]{appendix_figures/content_img/sailboat.jpg}}&
\parbox[c]{\MultipleSerialLength{}}{\includegraphics[width=\MultipleSerialLength{}]{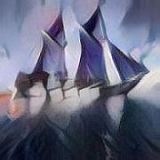}}&
\parbox[c]{\MultipleSerialLength{}}{\includegraphics[width=\MultipleSerialLength{}]{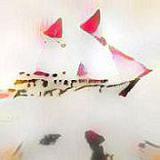}}&
\parbox[c]{\MultipleSerialLength{}}{\includegraphics[width=\MultipleSerialLength{}]{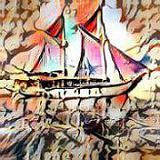}}&
\parbox[c]{\MultipleSerialLength{}}{\includegraphics[width=\MultipleSerialLength{}]{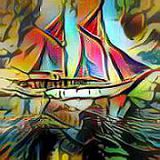}}&
\parbox[c]{\MultipleSerialLength{}}{\includegraphics[width=\MultipleSerialLength{}]{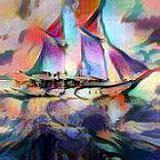}}&
\parbox[c]{\MultipleSerialLength{}}{\includegraphics[width=\MultipleSerialLength{}]{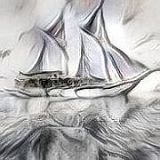}}&
\parbox[c]{\MultipleSerialLength{}}{\includegraphics[width=\MultipleSerialLength{}]{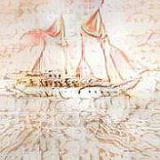}}\\

\parbox[c]{0.5em}{(4)}&
\parbox[c]{\MultipleSerialLength{}}{\includegraphics[width=\MultipleSerialLength{}]{appendix_figures/content_img/sailboat.jpg}}&
\parbox[c]{\MultipleSerialLength{}}{\includegraphics[width=\MultipleSerialLength{}]{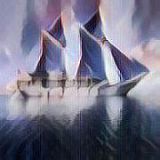}}&
\parbox[c]{\MultipleSerialLength{}}{\includegraphics[width=\MultipleSerialLength{}]{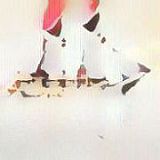}}&
\parbox[c]{\MultipleSerialLength{}}{\includegraphics[width=\MultipleSerialLength{}]{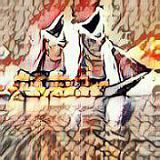}}&
\parbox[c]{\MultipleSerialLength{}}{\includegraphics[width=\MultipleSerialLength{}]{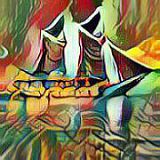}}&
\parbox[c]{\MultipleSerialLength{}}{\includegraphics[width=\MultipleSerialLength{}]{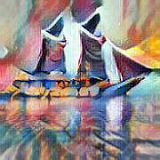}}&
\parbox[c]{\MultipleSerialLength{}}{\includegraphics[width=\MultipleSerialLength{}]{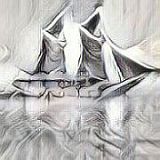}}&
\parbox[c]{\MultipleSerialLength{}}{\includegraphics[width=\MultipleSerialLength{}]{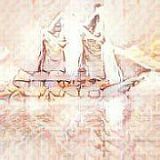}}\\
\end{tabular}

\caption{\small{Three sets of example results of serial style transfer for multiple times. The top row contains the style images used in each serial style transfer.
The rows in each set sequentially show the results generated by (1) Gatys \etal~\cite{Gatys2016ImageStyleTransferUsingCNN}, (2) AdaIN~\cite{Huang2017AdaIN}, (3) our two-stage model, and (4) our end-to-end model. Except the leftmost column, which are the content images, every stylized image is generated with the content feature of the image in its left, and the style feature of the image at the top of the column. The content of the results produced by our proposed models are less distorted by the intermediate style transfer operations.}}
\label{fig:result_multi_serial}
\end{figure*}


\newcommand\AdversarialAblationLength{0.19\columnwidth}

\begin{figure*}[htbp]
 \centering
 \begin{tabular}{cccccccc}
\parbox[c]{0.5em}{(1)}&
\parbox[c]{\AdversarialAblationLength{}}{\includegraphics[width=\AdversarialAblationLength{}]{appendix_figures/content_img/avril.jpg}}&
\parbox[c]{\AdversarialAblationLength{}}{\includegraphics[width=\AdversarialAblationLength{}]{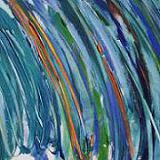}}&
\parbox[c]{\AdversarialAblationLength{}}{\includegraphics[width=\AdversarialAblationLength{}]{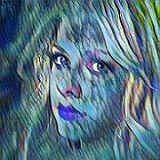}}&
\parbox[c]{\AdversarialAblationLength{}}{\includegraphics[width=\AdversarialAblationLength{}]{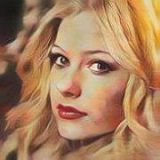}}&
\parbox[c]{\AdversarialAblationLength{}}{\includegraphics[width=\AdversarialAblationLength{}]{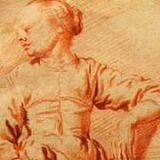}}&
\parbox[c]{\AdversarialAblationLength{}}{\includegraphics[width=\AdversarialAblationLength{}]{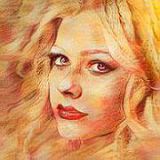}}&
\parbox[c]{\AdversarialAblationLength{}}{\includegraphics[width=\AdversarialAblationLength{}]{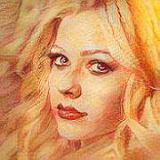}}\\
\parbox[c]{0.5em}{(2)}&
\parbox[c]{\AdversarialAblationLength{}}{\includegraphics[width=\AdversarialAblationLength{}]{appendix_figures/content_img/avril.jpg}}&
\parbox[c]{\AdversarialAblationLength{}}{\includegraphics[width=\AdversarialAblationLength{}]{appendix_figures/style_img/brushstrokes.jpg}}&
\parbox[c]{\AdversarialAblationLength{}}{\includegraphics[width=\AdversarialAblationLength{}]{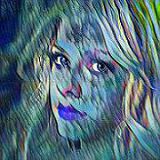}}&
\parbox[c]{\AdversarialAblationLength{}}{\includegraphics[width=\AdversarialAblationLength{}]{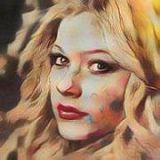}}&
\parbox[c]{\AdversarialAblationLength{}}{\includegraphics[width=\AdversarialAblationLength{}]{appendix_figures/style_img/woman_in_peasant_dress_cropped.jpg}}&
\parbox[c]{\AdversarialAblationLength{}}{\includegraphics[width=\AdversarialAblationLength{}]{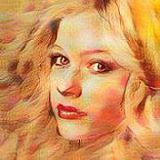}}&
\parbox[c]{\AdversarialAblationLength{}}{\includegraphics[width=\AdversarialAblationLength{}]{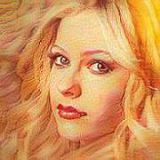}}\\
\hline

\parbox[c]{0.5em}{(1)}&
\parbox[c]{\AdversarialAblationLength{}}{\includegraphics[width=\AdversarialAblationLength{}]{appendix_figures/content_img/blonde_girl.jpg}}&
\parbox[c]{\AdversarialAblationLength{}}{\includegraphics[width=\AdversarialAblationLength{}]{appendix_figures/style_img/impronte_d_artista.jpg}}&
\parbox[c]{\AdversarialAblationLength{}}{\includegraphics[width=\AdversarialAblationLength{}]{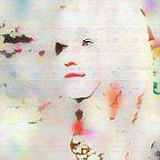}}&
\parbox[c]{\AdversarialAblationLength{}}{\includegraphics[width=\AdversarialAblationLength{}]{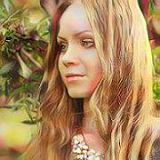}}&
\parbox[c]{\AdversarialAblationLength{}}{\includegraphics[width=\AdversarialAblationLength{}]{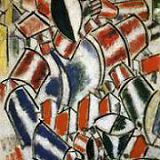}}&
\parbox[c]{\AdversarialAblationLength{}}{\includegraphics[width=\AdversarialAblationLength{}]{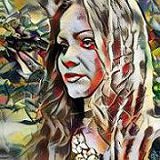}}&
\parbox[c]{\AdversarialAblationLength{}}{\includegraphics[width=\AdversarialAblationLength{}]{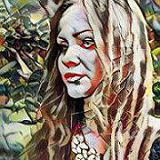}}\\
\parbox[c]{0.5em}{(2)}&
\parbox[c]{\AdversarialAblationLength{}}{\includegraphics[width=\AdversarialAblationLength{}]{appendix_figures/content_img/blonde_girl.jpg}}&
\parbox[c]{\AdversarialAblationLength{}}{\includegraphics[width=\AdversarialAblationLength{}]{appendix_figures/style_img/impronte_d_artista.jpg}}&
\parbox[c]{\AdversarialAblationLength{}}{\includegraphics[width=\AdversarialAblationLength{}]{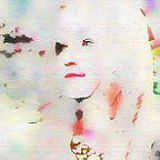}}&
\parbox[c]{\AdversarialAblationLength{}}{\includegraphics[width=\AdversarialAblationLength{}]{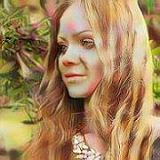}}&
\parbox[c]{\AdversarialAblationLength{}}{\includegraphics[width=\AdversarialAblationLength{}]{appendix_figures/style_img/contrast_of_forms.jpg}}&
\parbox[c]{\AdversarialAblationLength{}}{\includegraphics[width=\AdversarialAblationLength{}]{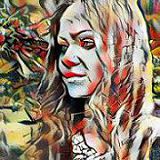}}&
\parbox[c]{\AdversarialAblationLength{}}{\includegraphics[width=\AdversarialAblationLength{}]{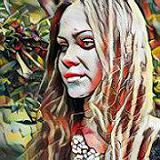}}\\
\hline

\parbox[c]{0.5em}{(1)}&
\parbox[c]{\AdversarialAblationLength{}}{\includegraphics[width=\AdversarialAblationLength{}]{appendix_figures/content_img/chicago.jpg}}&
\parbox[c]{\AdversarialAblationLength{}}{\includegraphics[width=\AdversarialAblationLength{}]{appendix_figures/style_img/scene_de_rue.jpg}}&
\parbox[c]{\AdversarialAblationLength{}}{\includegraphics[width=\AdversarialAblationLength{}]{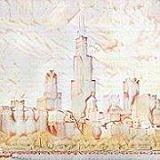}}&
\parbox[c]{\AdversarialAblationLength{}}{\includegraphics[width=\AdversarialAblationLength{}]{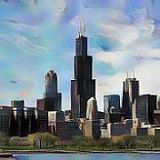}}&
\parbox[c]{\AdversarialAblationLength{}}{\includegraphics[width=\AdversarialAblationLength{}]{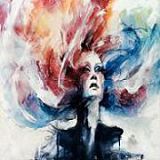}}&
\parbox[c]{\AdversarialAblationLength{}}{\includegraphics[width=\AdversarialAblationLength{}]{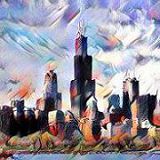}}&
\parbox[c]{\AdversarialAblationLength{}}{\includegraphics[width=\AdversarialAblationLength{}]{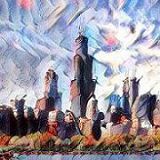}}\\
\parbox[c]{0.5em}{(2)}&
\parbox[c]{\AdversarialAblationLength{}}{\includegraphics[width=\AdversarialAblationLength{}]{appendix_figures/content_img/chicago.jpg}}&
\parbox[c]{\AdversarialAblationLength{}}{\includegraphics[width=\AdversarialAblationLength{}]{appendix_figures/style_img/scene_de_rue.jpg}}&
\parbox[c]{\AdversarialAblationLength{}}{\includegraphics[width=\AdversarialAblationLength{}]{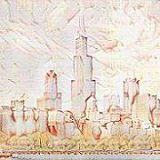}}&
\parbox[c]{\AdversarialAblationLength{}}{\includegraphics[width=\AdversarialAblationLength{}]{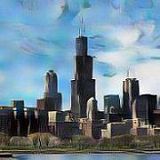}}&
\parbox[c]{\AdversarialAblationLength{}}{\includegraphics[width=\AdversarialAblationLength{}]{appendix_figures/style_img/antimonocromatismo.jpg}}&
\parbox[c]{\AdversarialAblationLength{}}{\includegraphics[width=\AdversarialAblationLength{}]{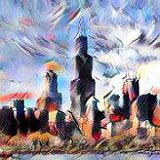}}&
\parbox[c]{\AdversarialAblationLength{}}{\includegraphics[width=\AdversarialAblationLength{}]{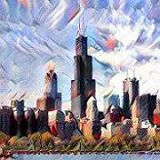}}\\
\hline

\parbox[c]{0.5em}{(1)}&
\parbox[c]{\AdversarialAblationLength{}}{\includegraphics[width=\AdversarialAblationLength{}]{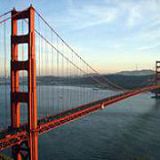}}&
\parbox[c]{\AdversarialAblationLength{}}{\includegraphics[width=\AdversarialAblationLength{}]{appendix_figures/style_img/the_resevoir_at_poitiers.jpg}}&
\parbox[c]{\AdversarialAblationLength{}}{\includegraphics[width=\AdversarialAblationLength{}]{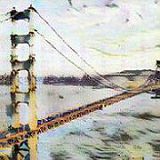}}&
\parbox[c]{\AdversarialAblationLength{}}{\includegraphics[width=\AdversarialAblationLength{}]{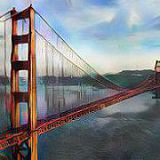}}&
\parbox[c]{\AdversarialAblationLength{}}{\includegraphics[width=\AdversarialAblationLength{}]{appendix_figures/style_img/la_muse.jpg}}&
\parbox[c]{\AdversarialAblationLength{}}{\includegraphics[width=\AdversarialAblationLength{}]{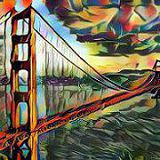}}&
\parbox[c]{\AdversarialAblationLength{}}{\includegraphics[width=\AdversarialAblationLength{}]{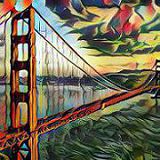}}\\
\parbox[c]{0.5em}{(2)}&
\parbox[c]{\AdversarialAblationLength{}}{\includegraphics[width=\AdversarialAblationLength{}]{appendix_figures/content_img/golden_gate.jpg}}&
\parbox[c]{\AdversarialAblationLength{}}{\includegraphics[width=\AdversarialAblationLength{}]{appendix_figures/style_img/the_resevoir_at_poitiers.jpg}}&
\parbox[c]{\AdversarialAblationLength{}}{\includegraphics[width=\AdversarialAblationLength{}]{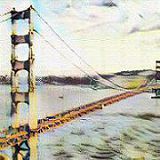}}&
\parbox[c]{\AdversarialAblationLength{}}{\includegraphics[width=\AdversarialAblationLength{}]{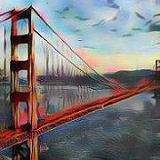}}&
\parbox[c]{\AdversarialAblationLength{}}{\includegraphics[width=\AdversarialAblationLength{}]{appendix_figures/style_img/la_muse.jpg}}&
\parbox[c]{\AdversarialAblationLength{}}{\includegraphics[width=\AdversarialAblationLength{}]{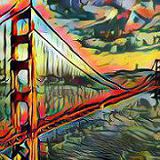}}&
\parbox[c]{\AdversarialAblationLength{}}{\includegraphics[width=\AdversarialAblationLength{}]{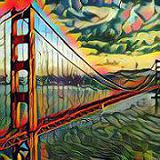}}\\
\hline
 &Content & Style & Regular & Reverse & \begin{tabular}[c]{@{}c@{}}Second\\Style\end{tabular} & Serial & \begin{tabular}[c]{@{}c@{}}Expected\\Serial\end{tabular}\\
 \end{tabular}
 \caption{Comparison between the expected results for reverse and serial issues and the actual results generated w/ adversarial learning (1) and w/o adversarial learning (2).}
 \label{fig:compare_adversarial}
 \end{figure*}

\newcommand\ReplaceAdaINLength{0.17\columnwidth}
\begin{figure*}
\centering
\begin{tabular}{cccccccc}
\parbox[c]{6em}{AdaIN\cite{Huang2017AdaIN}}&
\parbox[c]{\ReplaceAdaINLength{}}{\includegraphics[width=\ReplaceAdaINLength{}]{appendix_figures/content_img/avril.jpg}}&
\parbox[c]{\ReplaceAdaINLength{}}{\includegraphics[width=\ReplaceAdaINLength{}]{appendix_figures/style_img/asheville.jpg}}&
\parbox[c]{\ReplaceAdaINLength{}}{\includegraphics[width=\ReplaceAdaINLength{}]{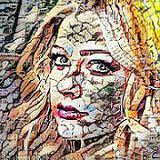}}&
\parbox[c]{\ReplaceAdaINLength{}}{\includegraphics[width=\ReplaceAdaINLength{}]{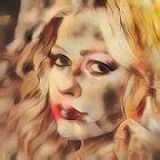}}&
\parbox[c]{\ReplaceAdaINLength{}}{\includegraphics[width=\ReplaceAdaINLength{}]{appendix_figures/style_img/contrast_of_forms.jpg}}&
\parbox[c]{\ReplaceAdaINLength{}}{\includegraphics[width=\ReplaceAdaINLength{}]{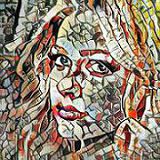}}&
\parbox[c]{\ReplaceAdaINLength{}}{\includegraphics[width=\ReplaceAdaINLength{}]{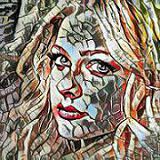}}\\
\parbox[c]{6em}{\begin{tabular}[c]{@{}c@{}}Two-Stage\\AdaIN\cite{Huang2017AdaIN}\end{tabular}}&
\parbox[c]{\ReplaceAdaINLength{}}{\includegraphics[width=\ReplaceAdaINLength{}]{appendix_figures/content_img/avril.jpg}}&
\parbox[c]{\ReplaceAdaINLength{}}{\includegraphics[width=\ReplaceAdaINLength{}]{appendix_figures/style_img/asheville.jpg}}&
\parbox[c]{\ReplaceAdaINLength{}}{\includegraphics[width=\ReplaceAdaINLength{}]{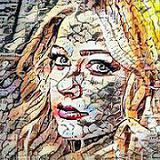}}&
\parbox[c]{\ReplaceAdaINLength{}}{\includegraphics[width=\ReplaceAdaINLength{}]{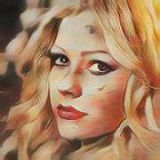}}&
\parbox[c]{\ReplaceAdaINLength{}}{\includegraphics[width=\ReplaceAdaINLength{}]{appendix_figures/style_img/contrast_of_forms.jpg}}&
\parbox[c]{\ReplaceAdaINLength{}}{\includegraphics[width=\ReplaceAdaINLength{}]{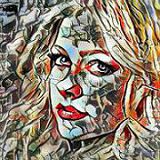}}&
\parbox[c]{\ReplaceAdaINLength{}}{\includegraphics[width=\ReplaceAdaINLength{}]{appendix_figures/Other_two-stage/AdaIN/avril_stylized_contrast_of_forms.jpg}}\\
\parbox[c]{6em}{AdaIN\cite{Huang2017AdaIN}}&
\parbox[c]{\ReplaceAdaINLength{}}{\includegraphics[width=\ReplaceAdaINLength{}]{appendix_figures/content_img/chicago.jpg}}&
\parbox[c]{\ReplaceAdaINLength{}}{\includegraphics[width=\ReplaceAdaINLength{}]{appendix_figures/style_img/sketch.png}}&
\parbox[c]{\ReplaceAdaINLength{}}{\includegraphics[width=\ReplaceAdaINLength{}]{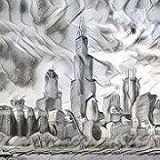}}&
\parbox[c]{\ReplaceAdaINLength{}}{\includegraphics[width=\ReplaceAdaINLength{}]{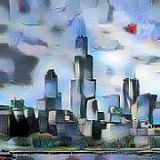}}&
\parbox[c]{\ReplaceAdaINLength{}}{\includegraphics[width=\ReplaceAdaINLength{}]{appendix_figures/style_img/brushstrokes.jpg}}&
\parbox[c]{\ReplaceAdaINLength{}}{\includegraphics[width=\ReplaceAdaINLength{}]{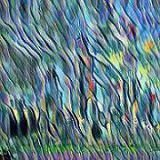}}&
\parbox[c]{\ReplaceAdaINLength{}}{\includegraphics[width=\ReplaceAdaINLength{}]{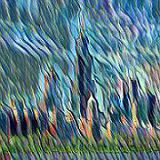}}\\
\parbox[c]{6em}{\begin{tabular}[c]{@{}c@{}}Two-Stage\\AdaIN\cite{Huang2017AdaIN}\end{tabular}}&
\parbox[c]{\ReplaceAdaINLength{}}{\includegraphics[width=\ReplaceAdaINLength{}]{appendix_figures/content_img/chicago.jpg}}&
\parbox[c]{\ReplaceAdaINLength{}}{\includegraphics[width=\ReplaceAdaINLength{}]{appendix_figures/style_img/sketch.png}}&
\parbox[c]{\ReplaceAdaINLength{}}{\includegraphics[width=\ReplaceAdaINLength{}]{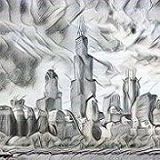}}&
\parbox[c]{\ReplaceAdaINLength{}}{\includegraphics[width=\ReplaceAdaINLength{}]{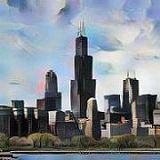}}&
\parbox[c]{\ReplaceAdaINLength{}}{\includegraphics[width=\ReplaceAdaINLength{}]{appendix_figures/style_img/brushstrokes.jpg}}&
\parbox[c]{\ReplaceAdaINLength{}}{\includegraphics[width=\ReplaceAdaINLength{}]{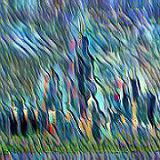}}&
\parbox[c]{\ReplaceAdaINLength{}}{\includegraphics[width=\ReplaceAdaINLength{}]{appendix_figures/Other_two-stage/AdaIN/chicago_stylized_brushstrokes.jpg}}\\
& Content & Style & Regular & Reverse & \begin{tabular}[c]{@{}c@{}}Second\\Style\end{tabular} & Serial & \begin{tabular}[c]{@{}c@{}}Expected\\Serial\end{tabular}\\
\parbox[c]{6em}{WCT~\cite{Li2017WCT}}&
\parbox[c]{\ReplaceAdaINLength{}}{\includegraphics[width=\ReplaceAdaINLength{}]{appendix_figures/content_img/avril.jpg}}&
\parbox[c]{\ReplaceAdaINLength{}}{\includegraphics[width=\ReplaceAdaINLength{}]{appendix_figures/style_img/asheville.jpg}}&
\parbox[c]{\ReplaceAdaINLength{}}{\includegraphics[width=\ReplaceAdaINLength{}]{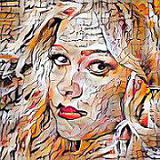}}&
\parbox[c]{\ReplaceAdaINLength{}}{\includegraphics[width=\ReplaceAdaINLength{}]{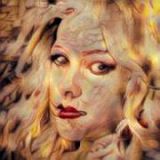}}&
\parbox[c]{\ReplaceAdaINLength{}}{\includegraphics[width=\ReplaceAdaINLength{}]{appendix_figures/style_img/contrast_of_forms.jpg}}&
\parbox[c]{\ReplaceAdaINLength{}}{\includegraphics[width=\ReplaceAdaINLength{}]{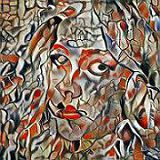}}&
\parbox[c]{\ReplaceAdaINLength{}}{\includegraphics[width=\ReplaceAdaINLength{}]{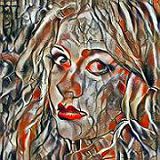}}\\
\parbox[c]{6em}{\begin{tabular}[c]{@{}c@{}}Two-Stage\\ WCT\\\cite{Li2017WCT}\end{tabular}}&
\parbox[c]{\ReplaceAdaINLength{}}{\includegraphics[width=\ReplaceAdaINLength{}]{appendix_figures/content_img/avril.jpg}}&
\parbox[c]{\ReplaceAdaINLength{}}{\includegraphics[width=\ReplaceAdaINLength{}]{appendix_figures/style_img/asheville.jpg}}&
\parbox[c]{\ReplaceAdaINLength{}}{\includegraphics[width=\ReplaceAdaINLength{}]{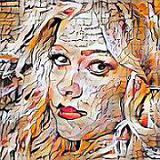}}&
\parbox[c]{\ReplaceAdaINLength{}}{\includegraphics[width=\ReplaceAdaINLength{}]{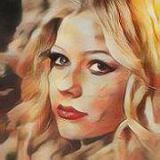}}&
\parbox[c]{\ReplaceAdaINLength{}}{\includegraphics[width=\ReplaceAdaINLength{}]{appendix_figures/style_img/contrast_of_forms.jpg}}&
\parbox[c]{\ReplaceAdaINLength{}}{\includegraphics[width=\ReplaceAdaINLength{}]{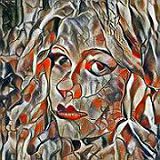}}&
\parbox[c]{\ReplaceAdaINLength{}}{\includegraphics[width=\ReplaceAdaINLength{}]{appendix_figures/Other_two-stage/WCT/avril_stylized_contrast_of_forms.jpg}}\\
\parbox[c]{6em}{WCT~\cite{Li2017WCT}}&
\parbox[c]{\ReplaceAdaINLength{}}{\includegraphics[width=\ReplaceAdaINLength{}]{appendix_figures/content_img/chicago.jpg}}&
\parbox[c]{\ReplaceAdaINLength{}}{\includegraphics[width=\ReplaceAdaINLength{}]{appendix_figures/style_img/sketch.png}}&
\parbox[c]{\ReplaceAdaINLength{}}{\includegraphics[width=\ReplaceAdaINLength{}]{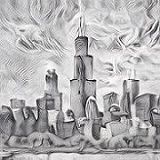}}&
\parbox[c]{\ReplaceAdaINLength{}}{\includegraphics[width=\ReplaceAdaINLength{}]{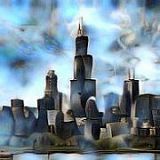}}&
\parbox[c]{\ReplaceAdaINLength{}}{\includegraphics[width=\ReplaceAdaINLength{}]{appendix_figures/style_img/brushstrokes.jpg}}&
\parbox[c]{\ReplaceAdaINLength{}}{\includegraphics[width=\ReplaceAdaINLength{}]{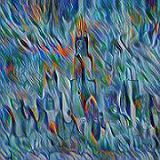}}&
\parbox[c]{\ReplaceAdaINLength{}}{\includegraphics[width=\ReplaceAdaINLength{}]{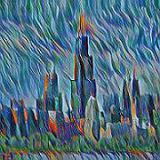}}\\
\parbox[c]{6em}{\begin{tabular}[c]{@{}c@{}}Two-Stage\\ WCT\\\cite{Li2017WCT}\end{tabular}}&
\parbox[c]{\ReplaceAdaINLength{}}{\includegraphics[width=\ReplaceAdaINLength{}]{appendix_figures/content_img/chicago.jpg}}&
\parbox[c]{\ReplaceAdaINLength{}}{\includegraphics[width=\ReplaceAdaINLength{}]{appendix_figures/style_img/sketch.png}}&
\parbox[c]{\ReplaceAdaINLength{}}{\includegraphics[width=\ReplaceAdaINLength{}]{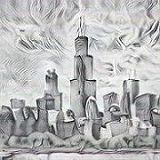}}&
\parbox[c]{\ReplaceAdaINLength{}}{\includegraphics[width=\ReplaceAdaINLength{}]{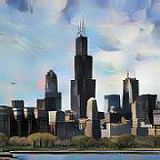}}&
\parbox[c]{\ReplaceAdaINLength{}}{\includegraphics[width=\ReplaceAdaINLength{}]{appendix_figures/style_img/brushstrokes.jpg}}&
\parbox[c]{\ReplaceAdaINLength{}}{\includegraphics[width=\ReplaceAdaINLength{}]{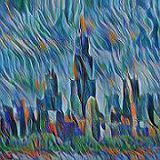}}&
\parbox[c]{\ReplaceAdaINLength{}}{\includegraphics[width=\ReplaceAdaINLength{}]{appendix_figures/Other_two-stage/WCT/chicago_stylized_brushstrokes.jpg}}\\
& Content & Style & Regular & Reverse & \begin{tabular}[c]{@{}c@{}}Second\\Style\end{tabular} & Serial & \begin{tabular}[c]{@{}c@{}}Expected\\Serial\end{tabular}\\
\parbox[c]{6em}{\begin{tabular}[c]{@{}c@{}}Ulyanov \etal\\\cite{Ulyanov2017InstanceNormalization}\end{tabular}}&
\parbox[c]{\ReplaceAdaINLength{}}{\includegraphics[width=\ReplaceAdaINLength{}]{appendix_figures/content_img/avril.jpg}}&
\parbox[c]{\ReplaceAdaINLength{}}{\includegraphics[width=\ReplaceAdaINLength{}]{appendix_figures/style_img/asheville.jpg}}&
\parbox[c]{\ReplaceAdaINLength{}}{\includegraphics[width=\ReplaceAdaINLength{}]{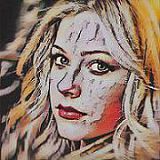}}&
\parbox[c]{\ReplaceAdaINLength{}}{\includegraphics[width=\ReplaceAdaINLength{}]{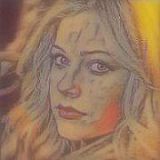}}&
\parbox[c]{\ReplaceAdaINLength{}}{\includegraphics[width=\ReplaceAdaINLength{}]{appendix_figures/style_img/contrast_of_forms.jpg}}&
\parbox[c]{\ReplaceAdaINLength{}}{\includegraphics[width=\ReplaceAdaINLength{}]{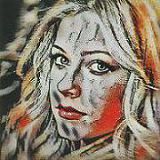}}&
\parbox[c]{\ReplaceAdaINLength{}}{\includegraphics[width=\ReplaceAdaINLength{}]{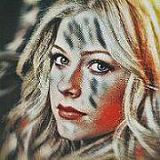}}\\
\parbox[c]{6em}{\begin{tabular}[c]{@{}c@{}}Two-Stage\\Ulyanov \etal\\\cite{Ulyanov2017InstanceNormalization}\end{tabular}}&
\parbox[c]{\ReplaceAdaINLength{}}{\includegraphics[width=\ReplaceAdaINLength{}]{appendix_figures/content_img/avril.jpg}}&
\parbox[c]{\ReplaceAdaINLength{}}{\includegraphics[width=\ReplaceAdaINLength{}]{appendix_figures/style_img/asheville.jpg}}&
\parbox[c]{\ReplaceAdaINLength{}}{\includegraphics[width=\ReplaceAdaINLength{}]{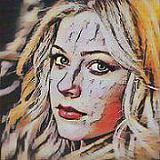}}&
\parbox[c]{\ReplaceAdaINLength{}}{\includegraphics[width=\ReplaceAdaINLength{}]{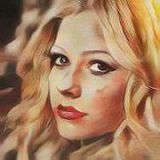}}&
\parbox[c]{\ReplaceAdaINLength{}}{\includegraphics[width=\ReplaceAdaINLength{}]{appendix_figures/style_img/contrast_of_forms.jpg}}&
\parbox[c]{\ReplaceAdaINLength{}}{\includegraphics[width=\ReplaceAdaINLength{}]{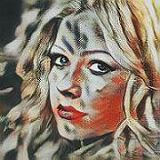}}&
\parbox[c]{\ReplaceAdaINLength{}}{\includegraphics[width=\ReplaceAdaINLength{}]{appendix_figures/Other_two-stage/Johnson/avril_stylized_contrast_of_forms.jpg}}\\
\parbox[c]{6em}{\begin{tabular}[c]{@{}c@{}}Ulyanov \etal\\\cite{Ulyanov2017InstanceNormalization}\end{tabular}}&
\parbox[c]{\ReplaceAdaINLength{}}{\includegraphics[width=\ReplaceAdaINLength{}]{appendix_figures/content_img/chicago.jpg}}&
\parbox[c]{\ReplaceAdaINLength{}}{\includegraphics[width=\ReplaceAdaINLength{}]{appendix_figures/style_img/sketch.png}}&
\parbox[c]{\ReplaceAdaINLength{}}{\includegraphics[width=\ReplaceAdaINLength{}]{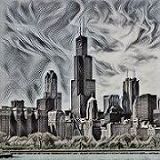}}&
\parbox[c]{\ReplaceAdaINLength{}}{\includegraphics[width=\ReplaceAdaINLength{}]{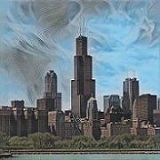}}&
\parbox[c]{\ReplaceAdaINLength{}}{\includegraphics[width=\ReplaceAdaINLength{}]{appendix_figures/style_img/brushstrokes.jpg}}&
\parbox[c]{\ReplaceAdaINLength{}}{\includegraphics[width=\ReplaceAdaINLength{}]{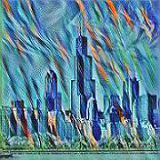}}&
\parbox[c]{\ReplaceAdaINLength{}}{\includegraphics[width=\ReplaceAdaINLength{}]{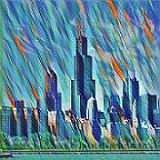}}\\
\parbox[c]{6em}{\begin{tabular}[c]{@{}c@{}}Two-Stage\\Ulyanov \etal\\\cite{Ulyanov2017InstanceNormalization}\end{tabular}}&
\parbox[c]{\ReplaceAdaINLength{}}{\includegraphics[width=\ReplaceAdaINLength{}]{appendix_figures/content_img/chicago.jpg}}&
\parbox[c]{\ReplaceAdaINLength{}}{\includegraphics[width=\ReplaceAdaINLength{}]{appendix_figures/style_img/sketch.png}}&
\parbox[c]{\ReplaceAdaINLength{}}{\includegraphics[width=\ReplaceAdaINLength{}]{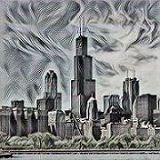}}&
\parbox[c]{\ReplaceAdaINLength{}}{\includegraphics[width=\ReplaceAdaINLength{}]{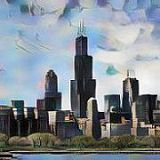}}&
\parbox[c]{\ReplaceAdaINLength{}}{\includegraphics[width=\ReplaceAdaINLength{}]{appendix_figures/style_img/brushstrokes.jpg}}&
\parbox[c]{\ReplaceAdaINLength{}}{\includegraphics[width=\ReplaceAdaINLength{}]{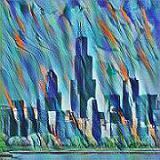}}&
\parbox[c]{\ReplaceAdaINLength{}}{\includegraphics[width=\ReplaceAdaINLength{}]{appendix_figures/Other_two-stage/Johnson/chicago_stylized_brushstrokes.jpg}}\\
& Content & Style & Regular & Reverse & \begin{tabular}[c]{@{}c@{}}Second\\Style\end{tabular} & Serial & \begin{tabular}[c]{@{}c@{}}Expected\\Serial\end{tabular}\\
\end{tabular}
\caption{Three sets of results to demonstrate the comparison between adopting different methods for the style transfer stage of our two-stage model to perform regular, reverse, and serial style transfer. The rows in each set sequentially show the results generated by (1) AdaIN~\cite{Huang2017AdaIN} and our two-stage model with AdaIN~\cite{Huang2017AdaIN}, (2) WCT~\cite{Li2017WCT} and our two-stage model with WCT~\cite{Li2017WCT}, and (3) instance normalization~\cite{Ulyanov2017InstanceNormalization} and our two-stage with instance normalization~\cite{Ulyanov2017InstanceNormalization}}
\label{fig:two-stage_other_method}
\end{figure*}

%% file: main.bbl
\begin{thebibliography}{10}\itemsep=-1pt

\bibitem{Baluja2017DeepSteganography}
S.~Baluja.
\newblock Hiding images in plain sight: Deep steganography.
\newblock In {\em Advances in Neural Information Processing Systems (NIPS)},
  2017.

\bibitem{cheddad2010digital}
A.~Cheddad, J.~Condell, K.~Curran, and P.~Mc~Kevitt.
\newblock Digital image steganography: Survey and analysis of current methods.
\newblock {\em Signal Processing}, 2010.

\bibitem{ImageQuilting}
A.~A. Efros and W.~T. Freeman.
\newblock Image quilting for texture synthesis and transfer.
\newblock {\em ACM Transactions on Graphics (TOG)}, 2001.

\bibitem{Gatys2015TextureSynthesisUsingCNN}
L.~A. Gatys, A.~S. Ecker, and M.~Bethge.
\newblock Texture synthesis using convolutional neural networks.
\newblock In {\em Advances in Neural Information Processing Systems (NIPS)},
  2015.

\bibitem{Gatys2016ImageStyleTransferUsingCNN}
L.~A. Gatys, A.~S. Ecker, and M.~Bethge.
\newblock Image style transfer using convolutional neural networks.
\newblock In {\em IEEE Conference on Computer Vision and Pattern Recognition
  (CVPR)}, 2016.

\bibitem{Gupta2012LSBSteganography}
S.~Gupta, A.~Goyal, and B.~Bhushan.
\newblock Information hiding using least significant bit steganography and
  cryptography.
\newblock {\em International Journal of Modern Education and Computer Science
  (IJMECS)}, 2012.

\bibitem{hayes2017generating}
J.~Hayes and G.~Danezis.
\newblock Generating steganographic images via adversarial training.
\newblock In {\em Advances in Neural Information Processing Systems (NIPS)},
  2017.

\bibitem{ImageAnalogies}
A.~Hertzmann, C.~E. Jacobs, N.~Oliver, B.~Curless, and D.~H. Salesin.
\newblock Image analogies.
\newblock {\em ACM Transactions on Graphics (TOG)}, 2001.

\bibitem{Huang2017AdaIN}
X.~Huang and S.~Belongie.
\newblock Arbitrary style transfer in real-time with adaptive instance
  normalization.
\newblock In {\em IEEE International Conference on Computer Vision (ICCV)},
  2017.

\bibitem{isola2017image}
P.~Isola, J.-Y. Zhu, T.~Zhou, and A.~A. Efros.
\newblock Image-to-image translation with conditional adversarial networks.
\newblock In {\em IEEE Conference on Computer Vision and Pattern Recognition
  (CVPR)}, 2017.

\bibitem{Johnson2016Perceptual}
J.~Johnson, A.~Alahi, and L.~Fei-Fei.
\newblock Perceptual losses for real-time style transfer and super-resolution.
\newblock In {\em European Conference on Computer Vision (ECCV)}, 2016.

\bibitem{jolicoeur2018relativistic}
A.~Jolicoeur-Martineau.
\newblock The relativistic discriminator: a key element missing from standard
  gan.
\newblock In {\em International Conference on Learning Representations (ICLR)},
  2019.

\bibitem{kessler2011overview}
G.~C. Kessler and C.~Hosmer.
\newblock An overview of steganography.
\newblock {\em Advances in Computers}, 2011.

\bibitem{kingma2014adam}
D.~P. Kingma and J.~Ba.
\newblock Adam: A method for stochastic optimization.
\newblock In {\em International Conference on Learning Representations (ICLR)},
  2015.

\bibitem{Li2017WCT}
Y.~Li, C.~Fang, J.~Yang, Z.~Wang, X.~Lu, and M.-H. Yang.
\newblock Universal style transfer via feature transforms.
\newblock In {\em Advances in Neural Information Processing Systems (NIPS)},
  2017.

\bibitem{lin2014microsoft}
T.-Y. Lin, M.~Maire, S.~Belongie, J.~Hays, P.~Perona, D.~Ramanan,
  P.~Doll{\'a}r, and C.~L. Zitnick.
\newblock Microsoft coco: Common objects in context.
\newblock In {\em European Conference on Computer Vision (ECCV)}, 2014.

\bibitem{wiki_art}
K.~Nichol.
\newblock Painter by numbers, wikiart.
\newblock \url{https://www.kaggle.com/c/painter-by-numbers}, 2016.

\bibitem{PyTorch}
A.~Paszke, S.~Gross, S.~Chintala, G.~Chanan, E.~Yang, Z.~DeVito, Z.~Lin,
  A.~Desmaison, L.~Antiga, and A.~Lerer.
\newblock Automatic differentiation in {PyTorch}.
\newblock In {\em NeurIPS Autodiff Workshop}, 2017.

\bibitem{Pevny2010HUGO}
T.~Pevný, T.~Filler, and P.~Bas.
\newblock Using high-dimensional image models to perform highly undetectable
  steganography.
\newblock In {\em Proceedings of the International Conference on Information
  Hiding (IH)}, 2010.

\bibitem{Sheng2018Avatar-Net}
L.~Sheng, Z.~Lin, J.~Shao, and X.~Wang.
\newblock Avatar-net: Multi-scale zero-shot style transfer by feature
  decoration.
\newblock In {\em IEEE Conference on Computer Vision and Pattern Recognition
  (CVPR)}, 2018.

\bibitem{shih2013data}
Y.~Shih, S.~Paris, F.~Durand, and W.~T. Freeman.
\newblock Data-driven hallucination of different times of day from a single
  outdoor photo.
\newblock {\em ACM Transactions on Graphics (TOG)}, 2013.

\bibitem{FaceDestylization}
F.~Shiri, X.~Yu, P.~Koniuszand, and F.~Porikli.
\newblock Face destylization.
\newblock In {\em Proceedings of the International Conference on Digital Image
  Computing: Techniques and Applications (DICTA)}, 2017.

\bibitem{FaceRecoveryfromPortraits}
F.~Shiri, X.~Yu, F.~Porikli, R.~Hartley, and P.~Koniusz.
\newblock Identity-preserving face recovery from portraits.
\newblock In {\em IEEE Winter Conference on Applications of Computer Vision
  (WACV)}, 2018.

\bibitem{VGG}
K.~Simonyan and A.~Zisserman.
\newblock Very deep convolutional networks for large-scale image recognition.
\newblock In {\em International Conference on Learning Representations (ICLR)},
  2015.

\bibitem{tomei2019art2real}
M.~Tomei, M.~Cornia, L.~Baraldi, and R.~Cucchiara.
\newblock {Art2Real: Unfolding the Reality of Artworks via Semantically-Aware
  Image-to-Image Translation}.
\newblock In {\em IEEE Conference on Computer Vision and Pattern Recognition
  (CVPR)}, 2019.

\bibitem{Ulyanov2016TextureNetworks}
D.~Ulyanov, V.~Lebedev, A.~Vedaldi, and V.~Lempitsky.
\newblock Texture networks: Feed-forward synthesis of textures and stylized
  images.
\newblock In {\em International Conference on Machine Learning (ICML)}, 2016.

\bibitem{Ulyanov2017InstanceNormalization}
D.~Ulyanov, A.~Vedaldi, and V.~Lempitsky.
\newblock Improved texture networks: Maximizing quality and diversity in
  feed-forward stylization and texture synthesis.
\newblock In {\em IEEE Conference on Computer Vision and Pattern Recognition
  (CVPR)}, 2017.

\bibitem{Dumoulin2017ConditionalIN}
M.~K. Vincent~Dumoulin, Jonathon~Shlens.
\newblock A learned representation for artistic style.
\newblock In {\em International Conference on Learning Representations (ICLR)},
  2017.

\bibitem{zhang2018perceptual}
R.~Zhang, P.~Isola, A.~A. Efros, E.~Shechtman, and O.~Wang.
\newblock The unreasonable effectiveness of deep features as a perceptual
  metric.
\newblock In {\em IEEE Conference on Computer Vision and Pattern Recognition
  (CVPR)}, 2018.

\bibitem{HiDDeN}
J.~Zhu, R.~Kaplan, J.~Johnson, and L.~Fei-Fei.
\newblock Hidden: Hiding data with deep networks.
\newblock In {\em European Conference on Computer Vision (ECCV)}, 2018.

\bibitem{CycleGAN2017}
J.-Y. Zhu, T.~Park, P.~Isola, and A.~A. Efros.
\newblock Unpaired image-to-image translation using cycle-consistent
  adversarial networks.
\newblock In {\em IEEE International Conference on Computer Vision (ICCV)},
  2017.

\end{thebibliography}
